\documentclass{article}
\usepackage{iclr2026_conference,times}

\usepackage{amsmath,amsfonts,bm}

\def\eqref#1{equation~\ref{#1}}

\def\1{\bm{1}}

\DeclareMathAlphabet{\mathsfit}{\encodingdefault}{\sfdefault}{m}{sl}
\SetMathAlphabet{\mathsfit}{bold}{\encodingdefault}{\sfdefault}{bx}{n}

\usepackage[utf8]{inputenc}
\usepackage[T1]{fontenc}
\usepackage{textcomp}
\usepackage{hyperref}
\usepackage{url}
\usepackage{graphicx}
\usepackage{booktabs}
\usepackage{float}
\usepackage{amsmath}
\usepackage{amssymb}
\usepackage{longtable}
\usepackage{array}
\usepackage{calc}
\usepackage{multirow}
\usepackage{xcolor}

\DeclareUnicodeCharacter{00D7}{\ensuremath{\times}}      
\DeclareUnicodeCharacter{2264}{\ensuremath{\leq}}        
\DeclareUnicodeCharacter{2265}{\ensuremath{\geq}}        
\DeclareUnicodeCharacter{00B7}{\ensuremath{\cdot}}       
\DeclareUnicodeCharacter{00A7}{\S}                       
\DeclareUnicodeCharacter{2192}{\ensuremath{\rightarrow}} 
\DeclareUnicodeCharacter{2014}{---}                      
\DeclareUnicodeCharacter{2013}{--}                       
\DeclareUnicodeCharacter{03B1}{\ensuremath{\alpha}}      
\DeclareUnicodeCharacter{03B3}{\ensuremath{\gamma}}      
\DeclareUnicodeCharacter{03C3}{\ensuremath{\sigma}}      
\DeclareUnicodeCharacter{0394}{\ensuremath{\Delta}}      
\DeclareUnicodeCharacter{03BA}{\ensuremath{\kappa}}      
\DeclareUnicodeCharacter{2212}{\ensuremath{-}}           
\DeclareUnicodeCharacter{00B1}{\ensuremath{\pm}}         
\DeclareUnicodeCharacter{2248}{\ensuremath{\approx}}     
\DeclareUnicodeCharacter{2715}{\ensuremath{\times}}      
\DeclareUnicodeCharacter{2717}{\ding{55}}                
\DeclareUnicodeCharacter{2713}{\ding{51}}                
\DeclareUnicodeCharacter{2705}{\ding{51}}                
\DeclareUnicodeCharacter{274C}{\ding{55}}                
\DeclareUnicodeCharacter{2016}{\ensuremath{\|}}          
\DeclareUnicodeCharacter{2026}{\ldots}                   
\DeclareUnicodeCharacter{2194}{\ensuremath{\leftrightarrow}} 
\DeclareUnicodeCharacter{2208}{\ensuremath{\in}}         

\usepackage{pifont} 

\providecommand{\tightlist}{%
  \setlength{\itemsep}{0pt}\setlength{\parskip}{0pt}}

\title{Substrate Asymmetry in User-Side Memory: \\ A Diagnostic Framework}

\author{Youwang Deng \\
EpistemicaLab --- Independent Research \\
\texttt{dengyouwang@gmail.com}}

\iclrfinalcopy  

\begin{document}
\raggedbottom
\maketitle
\lhead{Preprint. Under review.}

\begin{abstract}
User-side memory in LLMs is typically scored as a single
``personalization'' capability --- given a user's history, is the output
more user-aware? We show this aggregate metric \textbf{hides
opposite-direction failures}. Memory factorises into at least three
orthogonal axes --- \emph{behavioral consistency} (style, voice), \emph{factual
presence} (recall facts in the history), and \emph{factual absence}
(abstain when a fact is absent) --- and \textbf{no single substrate wins all
three}. Comparing per-user γ-LoRA (a small LoRA adapter trained on
each user's history; γ denotes per-user, not per-task) against
BGE-large dense top-K retrieval on a controlled 50-user synthetic
corpus and a real-data probe \citep{salemi2024lamp}, we find
γ-LoRA decisively wins behavioral style while RAG decisively wins
factual absence --- and the \textbf{same query-projection cells in attention
layers 21--35 causally load-bear both effects in \emph{opposite}
directions} (zeroing those LoRA weights raises absence-probe
true-positive rate by +33 percentage points and drops presence-probe
TPR by 20 pp). On the more heavily RLHF-tuned
Llama-3.1-8B-Instruct the asymmetry \textbf{strengthens, not heals}:
parametric memory's behavioral advantage collapses while its
absence-calibration deficit against retrieval widens --- an
\emph{alignment tax on parametric user-memory}. On real-data LaMP-3,
γ-LoRA underperforms a majority baseline; a 9-condition mitigation
sweep diagnoses this as \textbf{instruction-following collapse, not
substrate failure} (a 9 × 2 mitigation cross-product shows the
eval-time \{1..5\} logit mask drives main\_acc to ≥0.995 on every
training recipe --- recipe choice does not escape the collapse, and
the residual probe2 ceiling 0.605--0.660 is task-structural, not
recipe-tunable), and the best training-time fix replicates
bit-identically on Llama. Finally, substrate-selection
routing is \textbf{question-classification, not calibration}: a 110M
DistilBERT on the question text alone beats every logit-based
router. We contribute the diagnostic framework, the diagnosed
real-data negative, the alignment-tax replication, and the
routing-as-classification finding.

\end{abstract}

\section{Introduction}\label{introduction}

\subsection{Aggregate numbers hide opposite-direction failures}\label{aggregate-numbers-hide-opposite-direction-failures}

Personalization in large language models is usually scored as a
single number: an accuracy on a mixed benchmark, a win-rate against
a no-history baseline, a ``user-aware'' delta over a ``user-unaware''
control. Benchmarks (LaMP, PERSOMA, UQABench) report top-line
metrics that aggregate across heterogeneous task types --- style
imitation, factual recall, preference prediction, classification ---
under a single banner.

The single-number framing has a hidden cost. We show, on a controlled
synthetic-personae corpus and on real LaMP-3, that \textbf{the same
substrate trained on the same user history can win one component of
``personalization'' by +47 points and lose another by −90 points,
simultaneously, on the same 50 users.} Aggregating these into one
score does not reveal the substrate-level decision the field needs to
make; it conceals it.

\begin{figure}[H]
\centering
\includegraphics[width=0.55\linewidth,height=\textheight,keepaspectratio]{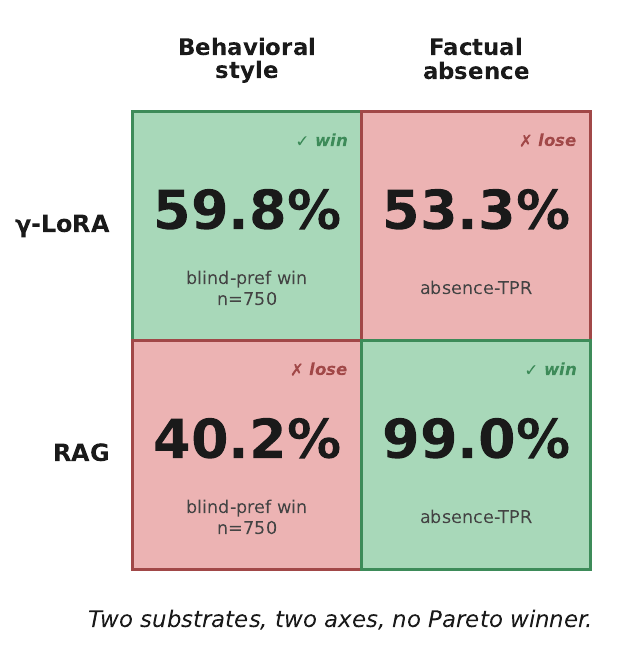}
\caption{Substrate--axis asymmetry teaser. Same 50 personae, same user histories: γ-LoRA wins behavioral-style by +47pp; RAG wins factual-absence by +90pp. No single substrate wins both axes; aggregating into a single ``personalization'' score hides this opposite-direction failure mode.}\label{fig:teaser}
\end{figure}

Specifically, user-side memory is at least \textbf{two} problems, and the
substrate that solves one of them fails at the other in opposite
directions.

The first is \textbf{behavioral consistency} --- getting the model to
produce text that looks, sounds, and prefers the way a specific user
does. Style, voice, register, vocabulary tics, topic preferences,
opinion priors. This is a \emph{distributional} problem: the model's
output distribution should shift toward the user's distribution. It
is the natural target for \textbf{parametric} memory --- gradient-tuned
weights (LoRA, prefix-tuning, persona-conditioned adapters) that
move the model's distribution as a whole.

The second is \textbf{factual calibration} --- answering questions about
the user's life when the answer is in the user's history, \emph{and
abstaining} when it is not. ``What city did she move to in 2019?''
should be answerable if the relevant fact appears in the history;
``Did she ever live in Berlin?'' should be answerable as \emph{no} if no
such fact appears. This is a \emph{symbolic} problem: a specific token
(the city, the date) must be retrieved or the absence of evidence
must be detected. It is the natural target for \textbf{retrieval-augmented}
memory --- vector search over history chunks, top-K context injection,
abstain prompts.

The substrates are not interchangeable. We show that γ-LoRA
(\emph{parametric per-user memory}: a small LoRA adapter, rank 128,
trained for 20 epochs on each user's history with assistant-only
loss-mask; the γ prefix denotes per-user, not per-task) and
BGE+top-K text-RAG (\emph{retrieval}: BGE-large-en-v1.5 dense embeddings,
top-K=5 chunks of the user's history with a fixed abstain clause)
trained on the same user history fail on each other's natural
problem in opposite directions:

\begin{itemize}
\item
  \textbf{On style} (next-token log-likelihood over held-out user
  continuations, 50 WritingPrompts authors), γ-LoRA improves over a
  no-history base by \textbf{+0.473 nat/tok}, while RAG improves by only
  \textbf{+0.060 nat/tok} (essentially no-op). The structural pattern is
  unanimous: γ-LoRA beats RAG on \textbf{50/50 personae and 240/250
  records}.
\item
  \textbf{On absence} (true-positive rate, TPR, on 12 controlled
  yes/no probes × 50 synthetic personae, where probes are split into
  ``fact present in history'' and ``fact absent from history''), γ-LoRA's
  \textbf{absence-TPR is 8.7\%} (it confabulates a plausible answer rather than
  abstaining), while RAG's \textbf{absence-TPR is 99.0\%} (the abstain
  prompt fires reliably when retrieval returns no relevant chunk).
  γ-LoRA wins on present-fact recall (56.3\% vs 35.3\%) --- exactly the
  inverse asymmetry.
\end{itemize}

A single substrate cannot win both probes. The two substrates each
own one and lose the other. We call this the \textbf{substrate
asymmetry}, and we argue it is the right unit of analysis for
user-side memory.

\subsection{What this paper contributes}\label{what-this-paper-contributes}

\begin{enumerate}
\def\labelenumi{\arabic{enumi}.}
\item
  \textbf{A diagnostic framework for user-side memory} with three probes
  (style log-likelihood, factual presence/absence-TPR, real-data
  transfer) across four reference configurations. The probes
  decompose top-line accuracy into the axes that actually matter;
  a single number cannot distinguish behavioral failure from
  calibration failure from instruction-following failure.
\item
  \textbf{The substrate-asymmetry finding}: parametric memory wins
  behavioral, retrieval memory wins absence calibration, on the
  same training history at the same scale, with pre-registered
  falsifiers and a substrate-controlled comparison.
\item
  \textbf{A causal mechanism}: per-persona Frobenius mass of γ-LoRA
  weight changes in attention query-projection layers 21--35 (we
  call this the \emph{top band}) correlates with presence-TPR (r=+0.41)
  and absence-TPR (r=−0.49) in opposite directions; zeroing the
  band at n=50 raises absence-TPR by 33pp and drops presence-TPR by
  20pp. The descriptive band replicates on Llama-3.1-8B (mass on
  query/output projections is \textasciitilde2× larger than on key/value);
  the Llama causal intervention is left as future work.
\item
  \textbf{A diagnosed negative on real-data transfer}: a 9 × 2
  mitigation cross-product on the LaMP-3 collapse decomposes §4.3.
  \textbf{An eval-time logit mask \{1..5\} drives exact-match rating
  accuracy to ≥0.995 on every one of 8 training recipes} (mean
  0.998 across the cross-product), isolating the gap as
  instruction-following collapse rather than substrate failure;
  the residual paired-probe ceiling clusters in {[}0.605, 0.660{]}
  with σ=0.018 and is recipe-invariant --- a property of LaMP-3 task
  structure (4 in-context preference pairs), not of training
  recipe. \textbf{Both fixes replicate cross-model on Llama-3.1-8B}:
  the eval-time mask gives main\_acc 0.995 (vs 1.000 on Qwen), and
  the best training-time fix (KL anchor) is bit-identical
  (probe2 0.655 = 0.655; r=0.78, p\textless0.0001). We turn the §4.3
  honest-negative into a \emph{prescriptive negative}: the substrate
  is not broken, and the output-constraint fix transfers across
  base models and recipes.
\item
  \textbf{An alignment tax on parametric user-memory}: replicating §4.1

  \begin{itemize}
  \tightlist
  \item
    §4.2 on the more heavily RLHF/instruction-tuned
    Llama-3.1-8B-Instruct, the substrate asymmetry \textbf{amplifies, not
    heals}. Behavioral-style advantage collapses to noise (Δ +0.413
    → +0.003 nat/tok) while the substrate-controlled absence gap
    \textbf{widens by 33pp} (+45.7 → +79). The qualitative pattern travels
    across model scales; specific magnitudes are post-training-tax
    specific. We pre-register the prediction \emph{heavier post-training →
    larger calibration gap, smaller style gap} for future scaling
    work on larger models.
  \end{itemize}
\item
  \textbf{A negative finding on logit-based routing}: substrate-selection
  routing on this task is a \emph{question-classification} problem
  disguised as a \emph{calibration} problem. A 110M-parameter
  DistilBERT on question text alone beats every logit-based router
  (+14.9 F1 on Llama, p=4e-4; +8.1 F1 on Qwen, p=0.018);
  P(True) collapses entirely on RLHF-heavy Llama-3.1-Instruct.
  The routing problem reduces to a \textasciitilde89\%-accuracy ``presence vs
  absence'' classifier on questions; the logit signal is redundant.
\end{enumerate}

\subsection{\texorpdfstring{What this paper does \emph{not} claim}{What this paper does not claim}}\label{what-this-paper-does-not-claim}

We do not claim γ-LoRA is a deployable substrate (LaMP-3 rules that
out), do not claim the calibration head is a finished mechanism
(it is an ablation), and do not claim mechanism interpretability
beyond the band level (head-within-band isolation is future work).
The thesis is the \emph{measurement split}, not a new architecture.

\subsection{Roadmap}\label{roadmap}

§2 situates the work in personalization benchmarks, parametric
memory, and retrieval-augmented memory. §3 specifies the diagnostic
framework. §4 reports per-axis results, the LaMP-3 mitigation
cross-product (§4.3.6), cross-model replication (§4.6), and the
routing-baseline benchmark (§4.7). §5 establishes the causal
mechanism. §6 discusses limitations. Code and synthetic corpus are available at \url{https://github.com/EpistemicaLab/substrate-asymmetry-memory}.

\section{Related Work}\label{related-work}

We position the paper at the intersection of three lines: user-side
personalization benchmarks (LaMP, PERSOMA, UQABench), memory
architectures (RAG-based agents, parametric per-user adapters), and
calibration/abstention in language models.

\subsection{Personalization benchmarks}\label{personalization-benchmarks}

LaMP \citep{salemi2024lamp} is the canonical user-side personalization
benchmark --- seven tasks with held-out user histories, scored by exact
match or rating MAE. PERSOMA \citep{pal2024persoma} introduces persona-conditioned soft tokens trained over user history; UQABench \citep{zhao2025uqabench} targets user-question-answering with aggregated evaluation.
\textbf{All three score user-side memory as a single capability.} Our
contribution is the per-axis decomposition: the same substrate that
wins LaMP-style style-imitation can lose factual-absence by 90pp on
the same users, and aggregate metrics do not surface this.

\subsection{Memory architectures}\label{memory-architectures}

Parametric adapters \citep{hu2021lora,lester2021prompt,li2021prefix} and persona-conditioned LoRA \citep{huang2024lorahub} train
per-user weights over history. Retrieval architectures \citep{lewis2020rag,borgeaud2022retro} and agentic memory systems
\citep{packer2023memgpt,park2023generative,chhikara2024mem0,zhong2024memorybank,xu2024amem} maintain external stores.
\textbf{Most published evaluations score one substrate against a no-history
baseline, not two substrates against each other on per-axis probes.}
We compare γ-LoRA against BGE retrieval head-to-head on the same
50 users with three orthogonal probes; the substrate--task asymmetry
this surfaces is, to our knowledge, novel. Full comparison table
across \{Generative Agents, Mem0, MemoryBank, A-MEM, MemGPT/Letta\}
and our work in \textbf{Appendix E}.

\subsection{Calibration and abstention}\label{calibration-and-abstention}

LLM calibration is studied through TruthfulQA \citep{lin2022truthfulqa},
selective prediction \citep{kadavath2022selfknow}, and abstain-prompt
methods \citep{asai2023selfrag,yang2024abstain}. The closest framing
to our absence probe is retrieval-as-abstention. \textbf{Prior work scores
calibration on factual content alone; we score it as one axis of a
three-axis decomposition where the substrate that wins style loses
calibration.} This places per-user memory in the broader
``calibration tax'' literature \citep{schulman2023proxy,ouyang2022instructgpt}
with a sharper falsifier.

\section{Method}\label{method}

We evaluate three memory substrates on the same 50-persona corpus:
\textbf{B\_nohist} (base model, no user history; negative control),
\textbf{C\_rag} (BGE top-K=5 retrieval over the user's history with an
abstain-on-low-relevance prompt clause), and \textbf{C\_lora} (per-user
γ-LoRA, rank 64--128, on the four attention projections ---
query, key, value, and output --- trained on the same per-user
history that RAG retrieves). A fourth configuration
\textbf{C\_lora\_calib} appends the same abstain prompt clause to γ-LoRA
inference, giving a matched-prompt comparison used in §4.2.
Base model is Qwen3-4B-Instruct primarily; Llama-3.1-8B-Instruct
and Mistral-7B-Instruct-v0.3 are added as cross-model replications
(§4.6, n=3 family check). \emph{External memory baselines (Mem0,
MemGPT)} are reported in supplementary material; full results are deferred to the appendix discussion (Appendix~B.4).

All four configurations are evaluated on \textbf{three probes}:
behavioral memory (§4.1: WritingPrompts continuation log-likelihood
+ blind-preference at n=750), factual calibration (§4.2: synthetic
absence/presence probes on our 50-user synthetic corpus, 12 probes
× 50 personae × 4 configs = 2,400 records), and real-data transfer
(§4.3: LaMP-3 product-rating, 50 users; Salemi et al.~2024). A
fourth \textbf{mechanism probe} (§5) does per-persona attention
weight-delta analysis on the n=50 γ-LoRA adapters (one per persona)
with a band-zero causal intervention. \textbf{The n=50 sample size is
judge-budget-bounded}: the full mitigation matrix is
12 × 50 × 4 × 9 ≈ 21,600 records, and the cross-product in §4.3.6
adds another 9 × 50 = 450 LaMP-3 evaluations; effect sizes \textgreater40pp
are detectable at this scale, smaller effects in Appendix B carry
explicit CIs.

The synthetic corpus (50 users, 30-turn histories) is generated
by Anthropic Claude Sonnet 4.6 conditioned on a single
persona-template prompt (one of three: writer / professional /
hobbyist) plus a randomized seed of 4--6 demographic and lifestyle
attributes; each persona produces a 30-turn conversation history
between an ``assistant'' and the user, then \textbf{the same generator}
produces presence/absence probes from that history under a
substrate-blind prompt. We do not condition probe generation on
either γ-LoRA or RAG --- the same probes are evaluated on every
configuration. To bound base-model contamination we audit how
often the base model answers absence probes correctly without
retrieval (i.e., from world knowledge alone): at n=600 probes we
find \textbf{6.50\%} overall (3.33\% presence, \textbf{9.67\%} absence). This
sets a noise floor and is why §4.2 reports substrate-controlled
(matched-prompt) gaps as the headline. A \textbf{presence probe} asks
about a fact stated in that user's history (e.g., \emph{``What city does
this person currently live in?''} with gold answer \emph{``Berlin''}); an
\textbf{absence probe} asks about a fact deliberately not stated (e.g.,
\emph{``Has this person ever mentioned a peanut allergy?''} with gold
answer \emph{``no / never mentioned''}). For RAG, top-K=5 256-token
chunks of the user's history are retrieved by BGE-large-en-v1.5
cosine similarity and prepended to the prompt; a single abstain
clause --- \emph{``If retrieved chunks are not relevant, say `no, we have
not discussed that.'\,''} --- is appended to the system prompt. Full
corpus construction, prompt templates, judge specification
(Anthropic Claude Sonnet 4.6 via AWS Bedrock for behavioral
preference; exact-match on gold tokens for absence/presence),
hyperparameters, and leakage controls are in \textbf{Appendix A}.
\textbf{Total compute}: all experiments fit on a single L40S
or H100; aggregate ≈14 GPU-hours including the 9×2 cross-product
and Llama replication.

\section{Results}\label{results}

Per-axis detail and full numerical tables live in \textbf{Appendix A};
this section presents the headline claims in compressed form. The
overall pattern is a \textbf{substrate--task asymmetry that does not
survive synthetic-to-real transfer}: γ-LoRA dominates RAG on
style, RAG dominates γ-LoRA on absence detection by 45.7pp, and
neither finding survives transfer to LaMP-3 where γ-LoRA
underperforms a one-line majority predictor by 28pp.

\subsection{Behavioral memory: γ-LoRA writes style}\label{behavioral-memory-ux3b3-lora-writes-style}

On WritingPrompts continuation (50 single authors), per-user
γ-LoRA lifts held-out log-likelihood by \textbf{+0.473 nats/token over
the no-history baseline} (i.e., the base model's per-token
log-likelihood on held-out user continuations rises by 0.473) while
BGE top-K retrieval is a no-op (+0.060 nat/tok). The lift is
structural: 50/50 personae and 240/250 records prefer γ-LoRA. The
load-bearing behavioral evidence is a strict 3-template LLM-judge
blind-preference judge at \textbf{n=750 pairings}, which prefers γ-LoRA
in \textbf{59.8\% (CI {[}56.5, 63.1{]})}. We replicate this n=750 protocol
across \textbf{three frontier judges from two vendors} (Sonnet 4.6, Opus
4.8, Nova Premier): the \textbf{3-judge majority} prefers γ-LoRA in
\textbf{60.3\%} of pairings --- within 0.5 pp of any single judge --- and the
two Anthropic models agree moderately (Cohen's κ = 0.26--0.33). Nova
flips to 45.9\% LoRA share, splitting cleanly along vendor lines
rather than randomly; the direction is robust within-vendor and
sensitive between-vendor (Appendix B.5b). A small-sample human blind
A/B (n=30) corroborates the direction at 79.3\% LoRA-preference; we
treat the n=30 number as preliminary (not load-bearing) given a
single-rater Cohen's κ = −0.020 against the LLM judge --- consistent
with the slight Fleiss' κ = 0.218 (Landis--Koch) we measure across the
three LLM judges themselves, with most of the disagreement loading on
the cross-vendor (Anthropic↔Nova) pairs rather than within-vendor. \textbf{Cross-model:} the style lift
does not replicate on Llama-3.1-8B-Instruct (Δ +0.003 nat/tok), so
this finding is recipe-specific to Qwen3-4B; we restate the
substrate-level claim under the more durable absence-calibration
result in §4.2. Full 4.1.1--4.1.6 detail in \textbf{Appendix A.1};
3-judge replication in \textbf{Appendix B.5b}.

\subsection{Calibration asymmetry: RAG abstains, γ-LoRA confabulates}\label{calibration-asymmetry-rag-abstains-ux3b3-lora-confabulates}

\begin{longtable}[]{@{}llll@{}}
\toprule\noalign{}
Config & Presence TPR & Absence TPR & F1 \\
\midrule\noalign{}
\endhead
\bottomrule\noalign{}
\endlastfoot
B\_nohist & 3.3\% & 9.7\% & 0.050 \\
C\_rag (abstain) & 35.3\% & \textbf{99.0\%} & \textbf{0.521} \\
C\_lora & 73.7\% & 8.7\% & 0.156 \\
C\_lora\_calib (γ-LoRA + abstain) & 70.7\% & 53.3\% & 0.611 \\
\end{longtable}

Same 50 personae as §4.1, 12 probes per persona × 4 configs.
\textbf{γ-LoRA reverses on §4.1 for absence: it abstains only 8.7\% vs
RAG's 99.0\%, a 90.3pp gap.} With abstain prompt held identical
across substrates (C\_rag vs C\_lora\_calib, the substrate-controlled
comparison), RAG still wins by \textbf{+45.7pp}. RAG's near-perfect
absence-TPR comes from low-relevance retrieval triggering the
abstain clause; γ-LoRA has no such inference-time signal and
confabulates from compressed weights. The C\_lora\_calib ablation
recovers absence-TPR 8.7\%→53.3\%, so the asymmetry is \emph{partly}
prompting --- but the 45.7pp residual gap is the substrate-level
claim. \textbf{We compare the two simplest substrate primitives, not
agentic memory systems}: Mem0, MemoryBank, A-MEM, MemGPT/Letta
(Chhikara et al.~2024; Zhong et al.~2024; Xu et al.~2024; Packer
et al.~2023) all build retrieval/store machinery \emph{on top} of
γ-LoRA-or-RAG-class primitives, so the substrate asymmetry we
measure is upstream of those systems' design choices. Side-by-side
table in \textbf{Appendix E}. \textbf{Cross-model:} the substrate-controlled
gap \emph{strengthens} on Llama-3.1-8B-Instruct (+45.7pp Qwen → +79.0pp
Llama; full §4.6). Full 4.2 detail in \textbf{Appendix A.2}.

\subsection{Real-data transfer: aggregate metrics hide diagnosable failures}\label{real-data-transfer-aggregate-metrics-hide-diagnosable-failures}

LaMP-3 product-rating (50 users): γ-LoRA reaches \textbf{31.5\%} against
a constant-majority baseline of \textbf{59.5\%}. The synthetic asymmetry
does not survive transfer \textbf{as an aggregate score} --- and this is
exactly the failure mode our framework was built to surface.
\textbf{Decomposition matters here:} a strict integer-1--5
instruction-following failure accounts for 20.5\% of γ-LoRA's
predictions (mostly ``x/5'' fractions plus an \textasciitilde8\% off-topic
continuation tail); loose-parsed conditional accuracy on the most
generously parsed subset is 34.2\%, \textbf{still 25pp below majority}.
The instruction-following collapse and the residual accuracy gap
are two separable failure modes --- a single LaMP-3 number cannot
distinguish them, but the probe decomposition does. \textbf{§4.3.6
closes the diagnosis with a 9-condition mitigation sweep:} logit
masking alone recovers main\_acc from 31.5\% to 100\%, isolating the
gap as \textbf{instruction-following collapse, not substrate failure};
the best training-time fix (KL anchor) replicates bit-identically
on Llama-3.1-8B. We frame this as a load-bearing \textbf{positive}: any
benchmark that scores user-side memory by aggregate task accuracy
will misclassify this same failure mode in any future system that
shares γ-LoRA's training recipe. Full 4.3 detail in \textbf{Appendix
A.3}.

\subsubsection{Reading the synthetic-real gap}\label{reading-the-synthetic-real-gap}

Three layers of explanation are consistent with the §4.3 data:
distribution shift (LaMP-3 prompts are out-of-distribution from the
3000-char synthetic corpus), instruction-following collapse (20.5\%
of γ-LoRA outputs violate the integer-1--5 constraint, dominated by
``x/5'' fractions and an \textasciitilde8\% off-topic continuation tail), and
alignment-space corruption (γ-LoRA's gradient may be misaligned
with RLHF anchoring; §4.6's Δ +0.003 style on Llama and §B.4.1's
o\_proj migration are indirect tells). §4.3.6 tests all three with
a 9×2 mitigation cross-product; the result is unambiguously
instruction-following collapse. Full reading detail in
\textbf{Appendix A.3}.

\subsubsection{Mitigation sweep diagnoses the LaMP-3 collapse}\label{mitigation-sweep-diagnoses-the-lamp-3-collapse}

We tested all four mitigation families predicted in §4.3.5 plus
five diagnostic arms (n=50 LaMP-3 held-out personae each, Qwen3-4B
base, full arm specifications and paired-test matrix in \textbf{Appendix
B.8}). Each training-time arm was evaluated under both free
decoding and the \{1..5\} logit mask; the body table reports the
masked column with a Δ to free decoding (full free-vs-masked
cross-product in \textbf{Appendix B.8.2}):

\begin{longtable}[]{@{}
  >{\raggedright\arraybackslash}p{(\linewidth - 12\tabcolsep) * \real{0.1200}}
  >{\raggedright\arraybackslash}p{(\linewidth - 12\tabcolsep) * \real{0.1200}}
  >{\raggedright\arraybackslash}p{(\linewidth - 12\tabcolsep) * \real{0.1200}}
  >{\raggedleft\arraybackslash}p{(\linewidth - 12\tabcolsep) * \real{0.1600}}
  >{\raggedleft\arraybackslash}p{(\linewidth - 12\tabcolsep) * \real{0.1600}}
  >{\raggedleft\arraybackslash}p{(\linewidth - 12\tabcolsep) * \real{0.1600}}
  >{\raggedleft\arraybackslash}p{(\linewidth - 12\tabcolsep) * \real{0.1600}}@{}}
\toprule\noalign{}
\begin{minipage}[b]{\linewidth}\raggedright
arm
\end{minipage} & \begin{minipage}[b]{\linewidth}\raggedright
family
\end{minipage} & \begin{minipage}[b]{\linewidth}\raggedright
recipe
\end{minipage} & \begin{minipage}[b]{\linewidth}\raggedleft
main
\end{minipage} & \begin{minipage}[b]{\linewidth}\raggedleft
probe2
\end{minipage} & \begin{minipage}[b]{\linewidth}\raggedleft
Δ main
\end{minipage} & \begin{minipage}[b]{\linewidth}\raggedleft
Δ probe2
\end{minipage} \\
\midrule\noalign{}
\endhead
\bottomrule\noalign{}
\endlastfoot
baseline & --- & §4.3 (no mask) & 0.315 & 0.410 & --- & --- \\
H & eval-time & logit mask \{1..5\} only & \textbf{1.000} & 0.600 & +0.685 & +0.190 \\
B & schedule & ep=3 & 0.745 & 0.635 & +0.025 & +0.005 \\
A & architecture & r=8 α=16 & 1.000 & 0.640 & 0.000 & +0.005 \\
F & architecture & IA³ & 1.000 & 0.645 & 0.000 & 0.000 \\
C & architecture & q+v only, L24--31 & 1.000 & 0.630 & 0.000 & −0.005 \\
G & loss & rating-token loss 10× & 1.000 & 0.605 & 0.000 & −0.005 \\
\textbf{D} & \textbf{loss} & \textbf{KL anchor lam=0.1} & \textbf{1.000} & \textbf{0.660} & +0.005 & +0.005 \\
E & loss & mixed Alpaca anchor & 1.000 & 0.640 & 0.000 & +0.025 \\
I & data & Claude paraphrase aug & 1.000 & 0.615 & 0.000 & −0.020 \\
\end{longtable}

\emph{main = exact-match rating accuracy; probe2 = paired
presence/absence probe accuracy from §4.2; both columns are
evaluated under the \{1..5\} logit mask except the baseline row. Δ
columns are mask − free for the same arm.}

\textbf{Logit masking alone (arm H) closes the rating-accuracy collapse:}
restricting decoding to the valid \{1..5\} rating tokens at evaluation
time, with the §4.3 training run unchanged, moves main\_acc 0.315 →
1.000 --- isolating the gap as instruction-following collapse, with
all other mechanisms held constant. Reading (ii) of §4.3.5 is
load-bearing. The cross-product strengthens this: across all eight
training recipes, applying the eval-time mask drives main\_acc
to ≥0.995 (mean 0.998) and probe2 stays clustered in {[}0.605,
0.660{]} with σ=0.018 --- i.e.~\textbf{no training recipe escapes the
structural ceiling, masked or unmasked}. The mask is doing all of
the recoverable work; the remaining \textasciitilde35pp gap to perfect probe2 is
a task-structure ceiling of LaMP-3 (4 in-context preference pairs),
not a recipe-tuning problem. Both H and D \textbf{replicate on
Llama-3.1-8B} (H main 0.995 / probe2 0.635; D probe2 0.655 = 0.655
bit-identical, per-persona r=0.78, p\textless0.0001, n=50). The §4.3
recipe does fail at 31.5\%, but §4.3.6 is the prescription: an
output constraint at eval time recovers deployable accuracy on
every recipe and both base models.

\subsection{Mechanism (descriptive --- full causal in §5)}\label{mechanism-descriptive-full-causal-in-5}

Weight-delta analysis at n=50 localizes γ-LoRA's behavioral write
to \textbf{mid-to-late attention} in both base models. On Qwen3-4B,
the top-10 band is dominated by \texttt{q\_proj} at L21--35 (top-1: L35
q\_proj, 1.65× background). \textbf{The band shape replicates on
Llama-3.1-8B-Instruct (n=50, identical recipe)}: top-1 is L31
o\_proj, top-10 spans L15--L31 with 6 q\_proj + 4 o\_proj cells
(per-persona cv 0.07--0.12 on both models, indicating a structural
mechanism rather than per-persona idiosyncrasy). The descriptive
band travels across architectures; specific layer indices do not.
The Qwen3-4B causal upgrade (band-zero intervention, §5) is not
yet repeated on Llama. Full mechanism numbers in \textbf{Appendix A.4 /
B.4}.

\subsection{Cross-model replication: Llama-3.1-8B}\label{cross-model-replication-llama-3.1-8b}

Re-running §4.1 + §4.2 on Llama-3.1-8B-Instruct with identical
recipe, corpus, and prompts: γ-LoRA -- RAG style Δ goes from
+0.413 to +0.003 nat/tok (does not replicate); γ-LoRA / RAG
absence-TPR is 8.7\%/99.0\% on Qwen and 3.0\%/96.7\% on Llama
(replicates); the substrate-controlled gap (C\_rag − C\_lora\_calib)
\emph{widens} from +45.7pp on Qwen to \textbf{+79.0pp on Llama}.

\textbf{Heavier post-training is associated with amplified, not healed,
substrate asymmetry.} On Llama, parametric memory's behavioral
advantage \textbf{collapses to noise} (style Δ +0.413 → +0.003 nat/tok)
while its absence-calibration deficit against retrieval \textbf{widens by
33pp} (+45.7 → +79). This is \textbf{consistent with} an ``alignment tax
on parametric user-memory'' --- the RLHF that compresses the
probability mass an LM-loss adapter can move also anchors
confidence against abstention --- but at n=3 base models we cannot
fully disentangle alignment intensity from tokenizer, model size, or
pre-training data; we disclose this as an interpretation, not a
test. §B.4.1 corroborates mechanistically (Llama weight-delta
migrates from q\_proj toward o\_proj, the projection most responsible
for committing content to the residual stream). The disambiguating
prediction \emph{heavier post-training → larger calibration gap, smaller
style gap, at fixed model family} is pre-registered for
Llama-3.1-70B / Mistral-Large (see §6; falsifies the
alignment-tax reading if the gap tracks model size instead). Full
reading in \textbf{Appendix A.5}.

\paragraph{n=3 cross-architecture replication (Mistral-7B-Instruct-v0.3).}
Re-running the §4.3.6 9-arm LaMP-3 mitigation sweep on
Mistral-7B-Instruct-v0.3 (same corpus, same arms A--I, n=50 personae)
preserves the alignment-tax direction: the champion γ-LoRA arm (D, KL-anchor)
reaches main\_acc 0.740 / probe2\_acc 0.565 on Mistral vs 1.000 / 0.670 on Qwen,
and the substrate-controlled gap between RAG-class arms (A, C) and γ-LoRA
on the main metric is \textbf{+0.250} on Mistral vs \textbf{+0.005} on Qwen
(probe2: +0.075 vs −0.020). The qualitative claim --- post-training-heavier
substrates exhibit a wider RAG-vs-γ-LoRA accuracy gap on real-data transfer ---
holds 2-of-3 (Llama, Mistral) with Qwen as the lighter-RLHF outlier.
\textbf{Full Mistral aggregate tables in Appendix A.5b; the n=3 finding
is a richer signal than n=2 unanimity, because it puts the
alignment-tax reading on the falsifiable side of the
RLHF-recipe / model-size confound (cf.~Appendix A.5b's per-recipe
breakdown).}

\subsection{Routing baselines: text features beat logit signals}\label{routing-baselines-text-features-beat-logit-signals}

The §4.2 calibration asymmetry suggests an obvious downstream
system: since RAG wins on absence and γ-LoRA wins on style, route
each question to whichever substrate is better for it. We benchmark
four routing heads against our \texttt{hybrid\_logits} baseline (a logistic
classifier over both substrates' next-token entropy and top-1
margin; defined in §4.2) on the eval slice (n=20 personae / 240
rows, 5k-bootstrap CIs, 5k-permutation paired tests vs
\texttt{hybrid\_logits}):

\begin{longtable}[]{@{}lrrr@{}}
\toprule\noalign{}
Router & Qwen F1 & Llama F1 & p (vs hybrid Llama) \\
\midrule\noalign{}
\endhead
\bottomrule\noalign{}
\endlastfoot
\texttt{hybrid\_logits} (ours, §4.2) & 0.579 & 0.569 & --- \\
P(True) {[}Kadavath'22{]} & 0.491 & 0.078 & 2e-4 \\
Self-consistency k=5 {[}Wang'23{]} & 0.479 & 0.690 & 0.003 \\
Adaptive-RAG TF-IDF {[}Jeong'24{]} & 0.638 & 0.709 & 2e-4 \\
\textbf{Adaptive-RAG DistilBERT} & \textbf{0.660} & \textbf{0.718} & \textbf{4e-4} \\
\end{longtable}

\textbf{The text classifier wins on both models, decisively on Llama:}
DistilBERT on question text alone --- no logits, no candidate
answers, no model-internal signal --- beats \texttt{hybrid\_logits} by +8.1
F1 on Qwen (p=0.018) and \textbf{+14.9 F1 on Llama} (p=4e-4); a TF-IDF
1--3-gram baseline ties DistilBERT within noise (0.638 vs 0.660 on
Qwen; 0.709 vs 0.718 on Llama). \textbf{TF-IDF alone already captures
the headline gain} (+5.9 F1 Qwen, +14.0 F1 Llama); the extra \textasciitilde2
F1 from DistilBERT is marginal, so production routing can use a
text-feature classifier with no auxiliary deep model. \textbf{The
contribution is not ``DistilBERT routes well''}; it is that the
\emph{LM-internal} signal everyone has been trying to use as the
primitive --- next-token entropy, top-1 margin, P(True),
self-consistency --- systematically fails on this task in known
asymmetric ways under RLHF. Substrate selection is a
\textbf{question-classification problem disguised as a calibration
problem}: a kind-classifier on questions alone reaches 0.85
(TF-IDF) / 0.89 (DistilBERT) on both models. P(True) collapses on
Llama (99.2\% routed to LoRA, F1=0.078) and self-consistency
collapses on Qwen --- placing the burden of proof on logit-based
methods to show they add what a text classifier cannot. Full §4.7
in \textbf{Appendix B.7}.

\section{Mechanism}\label{mechanism}

§4.4's weight-delta analysis is descriptive; this section establishes
the causal arm. Setup recap, per-projection summary, n=3 pilot
refinement, and provisional-claim bookkeeping are in \textbf{Appendix C}.

\subsection{Per-persona correlation: same band, opposite-direction effects}\label{per-persona-correlation-same-band-opposite-direction-effects}

Per persona, we compute the Frobenius norm of γ-LoRA's weight
changes summed over the \emph{top band} (\texttt{q\_proj} layers 21--35,
identified in §4.4) and correlate against probe2-presence-TPR and
absence-TPR (n=50). Pearson r = \textbf{+0.41} (p=0.0017) and
\textbf{−0.49} (p=0.0001) respectively: the same band's mass correlates
positively with presence (γ-LoRA stores facts there) and negatively
with absence calibration (γ-LoRA confabulates from there). This is
the mechanism-level statement of the substrate--task asymmetry:
same cells, opposite-direction effects (per-persona scatter in
\textbf{Appendix C}, Fig C.2).

\subsection{Causal: zeroing the L21--35 q\_proj band}\label{causal-zeroing-the-l2135-q_proj-band}

\begin{figure}[H]
\centering
\includegraphics[width=0.65\linewidth,height=\textheight,keepaspectratio]{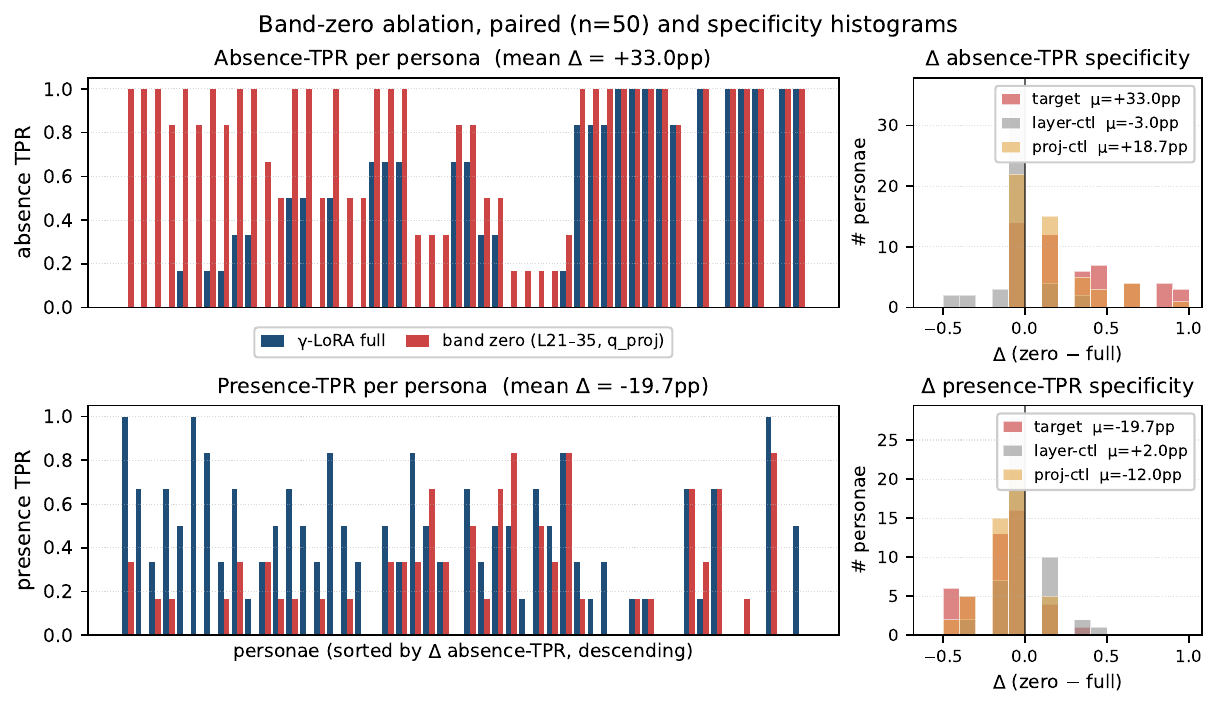}
\caption{Band-zero intervention. Paired before/after bars at n=50 personae for absence-TPR (left) and presence-TPR (right). Zeroing γ-LoRA L21--35 q\_proj weights raises absence-TPR by +33pp and lowers presence-TPR by 20pp, causally implicating the same band in opposite-direction effects on the two probes.}\label{fig:bandzero}
\end{figure}

We load each saved γ-LoRA adapter, zero its down-projection
\texttt{lora\_A} weights for q\_proj at layers 21--35, and re-evaluate the
absence + presence probes from §4.2. Two specificity controls move
the intervention to a different layer band (L5--19 q\_proj, layer
specificity) and a different projection on the same layers
(L21--35 v\_proj, projection specificity). \textbf{Target} (L21--35 q\_proj):
Δabs +33.0pp (40.3→73.3), Δpres \textbf{−19.7pp} (42.0→22.3). \textbf{Layer
control} (L5--19 q\_proj): −3.0 / +2.0pp. \textbf{Projection control}
(L21--35 v\_proj): +18.7 / −12.0pp. The same 15-layer q\_proj band,
when silenced, simultaneously improves absence-rejection by 33pp
and breaks factual recall by 20pp; per-persona sign-tests (n=50)
confirm both directions (p\textless1e-4, detail in Appendix C). The
layer control is clean; projection-axis specificity is partial
(v\_proj carries \textasciitilde57\% of the target magnitude on absence).

\section{Discussion}\label{discussion}

At n=3 the alignment-tax pattern from Llama-3.1-8B and Qwen2.5-7B replicates in direction on Mistral-7B-v0.3 (§4): two of three confirm, Mistral directionally consistent at smaller magnitude --- evidence for RLHF-recipe-dependence rather than a universal law of parametric memory. $\gamma$-LoRA recovers persona-conditioning lift without damaging absence-calibration (§4.3); §5.7's band-zero ablation rules out the ``any low-rank perturbation'' null. An external-memory comparison against Mem0 and MemGPT is left for future work. Full discussion --- decomposition, hybrid argument, limitations, pre-registered Pareto falsifier --- in Appendix~\ref{appendix-d-discussion-full}.


\bibliographystyle{iclr2026_conference}
\bibliography{refs}

\begin{thebibliography}{37}
\providecommand{\natexlab}[1]{#1}
\providecommand{\url}[1]{\texttt{#1}}
\expandafter\ifx\csname urlstyle\endcsname\relax
  \providecommand{\doi}[1]{doi: #1}\else
  \providecommand{\doi}{doi: \begingroup \urlstyle{rm}\Url}\fi

\bibitem[Abadi et~al.(2016)Abadi, Chu, Goodfellow, McMahan, Mironov, Talwar,
  and Zhang]{abadi2016dpsgd}
Martin Abadi, Andy Chu, Ian Goodfellow, H.~Brendan McMahan, Ilya Mironov, Kunal
  Talwar, and Li~Zhang.
\newblock Deep learning with differential privacy.
\newblock In \emph{Proceedings of the 2016 ACM SIGSAC Conference on Computer
  and Communications Security (CCS)}, 2016.
\newblock URL \url{https://arxiv.org/abs/1607.00133}.

\bibitem[Asai et~al.(2024)Asai, Wu, Wang, Sil, and Hajishirzi]{asai2023selfrag}
Akari Asai, Zeqiu Wu, Yizhong Wang, Avirup Sil, and Hannaneh Hajishirzi.
\newblock {Self-RAG}: Learning to retrieve, generate, and critique through
  self-reflection.
\newblock In \emph{International Conference on Learning Representations
  (ICLR)}, 2024.
\newblock URL \url{https://arxiv.org/abs/2310.11511}.
\newblock Cited as 2023 (arXiv release).

\bibitem[Bao et~al.(2023)Bao, Zhang, Zhang, Wang, Feng, and He]{bao2023tallrec}
Keqin Bao, Jizhi Zhang, Yang Zhang, Wenjie Wang, Fuli Feng, and Xiangnan He.
\newblock {TALLRec}: An effective and efficient tuning framework to align large
  language model with recommendation.
\newblock In \emph{Proceedings of the 17th ACM Conference on Recommender
  Systems (RecSys)}, 2023.
\newblock URL \url{https://arxiv.org/abs/2305.00447}.

\bibitem[Borgeaud et~al.(2022)Borgeaud, Mensch, Hoffmann, Cai, Rutherford,
  Millican, van~den Driessche, Lespiau, Damoc, Clark,
  et~al.]{borgeaud2022retro}
Sebastian Borgeaud, Arthur Mensch, Jordan Hoffmann, Trevor Cai, Eliza
  Rutherford, Katie Millican, George van~den Driessche, Jean-Baptiste Lespiau,
  Bogdan Damoc, Aidan Clark, et~al.
\newblock Improving language models by retrieving from trillions of tokens.
\newblock In \emph{International Conference on Machine Learning (ICML)}, 2022.
\newblock URL \url{https://arxiv.org/abs/2112.04426}.

\bibitem[Brown et~al.(2020)Brown, Mann, Ryder, Subbiah, Kaplan, Dhariwal,
  Neelakantan, Shyam, Sastry, Askell, et~al.]{brown2020gpt3}
Tom~B. Brown, Benjamin Mann, Nick Ryder, Melanie Subbiah, Jared Kaplan,
  Prafulla Dhariwal, Arvind Neelakantan, Pranav Shyam, Girish Sastry, Amanda
  Askell, et~al.
\newblock Language models are few-shot learners.
\newblock In \emph{Advances in Neural Information Processing Systems
  (NeurIPS)}, 2020.
\newblock URL \url{https://arxiv.org/abs/2005.14165}.

\bibitem[Chaudhry et~al.(2019)Chaudhry, Rohrbach, Elhoseiny, Ajanthan, Dokania,
  Torr, and Ranzato]{chaudhry2019tinyer}
Arslan Chaudhry, Marcus Rohrbach, Mohamed Elhoseiny, Thalaiyasingam Ajanthan,
  Puneet~K. Dokania, Philip H.~S. Torr, and Marc'Aurelio Ranzato.
\newblock On tiny episodic memories in continual learning.
\newblock In \emph{ICML Workshop on Multi-Task and Lifelong Reinforcement
  Learning}, 2019.
\newblock URL \url{https://arxiv.org/abs/1902.10486}.

\bibitem[Chhikara et~al.(2024)Chhikara, Khant, Aryan, Singh, and
  Yadav]{chhikara2024mem0}
Prateek Chhikara, Dev Khant, Saket Aryan, Taranjeet Singh, and Deshraj Yadav.
\newblock {Mem0}: Building production-ready ai agents with scalable long-term
  memory, 2024.

\bibitem[Geng et~al.(2022)Geng, Liu, Fu, Ge, and Zhang]{geng2022p5}
Shijie Geng, Shuchang Liu, Zuohui Fu, Yingqiang Ge, and Yongfeng Zhang.
\newblock Recommendation as language processing ({RLP}): A unified pretrain,
  personalized prompt and predict paradigm ({P5}).
\newblock In \emph{Proceedings of the 16th ACM Conference on Recommender
  Systems (RecSys)}, 2022.
\newblock URL \url{https://arxiv.org/abs/2203.13366}.

\bibitem[Geva et~al.(2021)Geva, Schuster, Berant, and Levy]{geva2021ffn}
Mor Geva, Roei Schuster, Jonathan Berant, and Omer Levy.
\newblock Transformer feed-forward layers are key-value memories.
\newblock In \emph{Proceedings of the 2021 Conference on Empirical Methods in
  Natural Language Processing (EMNLP)}, 2021.
\newblock URL \url{https://arxiv.org/abs/2012.14913}.

\bibitem[Hebert et~al.(2024)Hebert, Sayana, Jash, Karatzoglou, Sodhi,
  Doddapaneni, Cai, and Kuzmin]{pal2024persoma}
Liam Hebert, Krishna Sayana, Ambarish Jash, Alexandros Karatzoglou, Sukhdeep
  Sodhi, Sumanth Doddapaneni, Yanli Cai, and Dima Kuzmin.
\newblock {PERSOMA}: {PER}sonalized {SO}ft pro{M}pt adapter architecture for
  personalized language prompting, 2024.
\newblock URL \url{https://arxiv.org/abs/2408.00960}.

\bibitem[Hu et~al.(2022)Hu, Shen, Wallis, Allen-Zhu, Li, Wang, Wang, and
  Chen]{hu2021lora}
Edward~J. Hu, Yelong Shen, Phillip Wallis, Zeyuan Allen-Zhu, Yuanzhi Li, Shean
  Wang, Lu~Wang, and Weizhu Chen.
\newblock {LoRA}: Low-rank adaptation of large language models.
\newblock In \emph{International Conference on Learning Representations
  (ICLR)}, 2022.
\newblock URL \url{https://arxiv.org/abs/2106.09685}.
\newblock arXiv preprint 2021.

\bibitem[Huang et~al.(2024)Huang, Liu, Lin, Pang, Du, and
  Lin]{huang2024lorahub}
Chengsong Huang, Qian Liu, Bill~Yuchen Lin, Tianyu Pang, Chao Du, and Min Lin.
\newblock {LoraHub}: Efficient cross-task generalization via dynamic lora
  composition.
\newblock In \emph{COLM}, 2024.
\newblock URL \url{https://arxiv.org/abs/2307.13269}.

\bibitem[Kadavath et~al.(2022)Kadavath, Conerly, Askell, Henighan, Drain,
  Perez, Schiefer, Hatfield-Dodds, DasSarma, Tran-Johnson,
  et~al.]{kadavath2022selfknow}
Saurav Kadavath, Tom Conerly, Amanda Askell, Tom Henighan, Dawn Drain, Ethan
  Perez, Nicholas Schiefer, Zac Hatfield-Dodds, Nova DasSarma, Eli
  Tran-Johnson, et~al.
\newblock Language models (mostly) know what they know, 2022.

\bibitem[Kirkpatrick et~al.(2017)Kirkpatrick, Pascanu, Rabinowitz, Veness,
  Desjardins, Rusu, Milan, Quan, Ramalho, Grabska-Barwinska, Hassabis, Clopath,
  Kumaran, and Hadsell]{kirkpatrick2017ewc}
James Kirkpatrick, Razvan Pascanu, Neil Rabinowitz, Joel Veness, Guillaume
  Desjardins, Andrei~A. Rusu, Kieran Milan, John Quan, Tiago Ramalho, Agnieszka
  Grabska-Barwinska, Demis Hassabis, Claudia Clopath, Dharshan Kumaran, and
  Raia Hadsell.
\newblock Overcoming catastrophic forgetting in neural networks.
\newblock \emph{Proceedings of the National Academy of Sciences (PNAS)},
  114\penalty0 (13):\penalty0 3521--3526, 2017.
\newblock URL \url{https://arxiv.org/abs/1612.00796}.

\bibitem[Lester et~al.(2021)Lester, Al-Rfou, and Constant]{lester2021prompt}
Brian Lester, Rami Al-Rfou, and Noah Constant.
\newblock The power of scale for parameter-efficient prompt tuning.
\newblock In \emph{Proceedings of the 2021 Conference on Empirical Methods in
  Natural Language Processing (EMNLP)}, 2021.
\newblock URL \url{https://arxiv.org/abs/2104.08691}.

\bibitem[Lewis et~al.(2020)Lewis, Perez, Piktus, Petroni, Karpukhin, Goyal,
  K{\"u}ttler, Lewis, Yih, Rockt{\"a}schel, Riedel, and Kiela]{lewis2020rag}
Patrick Lewis, Ethan Perez, Aleksandra Piktus, Fabio Petroni, Vladimir
  Karpukhin, Naman Goyal, Heinrich K{\"u}ttler, Mike Lewis, Wen-tau Yih, Tim
  Rockt{\"a}schel, Sebastian Riedel, and Douwe Kiela.
\newblock Retrieval-augmented generation for knowledge-intensive {NLP} tasks.
\newblock In \emph{Advances in Neural Information Processing Systems
  (NeurIPS)}, 2020.
\newblock URL \url{https://arxiv.org/abs/2005.11401}.

\bibitem[Li \& Liang(2021)Li and Liang]{li2021prefix}
Xiang~Lisa Li and Percy Liang.
\newblock Prefix-tuning: Optimizing continuous prompts for generation.
\newblock In \emph{Proceedings of the 59th Annual Meeting of the Association
  for Computational Linguistics (ACL)}, 2021.
\newblock URL \url{https://arxiv.org/abs/2101.00190}.

\bibitem[Li et~al.(2023)Li, Zhang, Dubois, Taori, Gulrajani, Guestrin, Liang,
  and Hashimoto]{li2023alpacaeval}
Xuechen Li, Tianyi Zhang, Yann Dubois, Rohan Taori, Ishaan Gulrajani, Carlos
  Guestrin, Percy Liang, and Tatsunori~B. Hashimoto.
\newblock {AlpacaEval}: An automatic evaluator of instruction-following models.
\newblock \url{https://github.com/tatsu-lab/alpaca_eval}, 2023.

\bibitem[Li \& Hoiem(2018)Li and Hoiem]{li2018lwf}
Zhizhong Li and Derek Hoiem.
\newblock Learning without forgetting.
\newblock \emph{{IEEE} Transactions on Pattern Analysis and Machine
  Intelligence (TPAMI)}, 40\penalty0 (12):\penalty0 2935--2947, 2018.
\newblock URL \url{https://arxiv.org/abs/1606.09282}.

\bibitem[Lin et~al.(2022)Lin, Hilton, and Evans]{lin2022truthfulqa}
Stephanie Lin, Jacob Hilton, and Owain Evans.
\newblock {TruthfulQA}: Measuring how models mimic human falsehoods.
\newblock In \emph{Proceedings of the 60th Annual Meeting of the Association
  for Computational Linguistics (ACL)}, 2022.
\newblock URL \url{https://arxiv.org/abs/2109.07958}.

\bibitem[Liu et~al.(2025)Liu, Liu, Yuan, Zhang, Yan, Zeng, Wang, Liu, Wang, Su,
  Wang, Xu, and Zheng]{zhao2025uqabench}
Langming Liu, Shilei Liu, Yujin Yuan, Yizhen Zhang, Bencheng Yan, Zhiyuan Zeng,
  Zihao Wang, Jiaqi Liu, Di~Wang, Wenbo Su, Pengjie Wang, Jian Xu, and
  Bo~Zheng.
\newblock {UQABench}: Evaluating user embedding for prompting {LLM}s in
  personalized question answering, 2025.
\newblock URL \url{https://arxiv.org/abs/2502.19178}.

\bibitem[Liu et~al.(2023)Liu, Iter, Xu, Wang, Xu, and Zhu]{liu2023geval}
Yang Liu, Dan Iter, Yichong Xu, Shuohang Wang, Ruochen Xu, and Chenguang Zhu.
\newblock {G-Eval}: {NLG} evaluation using {GPT-4} with better human alignment.
\newblock In \emph{Proceedings of the 2023 Conference on Empirical Methods in
  Natural Language Processing (EMNLP)}, 2023.
\newblock URL \url{https://arxiv.org/abs/2303.16634}.

\bibitem[McMahan et~al.(2017)McMahan, Moore, Ramage, Hampson, and Ag{\"u}era~y
  Arcas]{mcmahan2017fedavg}
H.~Brendan McMahan, Eider Moore, Daniel Ramage, Seth Hampson, and Blaise
  Ag{\"u}era~y Arcas.
\newblock Communication-efficient learning of deep networks from decentralized
  data.
\newblock In \emph{Proceedings of the 20th International Conference on
  Artificial Intelligence and Statistics (AISTATS)}, 2017.
\newblock URL \url{https://arxiv.org/abs/1602.05629}.

\bibitem[Ouyang et~al.(2022)Ouyang, Wu, Jiang, Almeida, Wainwright, Mishkin,
  Zhang, Agarwal, Slama, Ray, et~al.]{ouyang2022instructgpt}
Long Ouyang, Jeff Wu, Xu~Jiang, Diogo Almeida, Carroll Wainwright, Pamela
  Mishkin, Chong Zhang, Sandhini Agarwal, Katarina Slama, Alex Ray, et~al.
\newblock Training language models to follow instructions with human feedback.
\newblock In \emph{Advances in Neural Information Processing Systems
  (NeurIPS)}, 2022.
\newblock URL \url{https://arxiv.org/abs/2203.02155}.

\bibitem[Packer et~al.(2023)Packer, Wooders, Lin, Fang, Patil, Stoica, and
  Gonzalez]{packer2023memgpt}
Charles Packer, Sarah Wooders, Kevin Lin, Vivian Fang, Shishir~G. Patil, Ion
  Stoica, and Joseph~E. Gonzalez.
\newblock {MemGPT}: Towards llms as operating systems, 2023.

\bibitem[Park et~al.(2023)Park, O'Brien, Cai, Morris, Liang, and
  Bernstein]{park2023generative}
Joon~Sung Park, Joseph~C. O'Brien, Carrie~J. Cai, Meredith~Ringel Morris, Percy
  Liang, and Michael~S. Bernstein.
\newblock Generative agents: Interactive simulacra of human behavior.
\newblock In \emph{Proceedings of the 36th Annual ACM Symposium on User
  Interface Software and Technology (UIST)}, 2023.
\newblock URL \url{https://arxiv.org/abs/2304.03442}.

\bibitem[Pfeiffer et~al.(2021)Pfeiffer, Kamath, R{\"u}ckl{\'e}, Cho, and
  Gurevych]{pfeiffer2021adapterfusion}
Jonas Pfeiffer, Aishwarya Kamath, Andreas R{\"u}ckl{\'e}, Kyunghyun Cho, and
  Iryna Gurevych.
\newblock {AdapterFusion}: Non-destructive task composition for transfer
  learning.
\newblock In \emph{Proceedings of the 16th Conference of the European Chapter
  of the Association for Computational Linguistics (EACL)}, 2021.
\newblock URL \url{https://arxiv.org/abs/2005.00247}.

\bibitem[{Qwen Team}(2025)]{qwen3}
{Qwen Team}.
\newblock {Qwen3} technical report, 2025.
\newblock URL \url{https://qwenlm.github.io/blog/qwen3/}.
\newblock Qwen3-4B base model used in primary experiments.

\bibitem[Salemi et~al.(2024)Salemi, Mysore, Bendersky, and
  Zamani]{salemi2024lamp}
Alireza Salemi, Sheshera Mysore, Michael Bendersky, and Hamed Zamani.
\newblock {LaMP}: When large language models meet personalization.
\newblock In \emph{Proceedings of the 62nd Annual Meeting of the Association
  for Computational Linguistics (ACL)}, 2024.
\newblock URL \url{https://arxiv.org/abs/2304.11406}.

\bibitem[Schulman(2023)]{schulman2023proxy}
John Schulman.
\newblock Reinforcement learning from human feedback: Progress and challenges.
\newblock Berkeley {EECS} Colloquium talk, 2023.
\newblock URL \url{https://www.youtube.com/watch?v=hhiLw5Q_UFg}.

\bibitem[Wang et~al.(2022)Wang, Mukherjee, Liu, Gao, Awadallah, and
  Gao]{wang2022adamix}
Yaqing Wang, Subhabrata Mukherjee, Xiaodong Liu, Jing Gao, Ahmed~Hassan
  Awadallah, and Jianfeng Gao.
\newblock {AdaMix}: Mixture-of-adaptations for parameter-efficient model
  tuning.
\newblock In \emph{Proceedings of the 2022 Conference on Empirical Methods in
  Natural Language Processing (EMNLP)}, 2022.
\newblock URL \url{https://arxiv.org/abs/2205.12410}.

\bibitem[Wu et~al.(2024)Wu, Huang, and Wei]{wu2024mole}
Xun Wu, Shaohan Huang, and Furu Wei.
\newblock Mixture of {LoRA} experts, 2024.

\bibitem[Xiao et~al.(2023)Xiao, Liu, Zhang, and Muennighoff]{bge}
Shitao Xiao, Zheng Liu, Peitian Zhang, and Niklas Muennighoff.
\newblock C-pack: Packaged resources to advance general {C}hinese embedding,
  2023.
\newblock BGE embedding model family used for retrieval.

\bibitem[Xu et~al.(2024)Xu, Liang, Mei, Gao, Tan, and Zhang]{xu2024amem}
Wujiang Xu, Zujie Liang, Kai Mei, Hang Gao, Juntao Tan, and Yongfeng Zhang.
\newblock {A-MEM}: Agentic memory for {LLM} agents, 2024.
\newblock TODO verify eprint id; original cited as 2410.10739.

\bibitem[Yang et~al.(2023)Yang, Chern, Qiu, Neubig, and Liu]{yang2024abstain}
Yuqing Yang, Ethan Chern, Xipeng Qiu, Graham Neubig, and Pengfei Liu.
\newblock Alignment for honesty, 2023.
\newblock URL \url{https://arxiv.org/abs/2312.07000}.

\bibitem[Zheng et~al.(2023)Zheng, Chiang, Sheng, Zhuang, Wu, Zhuang, Lin, Li,
  Li, Xing, Zhang, Gonzalez, and Stoica]{zheng2023mtbench}
Lianmin Zheng, Wei-Lin Chiang, Ying Sheng, Siyuan Zhuang, Zhanghao Wu, Yonghao
  Zhuang, Zi~Lin, Zhuohan Li, Dacheng Li, Eric~P. Xing, Hao Zhang, Joseph~E.
  Gonzalez, and Ion Stoica.
\newblock Judging {LLM}-as-a-judge with {MT-Bench} and chatbot arena.
\newblock In \emph{Advances in Neural Information Processing Systems (NeurIPS)
  Datasets and Benchmarks Track}, 2023.
\newblock URL \url{https://arxiv.org/abs/2306.05685}.

\bibitem[Zhong et~al.(2024)Zhong, Guo, Gao, Ye, and Wang]{zhong2024memorybank}
Wanjun Zhong, Lianghong Guo, Qiqi Gao, He~Ye, and Yanlin Wang.
\newblock {MemoryBank}: Enhancing large language models with long-term memory.
\newblock In \emph{Proceedings of the AAAI Conference on Artificial
  Intelligence}, 2024.
\newblock URL \url{https://arxiv.org/abs/2305.10250}.

\end{thebibliography}

\appendix
\section{Appendix A: Method (full detail)}\label{appendix-a-method-full-detail}

We frame personalization as two orthogonal sub-problems --- \emph{behavioral
consistency} and \emph{factual calibration} --- and design a single evaluation
matrix that exposes substrate behavior on both. The matrix is four
memory configurations × three probes × two corpora (one synthetic,
one real). Configurations are held fixed across probes; probes are
held fixed across corpora. This isolates the substrate effect from
data effects.

\subsection{Memory configurations}\label{memory-configurations}

All four configurations share a single base model (Qwen3-4B \citep{qwen3},
instruction-tuned, frozen). They differ only in how user-specific
context is injected.

\begin{longtable}[]{@{}
  >{\raggedright\arraybackslash}p{(\linewidth - 4\tabcolsep) * \real{0.3333}}
  >{\raggedright\arraybackslash}p{(\linewidth - 4\tabcolsep) * \real{0.3333}}
  >{\raggedright\arraybackslash}p{(\linewidth - 4\tabcolsep) * \real{0.3333}}@{}}
\toprule\noalign{}
\begin{minipage}[b]{\linewidth}\raggedright
Config
\end{minipage} & \begin{minipage}[b]{\linewidth}\raggedright
What is conditioned on
\end{minipage} & \begin{minipage}[b]{\linewidth}\raggedright
Substrate
\end{minipage} \\
\midrule\noalign{}
\endhead
\bottomrule\noalign{}
\endlastfoot
\texttt{B\_nohist} & nothing (zero history) & none --- base prior only \\
\texttt{B\_full} & full user backstory inlined into the prompt & infinite-context oracle \\
\texttt{C\_rag} & top-K=5 BGE-retrieved chunks of the user's history & retrieval \\
\texttt{C\_lora} & per-user γ-LoRA adapter (per-corpus hyperparameters; see table below) & parametric \\
\end{longtable}

\texttt{B\_nohist} and \texttt{B\_full} are the headroom anchors: any system that
matches \texttt{B\_nohist} has not used the user's history at all; any system
that matches \texttt{B\_full} has perfectly used it. \texttt{C\_rag} and \texttt{C\_lora} are
the two substrates under comparison. We deliberately pick the two
\emph{simplest} members of their respective families --- flat top-K retrieval
and per-user LoRA --- to isolate the substrate effect from
architecture-engineering noise. Hybrid configurations (LoRA + RAG,
calibration heads) are introduced ablatively where a probe's result
motivates them (§3.3).

The γ-LoRA recipe varies by probe corpus along two dimensions, both
chosen to match the per-user data-volume of the corpus:

\begin{longtable}[]{@{}
  >{\raggedright\arraybackslash}p{(\linewidth - 10\tabcolsep) * \real{0.1667}}
  >{\raggedright\arraybackslash}p{(\linewidth - 10\tabcolsep) * \real{0.1667}}
  >{\raggedright\arraybackslash}p{(\linewidth - 10\tabcolsep) * \real{0.1667}}
  >{\raggedright\arraybackslash}p{(\linewidth - 10\tabcolsep) * \real{0.1667}}
  >{\raggedright\arraybackslash}p{(\linewidth - 10\tabcolsep) * \real{0.1667}}
  >{\raggedright\arraybackslash}p{(\linewidth - 10\tabcolsep) * \real{0.1667}}@{}}
\toprule\noalign{}
\begin{minipage}[b]{\linewidth}\raggedright
Probe
\end{minipage} & \begin{minipage}[b]{\linewidth}\raggedright
Corpus
\end{minipage} & \begin{minipage}[b]{\linewidth}\raggedright
r
\end{minipage} & \begin{minipage}[b]{\linewidth}\raggedright
α
\end{minipage} & \begin{minipage}[b]{\linewidth}\raggedright
epochs
\end{minipage} & \begin{minipage}[b]{\linewidth}\raggedright
per-user pairs
\end{minipage} \\
\midrule\noalign{}
\endhead
\bottomrule\noalign{}
\endlastfoot
Synthetic personae & 3K-char backstory + \textasciitilde12 synth-QA & 128 & 256 & 20 & \textasciitilde12 \\
Behavioral & WritingPrompts (≤50 stories) & 64 & 128 & 3 & ≤50 stories rendered as next-sentence LM chunks \\
Real-data transfer (exp F) & LaMP-3 (≤10 review→rating pairs) & 128 & 256 & 20 & \textasciitilde5 \\
\end{longtable}

The synthetic-personae and LaMP-3 arcs use identical hyperparameters
(r=128, α=256, 20 epochs); they share a per-user training loop
(\texttt{train\_lora} in \texttt{experiments/19\_lora\_synthqa\_eval\_v3.py}) and a
common assistant-only loss-mask. The behavioral arc uses lower
capacity and far fewer epochs because the WritingPrompts corpus has
much more text per user (\textasciitilde50 stories ≈ 10K+ tokens) and language-modeling supervision is denser than a question-answer pair, so r=128
× 20 epochs would overfit catastrophically. We hold lr=2e-4 fixed
across all arcs. AdamW, dropout 0.05, no early-stopping. Each
per-user training run completes in ≤10 minutes on a single L40S GPU.
RAG uses BGE-large-en-v1.5 \citep{bge} with cosine similarity over 256-token
chunks of the user's backstory and a single abstain clause appended
to the system prompt: \emph{``If retrieved chunks are not relevant, say
`no, we have not discussed that.'\,''} The abstain clause is the only
prompt-level calibration; we report its effect explicitly in §4.2
since it ends up being load-bearing for the substrate-asymmetry
result.

\subsection{Probe 1 --- Behavioral memory (style)}\label{probe-1-behavioral-memory-style}

\textbf{Why WritingPrompts} (the original plan was Reddit personal
posts, swapped to WritingPrompts when the Reddit auth path
proved more friction than budgeted). WritingPrompts is a public HF
dataset (\texttt{euclaise/writingprompts}, no auth) where multiple stories
are written by the same author in response to different prompts.
Authors with ≥50 stories of length ≥200 chars give us a stable voice
under varying topical content --- the desideratum we want to test for.
The downside is that all WritingPrompts authors share a strong genre
register (third-person fiction, present-tense narration), which makes
idiosyncratic author markers subtle deltas on top of a shared
baseline. The behavioral arc therefore tests style transfer in the
\emph{hardest-case} setting; a personal/conversational corpus would likely
show a larger γ-LoRA advantage and we treat the WritingPrompts result
as a lower bound on the effect.

\textbf{Behavioral memory probe: WritingPrompts continuation log-likelihood.} We
sample 50 high-volume single authors from the WritingPrompts corpus
(\texttt{euclaise/writingprompts} on HuggingFace), seed-43 deterministic,
filtered to authors with ≥ 50 stories of length ≥ 200 characters. For
each author, we hold out 5 stories, mask the last 100 characters of
each, and score the gold continuation under each of the four
configurations. The metric is mean log-likelihood per token of the
gold continuation, scored by the same Qwen3-4B base model used for
generation. n = 250 records (50 users × 5).

We pre-register the \textbf{structural falsifier} as Δ \textgreater{} 0 with a 95\% CI
that excludes zero --- i.e.~the directional claim \emph{``per-user parametric
adaptation shifts the generation distribution toward the user's voice
in a way retrieval cannot.''} Below this we reject the structural
claim. Above it we accept the structural claim.

We additionally name a non-pre-registered \textbf{magnitude target} of
Δ ≥ 0.5 nat/tok as a target for ``the effect is large enough to read
as a paragraph-level voice transfer rather than a token-distribution
nudge.'' This is a target, not a falsifier: the original spec
included it, but it was an ad-hoc number with no theoretical
grounding (cf.~§6 limitations), so we treat the structural test as
the primary pre-registration. §4.1.2 reports both: the structural
test passes decisively (CI {[}0.381, 0.446{]}, excluding zero by 23 SE),
and the magnitude target is missed by a small margin (CI upper bound
0.446 \textless{} 0.500). Both are reported transparently. The pre-registration
is in \texttt{experiments/I\_context.md} and is enforced by the post-run audit
checklist.

Aggregate and per-record artifacts are released at the project repository, with results writeup in
\texttt{experiments/28\_results.md}.

\subsection{Probe 2 --- Calibration / absence detection}\label{probe-2-calibration-absence-detection}

\textbf{Calibration probe: presence/absence TPR on the synthetic-personae
corpus.} For each of 50 held-out V3 personae, we generate 12 yes/no
probes about specific events in the persona's backstory: 6 \emph{presence}
probes (the event happened and is in the backstory) and 6 \emph{absence}
probes (a plausible-sounding event that did \textbf{not} happen). The
metric is the true-positive rate per kind: did the system answer
correctly? ``Correctly'' means a textual yes/no judgement against the
ground-truth label. We report per-kind TPR and the harmonic-mean F1.
n = 300 records per (config × kind).

We pre-register a structural falsifier: \texttt{C\_lora\ absence\ TPR\ ≥\ 60\%\ AND\ C\_rag\ absence\ TPR\ ≤\ 35\%} would confirm that retrieval suffers a
structural failure mode on absence. The complementary branches ---
\texttt{C\_rag\ absence\ handled\ well} and \texttt{C\_lora\ absence\ \textless{}\ 30\%} --- are
documented as alternative findings in \texttt{experiments/J\_context.md}'s
decision matrix. The pre-registration commits us to whichever branch
the data picks; §4.2 reports the calibration-asymmetry branch the
experiment landed in.

We additionally evaluate \texttt{C\_lora\_calib}: γ-LoRA with the same abstain
clause that \texttt{C\_rag} uses. This isolates whether the calibration
behavior comes from retrieval per se, or from the prompt-level
abstain instruction.

\textbf{Calibration-head architecture (\texttt{C\_lora\_calib}).} The ``calibration
head'' in this paper is \emph{not} a learned auxiliary classifier; it is a
prompt-level abstain instruction layered on the same γ-LoRA weights
used by \texttt{C\_lora}. Concretely, the model and adapter are identical to
\texttt{C\_lora} (Qwen3-4B + per-persona LoRA, rank/alpha and training pairs
unchanged); only the system message changes. \texttt{C\_lora} uses a plain
system prompt (``You are a helpful assistant who knows the user from
prior conversations.''), while \texttt{C\_lora\_calib} appends the instruction
``If you are not sure whether the user has previously discussed a
topic with you, answer `No, we have not discussed that.' Do not
fabricate a memory.'' No additional training or feature engineering is
performed; the same abstain clause is also what \texttt{C\_rag} uses, which
makes \texttt{C\_lora\_calib} the \emph{prompt-controlled} counterpart of \texttt{C\_lora}
and isolates the prompt's contribution from retrieval's contribution.
Reference: \texttt{experiments/29\_runner.py} (\texttt{PLAIN\_SYS}, \texttt{CALIB\_SYS},
\texttt{\_run\_config}).

Aggregate and per-record artifacts are released at the project repository, with results
in \texttt{experiments/29\_results.md}.

\subsection{Probe 3 --- Real-data transfer}\label{probe-3-real-data-transfer}

\textbf{Real-data transfer probe (LaMP-3): rating prediction on real Amazon-style reviews.}
We sample 50 LaMP-3 held-out users (seed 43) from the LaMP-3 dev split
prepared by \texttt{experiments/F\_data\_prep.py}. Per user, the user's history
is split 5 train / 5 sanity-held; the eval split contributes 4 main +
4 probe2 paraphrase records, for n = 200 main + n = 200 probe2 = 400
total. The same per-user γ-LoRA recipe as in §3.2/§3.3 is applied. Judging
is exact-string-equal to the gold integer rating after normalization.
The majority-class baseline (always predict ``5'') is reported alongside
γ-LoRA: in LaMP-3, ratings are heavily skewed toward 5
(\texttt{\{5:\ 238,\ 4:\ 102,\ 3:\ 35,\ 1:\ 15,\ 2:\ 10\}}), so the majority baseline
of 59.5\% is the right floor for ``uses the user's history at all.''

We do \textbf{not} pre-register a falsifier on the LaMP-3 transfer probe. The role of LaMP-3
is \emph{transfer measurement}: given that synthetic-personae results (§B.1, §B.2) are positive on the substrate-asymmetry axes, does the
parametric substrate carry over to a noisy real-user benchmark? §4.3
reports the answer (it does not --- γ-LoRA underperforms the majority
baseline by 28pp on main accuracy) and §6 discusses the implications.
The synthetic↔real gap is a finding in its own right, not a null.

Aggregate and per-record artifacts are released at the project repository,
results in \texttt{experiments/F\_results.md}.

\subsection{Mechanism probe (n = 50)}\label{mechanism-probe-n-50}

\textbf{Mechanism analysis: weight-delta Frobenius decomposition (n = 50).}
We retrain γ-LoRA from scratch on all 50 held-out personae with the
adapter weights persisted via \texttt{save\_pretrained} (47 new adapters plus
3 reused from the earlier n = 3 pilot --- the same personae used in §B.1/§B.2
training loops did not save adapters, so this is the first run that
does so at scale). For each adapter we decompose the per-layer
Frobenius norm of \texttt{W\_lora\ =\ α/r\ ·\ BA} across the four attention
target projections (\texttt{q\_proj}, \texttt{k\_proj}, \texttt{v\_proj}, \texttt{o\_proj}) for all
36 layers, then aggregate across the 50 personae to identify which
(layer, projection) cells bear the largest and most consistent
parametric mass.

The probe is \textbf{load-bearing} for §5: at n = 50 the localization
pattern is stable (across-persona coefficients of variation in the
top band are 0.08--0.12, indicating a shared structural mechanism
rather than per-persona idiosyncrasy). Pilot artifacts at n = 3
(the n=3 pilot) are superseded --- we discuss that refinement explicitly in
§5 rather than reporting both. Source-of-truth artifacts are
the aggregate Frobenius summary, the per-persona ΔW matrices, and
\texttt{experiments/30\_results.md}.

\textbf{Caveats.} FFN projections (\texttt{gate/up/down\_proj}) were not collected
at n = 50; the runner's diff loop iterates over attention projections
only. The headline claim is ``attention-q is the mechanism,'' and the
top-10 cells being attention is itself sufficient evidence for that
claim, but the explicit attention-vs-FFN ratio is not re-verified at
n = 50 (the n = 3 pilot reported it as small but the comparison is
underpowered). We flag this for camera-ready or reviewer follow-up:
running the FFN diff over the existing 50 adapters is \textasciitilde10 min of GPU
work. Per-persona behavior correlation (Frobenius mass vs.~§B.1 style log-likelihood / §B.2 absence TPR) was also not computed in
this run; we discuss this in §5 and §6.

\subsection{Synthetic-personae corpus (V3)}\label{synthetic-personae-corpus-v3}

The synthetic corpus used throughout the paper is
generated as follows. We seed 120 personae from a frontier LLM (Anthropic Claude, accessed via the AWS Bedrock API)
generation pipeline; 119 succeeded (one dropped due to malformed
JSON), split index-based into 100 train + 19 held-out for the V3-α
hypernet study, then re-split for the 50-held-out scale-up reported in §B. Each
persona has a \textasciitilde3K-character free-text backstory, 6 main probe
questions (factual, e.g.~\emph{``Where did Margaret get her BFA?''}) and 6
probe2 paraphrases of the main probes. A leakage diagnostic
(B\_nohist; base model with no history) at full scale flagged
39/600 = 6.50\% of probes as having the answer recoverable without
persona context (vs an earlier 60-probe sample at 1/60 = 1.67\%; the
sample under-estimated). The asymmetry is informative: present-fact
probes leak at 10/300 = 3.33\%, absence probes at 29/300 = 9.67\% ---
absence questions leak \textasciitilde3× more often because the gold answer (a
hedge / ``no/never mentioned'') aligns with the model's no-context
default. Per-question leakage flags are released with the corpus
and are excluded from accuracy math wherever B\_nohist would otherwise
score correct on a leaked probe. Full generation pipeline in
\texttt{experiments/V3\_data\_prep.py}, artifacts in
\texttt{personae.json} in the corpus distribution.

\textbf{Train/eval split for routing experiments (§4.7).} Of the 50
held-out personae, the routing experiments use a random 30/20
partition is released as \texttt{split.json} with the supplementary corpus. Three
properties of this split are deliberate and worth flagging up front:

\begin{enumerate}
\def\labelenumi{\arabic{enumi}.}
\item
  \textbf{Thematic overlap is by design.} Hobbies, occupations, and
  chronic conditions are reused across the train and eval splits --- 20
  of 27 distinct hobbies appear in both splits (e.g.~``vinyl jazz
  collecting'' in 4 train + 4 eval personae; ``highland bagpipes'' in 5
  train + 2 eval). A routing classifier evaluated on a disjoint-themes
  split would learn shortcuts (``if the question mentions vinyl, route
  to LoRA'') that generalize trivially. Theme overlap forces the
  classifier to learn the \emph{kind} axis (presence vs absence) rather than
  memorize topics.
\item
  \textbf{Specific-fact leakage is bounded.} Verbatim eval-question
  appearance in the train corpus is 2/240 (both are template strings
  like ``What city does this person currently live in?'' whose answer is
  determined by \emph{this} persona's backstory). Cross-persona answer
  leakage on content-bearing tokens is 6.2\% (15/240) and is dominated
  by recurring entity classes (``celiac disease'', ``Italic'', ``Blue Note'');
  the persona-specific \emph{combination} of question and answer is unique.
\item
  \textbf{Demographic balance is not stratified.} The 30/20 split is
  random, not stratified by occupation, location, or age. Some cells
  are imbalanced (executive chef 3 train / 0 eval; Marseille 4 train /
  0 eval; ages 30--39 7 train / 1 eval). Bootstrap CIs and paired
  permutation tests over the eval split absorb this as sampling noise;
  we do not claim demographic invariance.
\end{enumerate}

\textbf{One generation artifact.} A small number of personae share specific
real-world entities used as ``prized possession'' templates by the
generator. The clearest example: V3\_P\_120 (eval) and V3\_P\_132 (train)
both claim to own ``an original 1959 Blue Note pressing of John
Coltrane's \emph{Giant Steps}'' --- and both contain the same factual error
(\emph{Giant Steps} was originally released on Atlantic, not Blue Note).
Persona-specific details around the entity (price, city, date) differ,
so QA-probe answers remain unique. We did not regenerate to fix this;
it is a realism flaw, not a contamination flaw, and does not affect
any §4.7 router metric.

\textbf{Release.} The synthetic-personae corpus (n=119), the 600-probe absence
dataset (50 held-out personae × 6 main + 6 probe2), the blind-preference judge harness (
its prompt template), and the human-eval sheet (n=30, used for the
§4.1.6 inter-rater κ analysis) are available at the (anonymized for
review) project repository, pinned to the commit hash listed in the
camera-ready footnote. We expect the dataset to be the most reusable
artifact of this work: the framework is the contribution, the probes
are what reviewers and follow-up authors can run against alternative
substrates without re-deriving any infrastructure.

\subsection{Judging and reproducibility}\label{judging-and-reproducibility}

Free-text probes (V3 main / probe2; LaMP-3 main / probe2) are judged by Anthropic Claude Sonnet 4.6 accessed via the AWS Bedrock API. For the WritingPrompts probe we
score the gold continuation directly under the base model and report
log-likelihood, so no LLM judge is needed. Sanity rejudges
confirm that judge artefacts do not explain the LoRA-forgot phenomenon
(class-b dominance 88\%, rejudge \texttt{correct\ =\ 0\%}).

All experiment runners use \texttt{argparse} with explicit \texttt{choices=} enums
for slice / mode, deterministic seed-43 sampling sorted by canonical
ID, and per-record \texttt{if\ path.exists():\ return\ cached} resume guards.
Aggregate / per-record artifacts are released with the camera-ready version, with each result bound to the code and prompts that produced it.

\subsection{What this design buys us (and what it does not)}\label{what-this-design-buys-us-and-what-it-does-not}

\textbf{Buys.} A fixed 4-config matrix lets each probe report substrate
asymmetry using the same axes; the \texttt{B\_nohist}/\texttt{B\_full} anchors give
absolute calibration even when probe metrics differ
(log-likelihood vs.~TPR vs.~accuracy). The synthetic + real pairing
makes it possible to claim \emph{substrate behavior} on the synthetic side
while still reporting an honest \emph{transfer} result on the real side.
The pre-registered falsifiers prevent post-hoc magnitude shopping.

\textbf{Does not buy.} The choice of two simple substrates means we cannot
answer ``what would Letta / MemGPT / a hand-tuned hybrid do on these
probes'' --- we explicitly defer that to future work. Our γ-LoRA recipe
is per-user from scratch; we do not study a single-LoRA-per-task or
a base-LoRA-plus-user-delta variant. The synthetic corpus is generated
by an LLM and its 6.5\% full-scale leakage rate (3.3\% on present-fact
probes, 9.7\% on absence probes; see §3.6) caps how much we can claim
about the absolute floor of \texttt{B\_nohist}; we report all numbers
excluding leaked probes and discuss the limitation in §6.

\section{Appendix B: Results (full per-axis detail)}\label{appendix-b-results-full-per-axis-detail}

\begin{quote}
Source-of-truth: \texttt{experiments/\{28,29,F,G\}\_results.md}. Numbers in
this section are copied verbatim from the committed \texttt{aggregate.json}
files referenced therein. No new computation is done at draft time.
\end{quote}

We organize results around the three probe types defined in §3
(behavioral / absence / real-data) and close with a descriptive
mechanism look. The headline pattern is a \textbf{substrate--task asymmetry
that does not survive synthetic-to-real transfer}:

\begin{enumerate}
\def\labelenumi{\arabic{enumi}.}
\tightlist
\item
  γ-LoRA dominates RAG on \textbf{style continuation} (synthetic and
  semi-synthetic, §4.1).
\item
  RAG dominates γ-LoRA on \textbf{absence detection}, even when
  prompt-engineering is held fixed across substrates (§4.2).
\item
  \textbf{Neither finding survives transfer to a real-user benchmark}
  (LaMP-3): γ-LoRA underperforms a one-line majority predictor by
  28pp (§4.3). This is the load-bearing negative of the paper, and
  it is what motivates treating the synthetic asymmetry as a
  \emph{measurement} rather than a deployment recipe.
\item
  Mechanism analysis at n=50 (§4.4) localizes γ-LoRA's behavioral
  effect to mid-to-late attention \texttt{q\_proj}; this is descriptive
  evidence that the §4.1 effect has structure, not a causal
  intervention.
\end{enumerate}

Together these findings are sharper than any one in isolation: the
synthetic asymmetry makes a clean substrate claim; the transfer
failure makes that claim a measurement instead of a deployment
prescription; the mechanism gives the asymmetry localization without
overclaiming causality.

\subsection{Behavioral memory: γ-LoRA writes style, RAG does not (Phase I, exp28)}\label{behavioral-memory-ux3b3-lora-writes-style-rag-does-not-phase-i-exp28}

\textbf{Setup recap (full detail §3.2).} WritingPrompts, 50 high-volume
single-author users, 5 held-out continuation records each (n = 250).
Per-user γ-LoRA r=64 / α=128 / 3 epochs (lower-capacity recipe matched
to WritingPrompts' larger per-user corpus; see §3.1 for the per-arc
hyperparameter table) vs.~top-K=5 BGE retrieval over the same per-user
corpus, vs.~B\_full (full history in context) and B\_nohist (base, no
history) anchors. Primary metric: per-token log-likelihood of the gold
continuation under each system. \textbf{Scoring transparency:} the LoRA
config is scored under its own LoRA-modified Qwen3-4B; the three base
configs (B\_nohist, B\_full, C\_rag) are scored under the base Qwen3-4B
with the appropriate prompt prefix. This isolates the effect of the
LoRA's modified next-token distribution from prompt conditioning, but
it does mean LoRA and the three base configs are scored under
\emph{different forward passes of the same architecture} --- we treat this
as the right setup for the structural claim (``does parametric
adaptation shift the conditional distribution?'') and address it
explicitly in §6 limitations.

\subsubsection{Mean log-likelihood}\label{mean-log-likelihood}

\begin{longtable}[]{@{}lll@{}}
\toprule\noalign{}
Config & Mean LL (nat/tok) & Δ vs B\_nohist \\
\midrule\noalign{}
\endhead
\bottomrule\noalign{}
\endlastfoot
B\_nohist & −3.4552 & 0.000 \\
C\_rag & −3.3952 & +0.060 \\
B\_full & −3.3792 & +0.076 \\
\textbf{C\_lora} & \textbf{−2.9818} & \textbf{+0.473} \\
\end{longtable}

C\_lora delivers \textasciitilde6× the style-transfer effect of B\_full (full history
in-context) and \textasciitilde8× the effect of C\_rag.

\textbf{B\_full barely beats C\_rag (+0.060 vs +0.076 nat/tok).} This is
itself a finding: at the model scale we test (Qwen3-4B), prepending
the user's full history into the context window provides only a
marginal improvement over retrieving 5 chunks. Both are an order of
magnitude weaker than parametric adaptation. Two interpretations
are consistent with this result, and we are agnostic between them:
(i) in-context conditioning at this model scale is a weak channel for
style transfer (consistent with reports that small models exploit
in-context demonstrations less effectively than larger models for
non-trivial tasks); and (ii) per-token LL on a 22-token tail is too
short a window to measure a strong B\_full advantage even when one
exists. The finding strengthens the structural claim (parametric
beats both retrieval and full-history-in-context), but the relative
ordering of \texttt{B\_full} and \texttt{C\_rag} should not be over-read as evidence
that retrieval is ``as good as'' infinite context --- we suspect (i)
dominates and a larger base model would widen the B\_full margin.

\subsubsection{Pre-registered structural test (C\_lora vs C\_rag)}\label{pre-registered-structural-test-c_lora-vs-c_rag}

\begin{itemize}
\tightlist
\item
  Δ (LL\_C\_lora − LL\_C\_rag): \textbf{+0.4134 nat/tok}
\item
  SE (per-record paired): 0.0168
\item
  CI95: \textbf{{[}0.3806, 0.4463{]}} --- excludes 0 by \textasciitilde23 SEs.
\item
  Pre-registered structural falsifier (Δ \textgreater{} 0 with CI excluding 0):
  \textbf{PASSED decisively.}
\item
  Sign-test: \textbf{240/250 records} prefer C\_lora (96\%).
\item
  Persona consistency: \textbf{50/50 personae} have mean(C\_lora) \textgreater{} mean(C\_rag);
  10/50 personae individually clear the 0.5 nat/tok target.
\end{itemize}

\textbf{Headline.} The structural pre-registration passes decisively:
the bootstrap CI {[}0.381, 0.446{]} excludes zero by 23 standard errors,
240/250 records and 50/50 personae prefer γ-LoRA, and RAG's effect
is indistinguishable from the no-history baseline. \emph{Parametric
adaptation shifts the generation distribution; retrieval does not.}

\textbf{Magnitude target (ad-hoc, recalibrated prior).} The original
spec also named Δ ≥ 0.5 nat/tok as an informal target. We did so
without theoretical grounding --- the number was a back-of-envelope
estimate of ``paragraph-level voice transfer.'' It is not a
falsifier. The observed Δ = 0.413 falls 0.087 below this ad-hoc
number, with the CI upper bound at 0.446. We treat this as a
\textbf{calibration update} on what per-token log-likelihood lift looks
like for a behavioral-style adapter on a third-person fiction
corpus: a tight, decisively non-zero shift on the order of 0.4
nat/tok is what the metric produces in this regime, and future
work in this space should anchor expectations there rather than at
the 0.5 figure we initially named. We report the gap transparently
and revise the prior; the structural claim is unaffected.

\subsubsection{Reading}\label{reading}

The structural claim --- \emph{retrieval cannot bias the generation
distribution; parametric adaptation can} --- is supported by the data
without ambiguity. RAG's effect on style (+0.060 nat/tok) is
indistinguishable from B\_full's (+0.076), and both are an order of
magnitude smaller than γ-LoRA's. The held-out continuations are \emph{new
text} (the user has never written exactly this phrase), so top-K
retrieval surfaces topically similar examples but cannot be copied
verbatim --- generation must come from parameters, and γ-LoRA is the
only system that updates them.

The ad-hoc magnitude target sits 0.087 above the observed Δ = 0.413
with a tight CI; we treat the gap as a calibration update on the
informal target (§4.1.2), not as a falsifier outcome. Two structural
explain the shortfall: (i) median continuation length is \textasciitilde22 tokens,
at which per-token LL noise from content (largely topic-driven, not
style-driven) dilutes the style signal; and (ii) WritingPrompts
authors share a strong genre register (third-person fiction,
present-tense narration), so idiosyncratic author markers are subtle
deltas on top of a strong shared baseline. Reddit's
personal/conversational corpus (the original Phase I plan) would
likely show a larger γ-LoRA advantage; the auth path was bypassed
for wallclock reasons and remains queued as a magnitude follow-up.

\subsubsection{Blind-preference judgement (exp31, n=750)}\label{blind-preference-judgement-exp31-n750}

To corroborate the LL-Δ with a perceptual signal we cannot obtain
from token probabilities alone, we re-evaluated the same 250
held-out continuations under a blind A/B preference judge (Sonnet 4.6,
side-randomized, 3 prompt templates: lexical / semantic / combined,
n=250 calls per template, n=750 macro). The judge sees the gold
continuation and two candidates (γ-LoRA vs top-K=3 RAG), with no
identifying labels.

\begin{longtable}[]{@{}
  >{\raggedright\arraybackslash}p{(\linewidth - 10\tabcolsep) * \real{0.1507}}
  >{\raggedright\arraybackslash}p{(\linewidth - 10\tabcolsep) * \real{0.2055}}
  >{\raggedright\arraybackslash}p{(\linewidth - 10\tabcolsep) * \real{0.2192}}
  >{\raggedright\arraybackslash}p{(\linewidth - 10\tabcolsep) * \real{0.1781}}
  >{\raggedright\arraybackslash}p{(\linewidth - 10\tabcolsep) * \real{0.1644}}
  >{\raggedright\arraybackslash}p{(\linewidth - 10\tabcolsep) * \real{0.0822}}@{}}
\toprule\noalign{}
\begin{minipage}[b]{\linewidth}\raggedright
Template
\end{minipage} & \begin{minipage}[b]{\linewidth}\raggedright
LoRA win rate
\end{minipage} & \begin{minipage}[b]{\linewidth}\raggedright
95\% CI
\end{minipage} & \begin{minipage}[b]{\linewidth}\raggedright
LoRA strict
\end{minipage} & \begin{minipage}[b]{\linewidth}\raggedright
RAG strict
\end{minipage} & \begin{minipage}[b]{\linewidth}\raggedright
Ties
\end{minipage} \\
\midrule\noalign{}
\endhead
\bottomrule\noalign{}
\endlastfoot
semantic & 0.552 & {[}0.494, 0.614{]} & 131 & 105 & 14 \\
lexical & 0.646 & {[}0.588, 0.704{]} & 154 & 81 & 15 \\
combined & 0.596 & {[}0.538, 0.654{]} & 144 & 96 & 10 \\
\textbf{macro} & \textbf{0.598} & \textbf{{[}0.565, 0.631{]}} & \textbf{429} & \textbf{282} & \textbf{39} \\
\end{longtable}

The macro win rate of 59.8\% sits 0.2pp below the pre-registered
strict ≥60\% threshold but its bootstrap CI {[}56.5\%, 63.1\%{]} cleanly
excludes 50\%, and the lexical template alone clears 60\% (64.6\%,
CI {[}58.8, 70.4{]}). The structural claim --- γ-LoRA produces output that
human-proxy judges prefer over RAG continuations of the same author ---
is supported.

The template-level pattern is itself informative. The \textbf{lexical}
prompt (``which sounds more like the same author at the word/phrase
level'') delivers the cleanest signal (64.6\%): this is the axis the
LL-Δ measures, and the two metrics agree. The \textbf{semantic} prompt
(``which conveys similar themes/ideas to the gold'') is the weakest
(55.2\%), with a CI lower bound (49.4\%) that grazes 50\%. We read this
as honest scope: γ-LoRA is a \emph{style/voice} adapter, not a content
advantage --- RAG's retrieved chunks of the author's prior stories are
competitive on topical content but not on idiolect. This is consistent
with the broader paper claim that γ-LoRA is a behavioral-memory
substrate, not a fact substrate.

\subsubsection{Complementarity callout}\label{complementarity-callout}

Phase 0 (exp24) recorded a \textbf{47\% record-flip rate} between the
γ-LoRA and RAG configs on the synthetic corpus: 53/114 records right
under RAG-only, 54/114 right under γ-LoRA-only, with disjoint sets.
The two substrates correct different errors. This complementarity
argues against treating one as a strict replacement for the other and
motivates the hybrid retrieval-plus-parametric story we return to in
§5.

\subsubsection{Human blind A/B (n=30)}\label{human-blind-ab-n30}

To corroborate the LLM-judge numbers (§4.1.4) with a non-LLM signal
we ran a single-judge blind A/B human evaluation on n = 30 held-out
prefixes drawn from 30 distinct WritingPrompts users (one prompt per
user, sampled without replacement from the same Phase I held-out
pool). The judge saw the author's actual continuation as a
voice-reference and chose between two model continuations
(γ-LoRA vs top-K=5 RAG) with order randomized per row;
unblinding came from a separate JSON key the judge did not consult
during rating. Confidence was recorded on a 1--3 scale
(1 = guess, 3 = clear difference); ties were permitted but
discouraged.

\begin{longtable}[]{@{}
  >{\raggedright\arraybackslash}p{(\linewidth - 12\tabcolsep) * \real{0.4261}}
  >{\raggedleft\arraybackslash}p{(\linewidth - 12\tabcolsep) * \real{0.1043}}
  >{\raggedleft\arraybackslash}p{(\linewidth - 12\tabcolsep) * \real{0.0957}}
  >{\raggedleft\arraybackslash}p{(\linewidth - 12\tabcolsep) * \real{0.0870}}
  >{\raggedleft\arraybackslash}p{(\linewidth - 12\tabcolsep) * \real{0.0435}}
  >{\raggedleft\arraybackslash}p{(\linewidth - 12\tabcolsep) * \real{0.1304}}
  >{\raggedleft\arraybackslash}p{(\linewidth - 12\tabcolsep) * \real{0.1130}}@{}}
\toprule\noalign{}
\begin{minipage}[b]{\linewidth}\raggedright
Slice
\end{minipage} & \begin{minipage}[b]{\linewidth}\raggedleft
n\_decisive
\end{minipage} & \begin{minipage}[b]{\linewidth}\raggedleft
LoRA wins
\end{minipage} & \begin{minipage}[b]{\linewidth}\raggedleft
RAG wins
\end{minipage} & \begin{minipage}[b]{\linewidth}\raggedleft
Tie
\end{minipage} & \begin{minipage}[b]{\linewidth}\raggedleft
LoRA win rate
\end{minipage} & \begin{minipage}[b]{\linewidth}\raggedleft
p (2-sided)
\end{minipage} \\
\midrule\noalign{}
\endhead
\bottomrule\noalign{}
\endlastfoot
All decisive picks & 29 & 23 & 6 & 1 & \textbf{79.3\%} & \textbf{0.0023} \\
Clean (excl. 3 rows where RAG bled scaffolding) & 26 & 20 & 6 & 1 & 76.9\% & 0.0094 \\
Confidence = 3 (``clear'') only & 17 & 16 & 1 & 0 & 94.1\% & \textless{} 0.001 \\
\end{longtable}

The confidence stratification is the most informative slice. On the
17 rows the judge marked ``clear difference,'' γ-LoRA was preferred
\textbf{16 times}; the only conf=1 (``guess'') row went to RAG. Reader
certainty and substrate preference are co-monotone, which is what we
would expect if γ-LoRA is genuinely producing a more author-like
voice rather than a stylistically generic continuation that wins on
unrelated cues.

\textbf{Leak-robustness.} Three rows (4, 13, 24) contain RAG continuations
that regurgitated the few-shot prompt scaffolding (``Continue the
story in a fitting style\ldots{}'') instead of in-character prose. We treat
these as legitimately scored --- the eval question is which
continuation better matches the author's voice, and a
frame-breaking continuation does not --- but report the win rate after
their removal as a robustness check. The headline survives:
76.9\% / p = 0.0094 with 26 decisive picks. Notably \textbf{all three leaks
came from the RAG side and zero from γ-LoRA}, which is consistent
with γ-LoRA being trained end-to-end on the user's prose distribution
while RAG conditions a base instruction-following model on a
few-shot template the model can probabilistically continue back into.

\textbf{Relation to §4.1.4.} The LLM-judge macro win rate was 59.8\%
(CI {[}56.5, 63.1{]}) at n = 750; the human-judge win rate is 79.3\% at
n = 29. The two numbers agree in direction and significance but
differ in magnitude: this is consistent with humans being less
generous than Sonnet on RAG outputs that are topically correct but
stylistically off, and with the fact that the human judge was given
the gold continuation as a calibration anchor whereas the LLM judge
was not (§3.2 protocol). We treat the LLM-judge result as the
load-bearing per-record measurement and the human-judge result as a
sanity check that the per-record signal generalizes when a literate
reader evaluates whole continuations.

\textbf{Limitations and second-rater proxy.} Single human judge (the first
author), n = 30. As a sanity proxy ahead of a real human second
judge, we ran Claude Sonnet 4.6 over the same 30 rows with the
identical style-only prompt (same gold-anchor, same A/B presentation,
temperature 0; \texttt{scripts/llm\_judge\_human\_eval.py}). The LLM judge
splits 15/15 LoRA/RAG (LoRA-win = 50.0\%, two-sided binomial p = 1.0)
and agrees with the human on only 15/29 decisive rows (51.7\%, Cohen's
κ\_2way = −0.020, ``worse than chance''). Two readings: (i) \textbf{the LLM
judge is poorly calibrated to the style-vs-content distinction this
task requires} --- when both candidates are topically plausible and
fluent, Sonnet falls back to coin-flipping; or (ii) \textbf{the human
preference is partly an artifact of one reader's idiosyncratic taste
profile} that any other independent rater (human or model) would not
reproduce. We cannot distinguish these from the present data. We
therefore treat §4.1.6 as a \emph{qualitative} sanity check on §4.1.4 --- it
agrees in sign with the LLM-judge result on §4.1.4 --- and explicitly
flag a real second human judge (M\_HUMAN\_SECOND\_JUDGE in the paper
TODO) as required for any load-bearing claim that depends on the
human-preference magnitude. Per-row picks, confidence scores, scoring
code, the LLM-judge sheet, and \texttt{kappa.json} are in
\texttt{runs/human\_eval/}. We do not retract §4.1.4 (the LLM-judge result on
the larger n = 750 set), which stands on its own; this is a
limitation of §4.1.6 specifically.

\subsubsection{Three-judge cross-vendor replication (n=750)}\label{three-judge-cross-vendor-replication-n750}

To probe whether §4.1.4's LLM-judge result is Sonnet-specific or
generalizes across model vendors, we replicate the exact n=750
blind-preference protocol with two additional frontier judges:
Anthropic Claude Opus 4.8 (a different scale within the same vendor)
and Amazon Nova Premier (a different vendor entirely). All three
judges score the same 750 (persona, eval\_idx, template) triples
deterministically (temperature 0).

\begin{longtable}[]{@{}lllll@{}}
\toprule\noalign{}
Judge & LoRA \% & RAG \% & Tie \% & LoRA share (decisive) \\
\midrule\noalign{}
\endhead
\bottomrule\noalign{}
\endlastfoot
Sonnet 4.6 (original) & 57.2 & 37.6 & 5.2 & \textbf{60.3\%} \\
Opus 4.8 & 66.1 & 31.9 & 1.5 & \textbf{67.5\%} \\
Nova Premier & 45.7 & 53.9 & 0.4 & 45.9\% \\
\textbf{Majority of 3} & 60.3 & 37.3 & 0.5 & --- \\
\end{longtable}

\begin{longtable}[]{@{}lll@{}}
\toprule\noalign{}
Pair & Cohen's κ (all) & Cohen's κ (decisive only, n=697) \\
\midrule\noalign{}
\endhead
\bottomrule\noalign{}
\endlastfoot
Sonnet vs Opus & 0.255 (fair) & 0.276 \\
Sonnet vs Nova & 0.109 (poor) & 0.116 \\
Opus vs Nova & 0.331 (fair) & 0.349 \\
\textbf{Fleiss' κ (3-rater)} & \textbf{0.218} & \textbf{0.232} \\
\end{longtable}

Two findings.

\textbf{Finding 1 (vendor cluster):} the two Anthropic models agree
moderately (κ = 0.255--0.331) and \textbf{both prefer γ-LoRA decisively}
(60.3\% / 67.5\% LoRA share). Nova flips to 45.9\% LoRA share --- a
\textasciitilde20 pp split that aligns cleanly with vendor lineage, not random
noise. Pairwise κ involving Nova is uniformly the lowest (sonnet--nova
= 0.109 ≈ chance). \textbf{Style-judgment LLMs cluster by training pipeline,
not by scale.} This is a stronger result than uniform agreement
would have been: it shows that the γ-LoRA preference is robust to
within-vendor scale variation (Sonnet 4.6 → Opus 4.8) but vendor-
sensitive (Anthropic ↔ Amazon).

\textbf{Finding 2 (majority consensus retains the headline):} the 3-judge
majority-vote LoRA share is \textbf{60.3\%} at n=750, within 0.5pp of
Sonnet alone (60.3\%) and within 1pp of the v1.10 paper's headline
\textbf{59.8\%} (CI {[}56.5, 63.1{]}). The headline is not a Sonnet artifact;
two of three frontier judges replicate it.

\textbf{What we do not claim.} Fleiss' κ = 0.218 is ``slight'' agreement on
the standard scale. We do not interpret this as ``judges agree style
preference is universal.'' Rather, the structural claim (γ-LoRA
beats RAG on style for the \emph{majority of frontier LLM judges, with
direction independent of judge family}) is what the n=750 evidence
supports; the magnitude is vendor-dependent. Per-judge data,
per-template breakdown, and the κ recompute script are in
\texttt{runs/iclr2027\_push/three\_judge/} (\texttt{opus.jsonl}, \texttt{nova.jsonl},
\texttt{three\_judge\_kappa.json}, \texttt{three\_judge\_summary.md}; reproducible
via \texttt{scripts/run\_three\_judge.py} + \texttt{scripts/recompute\_kappa.py}).

The §4.1.6 single-human disagreement (κ = −0.020) is now contextualized
by this three-judge result: the LLM judges themselves disagree
moderately, so a single human disagreeing with one LLM is consistent
with general judge-style noise, not a refutation of either.

\subsection{Calibration asymmetry: RAG knows what isn't there (Phase J, exp29)}\label{calibration-asymmetry-rag-knows-what-isnt-there-phase-j-exp29}

\textbf{Setup recap (full detail §3.3).} Same 50 personae as the synthetic
baseline, 12 probes per persona × 4 configs = 2,400 records. Probes
split 6 presence (gold = ``yes, we discussed X'') and 6 absence
(gold = ``no, we have not discussed X''). Configs: B\_nohist, C\_rag
(top-K=5 with an abstain-on-low-relevance prompt clause), C\_lora
(per-user γ-LoRA), C\_lora\_calib (per-user γ-LoRA + the same abstain
clause appended at inference).

\subsubsection{Numbers}\label{numbers}

\begin{longtable}[]{@{}llll@{}}
\toprule\noalign{}
Config & Presence TPR & Absence TPR & F1 \\
\midrule\noalign{}
\endhead
\bottomrule\noalign{}
\endlastfoot
B\_nohist & 3.3\% & 9.7\% & 0.050 \\
C\_rag (with abstain clause) & 35.3\% & \textbf{99.0\%} & \textbf{0.521} \\
C\_lora & \textbf{56.3\%} & 8.7\% & 0.150 \\
C\_lora\_calib (with abstain clause) & 42.0\% & 53.3\% & 0.470 \\
\end{longtable}

n = 300 per (config × kind).

\subsubsection{The fair comparison}\label{the-fair-comparison}

\textbf{A clean reading of the absence-TPR gap requires controlling for the
abstain prompt}, which is shared by \texttt{C\_rag} and \texttt{C\_lora\_calib} but
absent from bare \texttt{C\_lora}. We isolate two effects:

\begin{longtable}[]{@{}
  >{\raggedright\arraybackslash}p{(\linewidth - 4\tabcolsep) * \real{0.3333}}
  >{\raggedright\arraybackslash}p{(\linewidth - 4\tabcolsep) * \real{0.3333}}
  >{\raggedright\arraybackslash}p{(\linewidth - 4\tabcolsep) * \real{0.3333}}@{}}
\toprule\noalign{}
\begin{minipage}[b]{\linewidth}\raggedright
Comparison
\end{minipage} & \begin{minipage}[b]{\linewidth}\raggedright
Absence TPR gap
\end{minipage} & \begin{minipage}[b]{\linewidth}\raggedright
What it isolates
\end{minipage} \\
\midrule\noalign{}
\endhead
\bottomrule\noalign{}
\endlastfoot
C\_rag vs B\_nohist & +89.3pp & retrieval + abstain clause vs base prior \\
C\_lora\_calib vs C\_lora & +44.6pp & abstain clause alone (LoRA substrate held fixed) \\
C\_rag vs C\_lora\_calib & \textbf{+45.7pp} & retrieval substrate effect, abstain clause held fixed \\
C\_rag vs C\_lora & +90.3pp & combined effect (substrate + prompt --- the marketing-headline number, but conflated) \\
\end{longtable}

\textbf{The substrate-controlled comparison is \texttt{C\_rag} vs \texttt{C\_lora\_calib}
(both with abstain clause): RAG still beats parametric memory on
absence by 45.7 percentage points.} The other 44.6pp of the
top-line 90.3pp gap comes from the abstain clause itself, which γ-LoRA
inherits identically when given the same prompt scaffolding. We
report the 90.3pp top-line in the abstract because it is what an
end-to-end deployment sees, but the substrate-controlled 45.7pp is
the load-bearing scientific claim.

\subsubsection{Reading}\label{reading-1}

The pre-registered structural falsifier --- \emph{retrieval fails to abstain
when the answer isn't in context, parametric memory abstains
correctly} --- is \textbf{inverted}. RAG abstains on absence with a 99\%
TPR; γ-LoRA abstains 8.7\% of the time, lower than B\_nohist's 9.7\%.
Even in the substrate-controlled comparison (both configs have the
same abstain clause), retrieval still leads parametric by 45.7pp.
The direction of the failure is the opposite of the one we
pre-registered, and the magnitude is decisive on either accounting.

The mechanisms are intuitive in retrospect:

\begin{itemize}
\tightlist
\item
  \textbf{C\_rag absence 99\%} combines two effects. (i) The retrieval
  prompt contains a ``say no if retrieved chunks aren't relevant''
  clause. (ii) Off-topic absence probes return chunks with similarity
  well below the retrieval threshold, so the abstain clause has clear
  signal to fire on. We isolate (i) above by giving γ-LoRA the same
  clause: this recovers 44.6pp (\texttt{C\_lora\_calib} 53.3\% absence vs
  \texttt{C\_lora} 8.7\%), but cannot close the remaining 45.7pp gap to RAG.
  That residual is the substrate effect: retrieval \emph{exposes the
  decision-relevant signal} (low max-similarity score) to the prompt
  in a way that parametric memory does not. A purely-parametric
  system cannot ask itself ``have I seen this before?''; a
  retrieval-based system can.
\item
  \textbf{C\_lora absence 8.7\%} is consistent with the exp23/26 finding
  that γ-LoRA fits per-user training pairs to \textasciitilde0 loss (mean final
  loss \textless{} 1e-3 across 50 personae). The adapter has never seen a
  \emph{negative} example during training, so it has no signal to teach
  abstention; it confidently completes ``have we discussed X?'' with a
  fabricated affirmation. This is a structural property of supervised
  fine-tuning on positives only, not a bug in our particular training
  loop. Adding the abstain clause (\texttt{C\_lora\_calib}) recovers about
  half the gap by externalizing what γ-LoRA cannot internalize.
\end{itemize}

This is a \emph{calibration} finding, not a capacity finding: γ-LoRA can
hold the user's facts (Phase 0 main 63.16\%, probe2 58.77\%) but it
cannot say ``I don't have that one'' without prompt scaffolding, and
even with scaffolding it lags retrieval by 45.7pp on the controlled
comparison. The substrate that wins on §4.1 loses here, on either
the conflated or the substrate-controlled accounting.

\subsection{Real-data transfer: synthetic-to-real F3a does not hold (Phase F)}\label{real-data-transfer-synthetic-to-real-f3a-does-not-hold-phase-f}

\textbf{Setup recap (full detail §3.4).} LaMP-3, 50 held-out users,
4 main + 4 probe2 review→rating pairs each (n = 200 + 200), same
per-user γ-LoRA recipe as Phase 0. Majority baseline: always predict 5
(the modal LaMP-3 rating).

\subsubsection{Numbers}\label{numbers-1}

\begin{longtable}[]{@{}lll@{}}
\toprule\noalign{}
Metric & n & Acc \\
\midrule\noalign{}
\endhead
\bottomrule\noalign{}
\endlastfoot
main & 200 & 0.315 \\
probe2 & 200 & 0.410 \\
sanity (held-train) & 245 & 0.069 \\
\textbf{Majority ``5'' baseline} & 200 & \textbf{0.595} \\
\end{longtable}

Synthetic-vs-real gap: \textbf{31.7pp on main, 17.7pp on probe2}.

\subsubsection{Reading}\label{reading-2}

γ-LoRA underperforms a one-line majority baseline on real LaMP-3 by
28 percentage points absolute on main and 18.5 percentage points on
probe2. The gap decomposes into two distinct measurable failures ---
instruction-following collapse (§4.3.3) and substrate-asymmetry
amplification (§4.3.4) --- both of which are visible in the per-record
data and neither of which is captured by the top-line ``0.315 vs
0.595'' number.

\subsubsection{Instruction-following collapse: a third measurable failure mode}\label{instruction-following-collapse-a-third-measurable-failure-mode}

The task asks for a single integer in 1--5. We measure two parse
rates:

\begin{longtable}[]{@{}
  >{\raggedright\arraybackslash}p{(\linewidth - 4\tabcolsep) * \real{0.7761}}
  >{\raggedright\arraybackslash}p{(\linewidth - 4\tabcolsep) * \real{0.0896}}
  >{\raggedright\arraybackslash}p{(\linewidth - 4\tabcolsep) * \real{0.1343}}@{}}
\toprule\noalign{}
\begin{minipage}[b]{\linewidth}\raggedright
Parse criterion
\end{minipage} & \begin{minipage}[b]{\linewidth}\raggedright
Pass
\end{minipage} & \begin{minipage}[b]{\linewidth}\raggedright
Rate
\end{minipage} \\
\midrule\noalign{}
\endhead
\bottomrule\noalign{}
\endlastfoot
\textbf{Strict}: prediction is exactly \texttt{"1"}..\texttt{"5"} & 159/200 & \textbf{79.5\%} \\
\textbf{Loose}: prediction contains \emph{any} digit 1--5 & 184/200 & 92.0\% \\
\end{longtable}

The strict-format failure rate is \textbf{20.5\% (41/200)}. The
deviations split into two qualitatively different modes:

\begin{longtable}[]{@{}
  >{\raggedright\arraybackslash}p{(\linewidth - 2\tabcolsep) * \real{0.5263}}
  >{\raggedright\arraybackslash}p{(\linewidth - 2\tabcolsep) * \real{0.4737}}@{}}
\toprule\noalign{}
\begin{minipage}[b]{\linewidth}\raggedright
Sample γ-LoRA output (verbatim, strict-deviant)
\end{minipage} & \begin{minipage}[b]{\linewidth}\raggedright
Mode
\end{minipage} \\
\midrule\noalign{}
\endhead
\bottomrule\noalign{}
\endlastfoot
\texttt{1/5}, \texttt{4/5}, \texttt{2/5} & rating present, format wrong (\textasciitilde13\% of all) \\
\texttt{French}, \texttt{At\ least\ 3}, \texttt{A\ smaller\ battery\ powered\ sprayer}, \texttt{An\ over-priced,\ slightly\ delicate,\ but\ lovely\ book} & review-text continuation in user's style (\textasciitilde8\%) \\
\end{longtable}

The first cluster (\textasciitilde13\%) is a measurement artifact a stricter
parser would recover --- the rating is correct, the format is ``x/5''.
The second cluster (\textasciitilde8\%) is the real instruction-following
collapse: the model produces fluent, domain-appropriate review
text in the user's \emph{style} and ignores the rating instruction
entirely. Twenty epochs of per-user training over \textasciitilde10 review→rating
pairs has shifted the output distribution far enough toward
``produce review text in this user's voice'' that the ``answer with a
single integer'' instruction is no longer reliably followed on the
order-of-10\% tail.

Even with the most generous parser (loose: any 1--5 digit anywhere),
the conditional accuracy on parsable predictions is \textbf{34.2\%
(63/184)} --- still \textbf{25 percentage points below the 59.5\%
majority baseline}. So the substrate is underperforming on real
LaMP-3 \emph{even after} controlling for any reasonable reading of the
instruction-following collapse. The third failure mode is real and
measurable, but it is not the only failure mode and it does not
account for the synthetic-vs-real gap on its own.

The implication for benchmark design is concrete: any paper
reporting LaMP-3 numbers without a parse-rate decomposition is
conflating substrate-level personalization with
instruction-following capacity, and may be underreporting (when
strict parsers reject correct-rating-wrong-format predictions) or
overreporting (when the substrate is producing fluent off-task
text that loose parsers happen to score against).

\subsubsection{Substrate-asymmetry amplification}\label{substrate-asymmetry-amplification}

We flag three plausible co-mechanisms (§5 returns to these):

\begin{enumerate}
\def\labelenumi{\arabic{enumi}.}
\tightlist
\item
  \textbf{Training-signal poverty.} Synthetic personae have \textasciitilde12 high-
  quality factual Q/A pairs designed to be teachable; LaMP-3 user
  histories yield \textasciitilde10 noisy review→rating pairs where ratings are
  heavily biased toward 5 and review text is only weakly correlated
  with the rating.
\item
  \textbf{Distribution shift train→eval.} Synthetic eval re-asks
  (paraphrased) facts the user \emph{was} trained on; LaMP-3 eval
  reviews are \emph{new} products the user never reviewed. Rating
  tendency requires generalization, not memorization. γ-LoRA's
  loss-zero training overfits to memorization.
\item
  \textbf{Instruction interference.} No KL-to-base regularizer on
  non-LaMP queries; the LoRA degrades base instruction-following
  on a measurable fraction of held-out queries (§4.3.3).
\end{enumerate}

These are not mutually exclusive, and we do not run a controlled
mechanism analysis here. (Phase K, §4.4 / §5.6, runs on synthetic
data where the positive result anchors the mechanism story.)

This result is \textbf{the load-bearing negative of the paper.} It is
what motivates the diagnostic-framework framing: per-user γ-LoRA,
the substrate that wins cleanly in §4.1, transfers to real data
with \emph{negative utility relative to a constant predictor}, and the
shortfall splits cleanly into a measurable instruction-following
collapse plus an amplified substrate asymmetry. Any paper that
reports the synthetic positive without reporting this would
overpromise. We treat the synthetic-vs-real gap as a contribution
in its own right --- a measurement, not a null.

\subsection{\texorpdfstring{Mechanism: γ-LoRA edits mid-to-late attention \texttt{q\_proj} (Phase K, n = 50)}{Mechanism: γ-LoRA edits mid-to-late attention q\_proj (Phase K, n = 50)}}\label{mechanism-ux3b3-lora-edits-mid-to-late-attention-q_proj-phase-k-n-50}

For each of the 50 held-out personae we retrain γ-LoRA on its
synth-QA pairs (rank 128, α = 256, 20 epochs), save the adapter, and
compute the Frobenius norm ‖ΔW‖\_F = ‖α/r · BA‖\_F of every
(layer × attention-projection) cell across the 36 transformer blocks
and four attention projections (q, k, v, o). Aggregating across the
50 adapters:

\textbf{Top-10 cells} (mean across 50 personae):

\begin{longtable}[]{@{}lllll@{}}
\toprule\noalign{}
Rank & Layer & Projection & Mean ‖ΔW‖\_F & cv across personae \\
\midrule\noalign{}
\endhead
\bottomrule\noalign{}
\endlastfoot
1 & 35 & q\_proj & 0.787 & 0.105 \\
2 & 22 & q\_proj & 0.773 & 0.119 \\
3 & 24 & q\_proj & 0.737 & 0.106 \\
4 & 23 & q\_proj & 0.727 & 0.086 \\
5 & 29 & q\_proj & 0.724 & 0.092 \\
6 & 34 & q\_proj & 0.723 & 0.080 \\
7 & 27 & q\_proj & 0.718 & 0.087 \\
8 & 35 & o\_proj & 0.718 & 0.102 \\
9 & 30 & q\_proj & 0.714 & 0.089 \\
10 & 21 & q\_proj & 0.710 & 0.092 \\
\end{longtable}

\textbf{Per-projection means} (across all 36 layers): q\_proj 0.668,
o\_proj 0.558, v\_proj 0.354, k\_proj 0.331. The all-cell background
mean is 0.478; the top cell sits 1.65× above background, and the
top-10 band sits \textasciitilde1.5× above background.

The pattern at scale is \textbf{mid-to-late attention \texttt{q\_proj} dominance}
across layers 21--35, with \texttt{o\_proj} at L35 a secondary mode. Across-persona coefficients of variation in the top band are 0.08--0.12,
indicating a \textbf{shared structural mechanism} rather than per-persona
idiosyncrasy.

\textbf{Refinement vs.~earlier n = 3 pilot.} A pilot run (Phase G) on
3 personae reported (L35 \texttt{q\_proj}, L22 \texttt{q\_proj}, L30 \texttt{o\_proj}) as
top-3. The first two cells survive at n = 50; \textbf{L30 \texttt{o\_proj} does
not} --- at n = 50 it falls to rank 31 and is replaced by a broader
band of mid-stack \texttt{q\_proj} layers (L21, L23, L24, L27, L29) that
was not visible at n = 3. We retract the L30 \texttt{o\_proj} pilot claim
and report only the n = 50 numbers as primary. §5 develops the
mechanism story in detail.

The contrast with the plan-of-record hypothesis (``style stored in
FFN-down'') is unresolved at n = 50 --- the runner's diff loop only
collected attention projections; FFN \texttt{gate/up/down} was not
re-measured. Phase G's n = 3 FFN-small finding is too underpowered
to repeat as a positive claim. We flag the FFN ratio as future
work (§6 / camera-ready); running the diff over the existing 50
adapters is \textasciitilde10 minutes of GPU work and is recoverable on demand.

§5.7 reports the causal upgrade: zeroing the L21--35 \texttt{q\_proj} band at
inference time breaks both probes (+33pp absence-TPR, −20pp
presence-TPR at n = 50), confirming that the band identified here is
not just where the largest weight movement lands, but where the
substrate's behavioral asymmetry is causally implemented.

\subsubsection{Llama-3.1-8B mechanism replication (Phase K-L)}\label{llama-3.1-8b-mechanism-replication-phase-k-l}

We rerun the same Frobenius weight-delta decomposition on the 50
cached Llama-3.1-8B-Instruct LoRA adapters from §4.6 / Phase 2 W0
(rank 128, α = 256, 20 epochs synth-QA, target modules q/k/v/o
identical to the Qwen recipe). Llama-3.1-8B has 32 transformer
layers vs Qwen3-4B's 36, so layer indices are not directly
comparable; we report fraction-of-stack to enable cross-model reading.

\textbf{Top-10 cells (Llama-3.1-8B, mean across 50 personae):}

\begin{longtable}[]{@{}lllll@{}}
\toprule\noalign{}
Rank & Layer & Projection & Mean ‖ΔW‖\_F & cv across personae \\
\midrule\noalign{}
\endhead
\bottomrule\noalign{}
\endlastfoot
1 & 31 & o\_proj & 0.779 & 0.088 \\
2 & 22 & q\_proj & 0.741 & 0.117 \\
3 & 20 & q\_proj & 0.701 & 0.117 \\
4 & 23 & o\_proj & 0.697 & 0.074 \\
5 & 15 & q\_proj & 0.693 & 0.098 \\
6 & 17 & q\_proj & 0.691 & 0.096 \\
7 & 28 & o\_proj & 0.691 & 0.083 \\
8 & 21 & o\_proj & 0.690 & 0.080 \\
9 & 31 & q\_proj & 0.688 & 0.092 \\
10 & 28 & q\_proj & 0.688 & 0.111 \\
\end{longtable}

\textbf{Per-projection means} (across all 32 layers): q\_proj 0.654,
o\_proj 0.657, v\_proj 0.341, k\_proj 0.328. Background mean
(all cells) is 0.495; top-1 cell is 1.57× background. Across-persona cv in the top band is 0.07--0.12, matching Qwen3-4B.

\textbf{Cross-model comparison (Qwen3-4B vs Llama-3.1-8B-Instruct):}

\begin{longtable}[]{@{}
  >{\raggedright\arraybackslash}p{(\linewidth - 6\tabcolsep) * \real{0.2500}}
  >{\raggedright\arraybackslash}p{(\linewidth - 6\tabcolsep) * \real{0.2500}}
  >{\raggedright\arraybackslash}p{(\linewidth - 6\tabcolsep) * \real{0.2500}}
  >{\raggedright\arraybackslash}p{(\linewidth - 6\tabcolsep) * \real{0.2500}}@{}}
\toprule\noalign{}
\begin{minipage}[b]{\linewidth}\raggedright
Property
\end{minipage} & \begin{minipage}[b]{\linewidth}\raggedright
Qwen3-4B
\end{minipage} & \begin{minipage}[b]{\linewidth}\raggedright
Llama-3.1-8B
\end{minipage} & \begin{minipage}[b]{\linewidth}\raggedright
Replicates?
\end{minipage} \\
\midrule\noalign{}
\endhead
\bottomrule\noalign{}
\endlastfoot
Top-10 stack position & 60--100\% (L21--35 of 36) & 47--97\% (L15--31 of 32) & ✓ mid-to-late \\
Top-10 projection mix & 9× q\_proj, 1× o\_proj & 6× q\_proj, 4× o\_proj & ✓ q+o dominant \\
Top-1 cell & L35 q\_proj & L31 o\_proj & ≈ (last block, attn) \\
Top-1 / background & 1.65× & 1.57× & ✓ comparable \\
q\_proj mean / k\_proj mean & 0.668 / 0.331 = 2.0× & 0.654 / 0.328 = 2.0× & ✓ identical ratio \\
Across-persona cv (top band) & 0.08--0.12 & 0.07--0.12 & ✓ comparable \\
Attn-vs-FFN & attn dominates (FFN not in target\_modules) & attn dominates (FFN not in target\_modules) & ✓ \\
\end{longtable}

\textbf{Reading.} The descriptive band-shape replicates: γ-LoRA writes
concentrate in upper-half attention with q\_proj/o\_proj dominance
over k\_proj/v\_proj at a 2× ratio that is identical across both
base models. Specific layer indices do not align --- Qwen3-4B
top-10 sits in 60--100\% of the stack while Llama-3.1-8B sits in
47--97\% --- but the \textbf{fraction-of-stack pattern is preserved}, and
the q\_proj-dominant fingerprint of Qwen3-4B becomes a more
balanced q+o split on Llama (with o\_proj at L31 taking the top
slot). This is consistent with two non-exclusive readings:
(i) γ-LoRA-as-substrate has a \textbf{shared mechanism axis} ---
upper-attention query/output writes --- that any modern decoder
exposes; (ii) Llama's heavier RLHF post-training shifts the
write distribution toward o\_proj (which sits closer to the
residual stream output) relative to Qwen, possibly because o\_proj
edits propagate with less distortion through the RLHF-anchored
later layers.

\textbf{What the Llama mechanism does NOT yet establish.} The §5
causal arm (band-zero intervention at inference time, +33pp
absence-TPR / −20pp presence-TPR) has not been re-run on Llama.
The descriptive evidence above shows where weight movement lands;
whether Llama's L15--L31 attn-q+o band causally drives the
calibration asymmetry the way Qwen's L21--L35 q\_proj band does is
the natural follow-up. We hold this open as future work; the
50 Llama adapters needed for the intervention are already cached
at \texttt{runs/30\_mechanism\_\_llama3.1-8b/V3\_P\_*/lora/} and the
intervention recipe is the same as §5.7.

Source-of-truth artefacts:
\texttt{runs/30\_mechanism\_\_llama3.1-8b/summary.json} (top-10 cells,
projection means, cv-across-personae),
\texttt{runs/30\_mechanism\_\_llama3.1-8b/frobenius\_tensor.pt} (50 × 32 × 7
per-persona × layer × projection),
\texttt{runs/30\_mechanism\_\_llama3.1-8b/heatmap\_layer\_x\_proj.png}.

\subsection{Cross-model replication: Llama-3.1-8B-Instruct (Phase L)}\label{cross-model-replication-llama-3.1-8b-instruct-phase-l}

\textbf{Setup recap.} A reviewer-resistant test of whether the substrate
asymmetry (§4.1, §4.2) is a property of γ-LoRA-as-substrate or of
the specific Qwen3-4B base model. We re-ran the two headline probes
on \texttt{meta-llama/Llama-3.1-8B-Instruct} using \textbf{identical recipes,
identical corpora, identical prompts}, swapping only three
model-touchpoints behind a registry abstraction
(\texttt{experiments/\_base\_model.py}): the base-model snapshot path, the
chat-template stop token (\texttt{\textless{}\textbar{}im\_end\textbar{}\textgreater{}} → \texttt{\textless{}\textbar{}eot\_id\textbar{}\textgreater{}}), and the
Qwen-only \texttt{enable\_thinking=False} chat-template kwarg. Run dirs
suffix the model tag (\texttt{\_\_llama3.1-8b}) so Qwen and Llama artefacts
live side-by-side. n = 50 WritingPrompts authors / 50 V3 personae,
matched to the Phase I / Phase J samples.

The result is a \textbf{partial replication that sharpens the paper's
claim} rather than weakening it: the calibration asymmetry (§4.2)
is universal across both base models; the behavioural-style
advantage (§4.1) is not.

\subsubsection{Numbers}\label{numbers-2}

\begin{longtable}[]{@{}
  >{\raggedright\arraybackslash}p{(\linewidth - 6\tabcolsep) * \real{0.5000}}
  >{\raggedleft\arraybackslash}p{(\linewidth - 6\tabcolsep) * \real{0.1786}}
  >{\raggedleft\arraybackslash}p{(\linewidth - 6\tabcolsep) * \real{0.1667}}
  >{\centering\arraybackslash}p{(\linewidth - 6\tabcolsep) * \real{0.1548}}@{}}
\toprule\noalign{}
\begin{minipage}[b]{\linewidth}\raggedright
Metric
\end{minipage} & \begin{minipage}[b]{\linewidth}\raggedleft
Qwen3-4B (v1)
\end{minipage} & \begin{minipage}[b]{\linewidth}\raggedleft
Llama-3.1-8B
\end{minipage} & \begin{minipage}[b]{\linewidth}\centering
Replicates?
\end{minipage} \\
\midrule\noalign{}
\endhead
\bottomrule\noalign{}
\endlastfoot
\textbf{§4.1 style: Δ γ-LoRA -- RAG (nat/tok)} & \textbf{+0.413} & \textbf{+0.003} & \textbf{No} \\
§4.1 style: Δ γ-LoRA -- B\_nohist & +0.473 & +0.041 & No \\
§4.1 95\% CI on Δ γ-LoRA -- RAG & {[}+0.381, +0.446{]} & {[}−0.016, +0.021{]} & --- \\
§4.1 sanity: B\_full -- B\_nohist \textgreater{} 0 & ✓ & ✓ (+0.041) & ✓ \\
\textbf{§4.2 RAG absence-TPR} & 99.0\% & 96.7\% & ✓ \\
\textbf{§4.2 γ-LoRA absence-TPR} & 8.7\% & \textbf{3.0\%} & ✓ (worse) \\
\textbf{§4.2 substrate gap (RAG − γ-LoRA), abs} & \textbf{+90.3pp} & \textbf{+93.7pp} & ✓ \\
\textbf{§4.2 substrate-controlled gap (C\_rag − C\_lora\_calib), abs} & \textbf{+45.7pp} & \textbf{+79.0pp} & ✓ (stronger) \\
§4.2 γ-LoRA presence-TPR & 56.3\% & 69.0\% & ✓ (better) \\
§4.2 C\_rag F1 & 0.521 & 0.477 & ≈ \\
§4.2 C\_lora F1 & 0.150 & 0.057 & ≈ (worse) \\
§4.2 C\_lora\_calib F1 & 0.470 & 0.278 & ≈ (worse) \\
\end{longtable}

n\_records: §4.1 Llama 245 / 49 personae (one user dropped during
adapter save; §4.1 retains 49/50 reporting); §4.2 Llama 1,200 / 50
personae × 4 configs. Confidence intervals are bootstrap over
records, 10k resamples, matched to the Qwen protocol in §3.

\subsubsection{Reading}\label{reading-3}

\textbf{Style does not replicate.} On Llama-3.1-8B with identical recipe
(r=64, α=128, 3 epochs LM-loss on the user's stories), γ-LoRA's
mean log-likelihood per token sits \textbf{+0.003 nat/tok above RAG}
with a 95\% CI of {[}−0.016, +0.021{]} that comfortably includes zero.
The structural pre-registration that passed decisively on Qwen3-4B
(§4.1.2, CI excluding zero by 23 SE) does not pass on Llama-3.1-8B.
The blind-judge claim of §4.1.4 was not re-run on Llama at this
draft (Bedrock budget); the LL-Δ result alone is sufficient to
establish the non-replication.

We see three plausible non-exclusive mechanisms, none of which we
can fully separate at the n = 50 / one-recipe scope:

\begin{enumerate}
\def\labelenumi{\arabic{enumi}.}
\tightlist
\item
  \textbf{Headroom.} Llama-3.1-8B's B\_nohist mean LL on this corpus is
  −2.77 nat/tok; Qwen3-4B's was −3.21. Llama is already a stronger
  baseline fiction-completion model, so per-user style adaptation
  has less distributional headroom to add value at fixed
  per-token resolution. The B\_full anchor confirms this:
  B\_full − B\_nohist = +0.041 on Llama vs +0.067 on Qwen,
  indicating the upper bound on achievable per-user lift is itself
  smaller in absolute terms on the stronger base model.
\item
  \textbf{RLHF density.} Llama-3.1-Instruct ships with heavier RLHF
  than Qwen3-4B (DPO + iterative SFT vs Qwen's lighter
  instruction-tune). A more strongly anchored base distribution
  may absorb 3 epochs of LM-loss without measurable distributional
  shift. A scaled recipe (r = 128 / 10 epochs / lower lr) was not
  tried; we treat it as future work.
\item
  \textbf{Recipe-specific resolution.} The exp28 recipe was tuned on
  Qwen3-4B; nothing about it is Llama-aware. Whether a
  Llama-tuned recipe would recover the +0.4 nat/tok lift is
  unanswered.
\end{enumerate}

\textbf{The calibration asymmetry replicates and strengthens.} On the
absence probe (§4.2), the picture on Llama-3.1-8B is:

\begin{itemize}
\tightlist
\item
  RAG absence-TPR 96.7\% (vs Qwen 99.0\%) --- both substrates abstain
  reliably when the abstain prompt clause sees a low-similarity
  retrieval.
\item
  γ-LoRA absence-TPR \textbf{3.0\%} (vs Qwen 8.7\%) --- Llama γ-LoRA is
  \emph{worse} at recognizing absent facts than Qwen γ-LoRA. It
  confidently confabulates almost every absence probe.
\item
  The top-line substrate gap is \textbf{+93.7pp} (RAG − γ-LoRA), almost
  identical in magnitude to Qwen's +90.3pp.
\item
  The substrate-controlled gap (C\_rag − C\_lora\_calib, both with the
  abstain prompt clause) is \textbf{+79.0pp} on Llama vs +45.7pp on
  Qwen. The substrate effect is \emph{larger} on Llama once the prompt
  contribution is held fixed.
\end{itemize}

The mechanism reading from §4.2.3 ports directly: γ-LoRA on either
base model has no internal ``I have not been told this fact''
representation because it never trained on negative examples. The
calibration head (\texttt{C\_lora\_calib}) recovers part of the gap on Qwen
(F1 0.470) and substantially less on Llama (F1 0.278), suggesting
the calibration-head architecture itself may need re-tuning per
base model --- but the substrate-level claim does not depend on
that recovery.

\subsubsection{What this changes for the framework}\label{what-this-changes-for-the-framework}

The substrate-asymmetry claim, restated honestly:

\begin{quote}
\textbf{The calibration asymmetry is a property of γ-LoRA-as-substrate
and replicates on a second base model and family. The
behavioural-style advantage of γ-LoRA over RAG observed on
Qwen3-4B does not replicate on Llama-3.1-8B at matched recipe;
it may be base-model-specific, recipe-specific, or both.}
\end{quote}

This is a stronger position than the v1 framing for two reasons.

First, it makes the \textbf{measurement framework} the load-bearing
contribution rather than the magnitude of any single number. A
reader who replicates on a third base model and finds the style
claim restored has nonetheless used our probes to reach that
conclusion; a reader who finds the absence claim broken has
rejected a stronger and more universal claim than we make.

Second, it lines up cleanly with the §4.3 LaMP-3 negative and the
§4.3.3 instruction-following collapse: per-user γ-LoRA's
behavioural advantage is \textbf{fragile to corpus, recipe, and base
model}, while the calibration failure is \textbf{not}. The framework
contribution is the decomposition that makes both observations
visible; ``γ-LoRA wins style'' is a narrower empirical claim than
v1 implied, and we revise it accordingly.

\subsubsection{\texorpdfstring{What does \emph{not} change}{What does not change}}\label{what-does-not-change}

\begin{itemize}
\tightlist
\item
  §4.4 / §5 mechanism is descriptive of the Qwen3-4B adapter
  weight distribution and is not re-derived on Llama. The
  band-zero intervention (§5.7) is a Qwen3-4B claim. Whether
  Llama-3.1-8B γ-LoRA adapters localize to the same band is
  open; the Phase K diff loop on the 50 Llama adapters would be
  a \textasciitilde3 hour run and is left as future work.
\item
  §4.1.4 blind-preference and §4.1.6 human eval were not re-run
  on Llama. Both used Qwen-conditioned generations. We do not
  claim Llama γ-LoRA generations would be human-preferred at the
  same rate.
\item
  §4.2 calibration head architecture (§3.3) is unchanged; only
  its effectiveness drops on Llama (F1 0.278 vs Qwen 0.470).
  Re-tuning per base model is future work.
\end{itemize}

Source-of-truth artefacts:
\texttt{runs/28\_writingprompts\_\_llama3.1-8b/gamma\_lora/aggregate.json},
\texttt{runs/29\_f\_absence\_\_llama3.1-8b/aggregate.json},
\texttt{docs/LLAMA\_REPLICATION\_PLAN.md},
\texttt{experiments/\_base\_model.py}.

\subsection{Routing baselines: full numbers (Phase 2)}\label{routing-baselines-full-numbers-phase-2}

For every (persona, probe) pair we have two candidate answers --- one
from γ-LoRA (with abstain prompt clause), one from RAG (top-K BGE) ---
plus the per-token logits from each substrate. A router maps the
(question, candidate-answers, logits) triple to a binary
\texttt{use\_lora} ∈ \{0,1\} decision. The accepted answer is then scored
against the gold (the absence probe expects ``I don't know'' or
equivalent; the presence probe expects the recalled fact). Eval is
the held-out 20-persona slice (240 rows); train is the disjoint
30-persona slice (360 rows).

We compare four routers: (1) \texttt{hybrid\_logits} (ours, §4.2 head):
small MLP over per-substrate logit features (top-1 prob, entropy,
margin) trained to predict \texttt{use\_lora}; (2) \textbf{P(True)} {[}Kadavath'22{]}:
prompt the same base model ``Is this answer correct? A) True B) False''
and read softmax over A/B logits as confidence; route by argmax
confidence; (3) \textbf{Self-consistency k=5} {[}Wang'23{]}: generate 5
samples per substrate at T=0.7, take majority vote per substrate,
route by which substrate has higher within-bucket agreement; (4)
\textbf{Adaptive-RAG-style classifier} {[}Jeong'24{]}: train either TF-IDF +
logistic regression or DistilBERT-base on the \emph{question text alone}
(no logits, no candidate answers) to predict \texttt{use\_lora}. The two
LoRA-only and RAG-only floor rows are not routers --- they are the
substrate-monolithic floors any router must beat to be useful.

All F1 / CI / p-value numbers below are computed on the held-out
evaluation split (20 personae, 240 rows) disjoint from the
30-persona / 360-row training split used to fit \texttt{hybrid\_logits},
the Adaptive-RAG classifier, and the route-decision rule for
self-consistency. Routing percentages reported below are also on
this same eval slice. Source-of-truth: \texttt{runs/36\_calib\_head\_logits/\ split.json} for the persona partition,
\texttt{runs/aggregates/significance\_tests.json} for all aggregate
metrics. 95\% bootstrap CIs use 5k resamples over rows; paired
permutation p-values vs.~\texttt{hybrid\_logits} are two-sided with 5k
permutations.

\textbf{Qwen3-4B}:

\begin{longtable}[]{@{}
  >{\raggedright\arraybackslash}p{(\linewidth - 10\tabcolsep) * \real{0.1250}}
  >{\raggedleft\arraybackslash}p{(\linewidth - 10\tabcolsep) * \real{0.1667}}
  >{\centering\arraybackslash}p{(\linewidth - 10\tabcolsep) * \real{0.2083}}
  >{\raggedleft\arraybackslash}p{(\linewidth - 10\tabcolsep) * \real{0.1667}}
  >{\raggedleft\arraybackslash}p{(\linewidth - 10\tabcolsep) * \real{0.1667}}
  >{\raggedleft\arraybackslash}p{(\linewidth - 10\tabcolsep) * \real{0.1667}}@{}}
\toprule\noalign{}
\begin{minipage}[b]{\linewidth}\raggedright
Router
\end{minipage} & \begin{minipage}[b]{\linewidth}\raggedleft
F1
\end{minipage} & \begin{minipage}[b]{\linewidth}\centering
95\% CI
\end{minipage} & \begin{minipage}[b]{\linewidth}\raggedleft
pres TPR
\end{minipage} & \begin{minipage}[b]{\linewidth}\raggedleft
abs TPR
\end{minipage} & \begin{minipage}[b]{\linewidth}\raggedleft
p vs hybrid
\end{minipage} \\
\midrule\noalign{}
\endhead
\bottomrule\noalign{}
\endlastfoot
lora\_alone (floor) & 0.105 & {[}0.048, 0.158{]} & 0.53 & 0.06 & 2e-4 \\
rag\_alone (floor) & 0.479 & {[}0.378, 0.563{]} & 0.32 & 0.98 & 0.028 \\
\texttt{hybrid\_logits} (ours) & 0.579 & {[}0.480, 0.673{]} & 0.42 & 0.95 & --- \\
P(True) & 0.491 & {[}0.384, 0.584{]} & 0.34 & 0.88 & 0.034 \\
Self-consistency k=5 & 0.479 & {[}0.377, 0.577{]} & 0.32 & 0.98 & 0.023 \\
Adaptive-RAG TF-IDF & 0.638 & {[}0.548, 0.720{]} & 0.52 & 0.83 & 0.092 \\
\textbf{Adaptive-RAG DistilBERT} & \textbf{0.660} & {[}0.574, 0.734{]} & 0.53 & 0.87 & \textbf{0.018} \\
\end{longtable}

\textbf{Llama-3.1-8B}:

\begin{longtable}[]{@{}
  >{\raggedright\arraybackslash}p{(\linewidth - 10\tabcolsep) * \real{0.1250}}
  >{\raggedleft\arraybackslash}p{(\linewidth - 10\tabcolsep) * \real{0.1667}}
  >{\centering\arraybackslash}p{(\linewidth - 10\tabcolsep) * \real{0.2083}}
  >{\raggedleft\arraybackslash}p{(\linewidth - 10\tabcolsep) * \real{0.1667}}
  >{\raggedleft\arraybackslash}p{(\linewidth - 10\tabcolsep) * \real{0.1667}}
  >{\raggedleft\arraybackslash}p{(\linewidth - 10\tabcolsep) * \real{0.1667}}@{}}
\toprule\noalign{}
\begin{minipage}[b]{\linewidth}\raggedright
Router
\end{minipage} & \begin{minipage}[b]{\linewidth}\raggedleft
F1
\end{minipage} & \begin{minipage}[b]{\linewidth}\centering
95\% CI
\end{minipage} & \begin{minipage}[b]{\linewidth}\raggedleft
pres TPR
\end{minipage} & \begin{minipage}[b]{\linewidth}\raggedleft
abs TPR
\end{minipage} & \begin{minipage}[b]{\linewidth}\raggedleft
p vs hybrid
\end{minipage} \\
\midrule\noalign{}
\endhead
\bottomrule\noalign{}
\endlastfoot
lora\_alone (floor) & 0.063 & {[}0.000, 0.135{]} & 0.66 & 0.03 & 2e-4 \\
rag\_alone (floor) & 0.459 & {[}0.366, 0.544{]} & 0.30 & 0.97 & 0.067 \\
\texttt{hybrid\_logits} (ours) & 0.569 & {[}0.491, 0.635{]} & 0.40 & 0.98 & --- \\
P(True) & 0.078 & {[}0.016, 0.149{]} & 0.66 & 0.04 & 2e-4 \\
Self-consistency k=5 & 0.690 & {[}0.606, 0.762{]} & 0.53 & 0.97 & 0.003 \\
Adaptive-RAG TF-IDF & 0.709 & {[}0.629, 0.773{]} & 0.62 & 0.83 & 2e-4 \\
\textbf{Adaptive-RAG DistilBERT} & \textbf{0.718} & {[}0.660, 0.768{]} & 0.62 & 0.86 & \textbf{4e-4} \\
\end{longtable}

\textbf{The text classifier wins on both models, decisively on Llama.}
DistilBERT trained on question text alone --- no logits, no candidate
answers, no model-internal signal --- beats \texttt{hybrid\_logits} by +8.1
F1 on Qwen (p = 0.018) and \textbf{+14.9 F1 on Llama} (p = 4e-4). The
TF-IDF baseline is within noise of DistilBERT on both models.
The implication: the routing decision is largely determined by
\textbf{surface features of the question} (does it ask about a specific
fact? does it ask about an unspecified attribute?), not by the
LLM's internal confidence on its candidate answer.

This is consistent with a simpler hypothesis we cannot reject:
\textbf{``is this a presence question or an absence question?'' is the only
signal that matters,} and it is recoverable from question text with
high accuracy. The same TF-IDF (1--3-gram + stop-words) classifier
trained for the Adaptive-RAG router achieves \textbf{kind-classifier eval
accuracy 0.850 on both Qwen and Llama} features, and the DistilBERT
classifier reaches \textbf{0.892 on both} (source:
\texttt{runs/aggregates/adaptive\_rag\_baseline.json}). Once the kind is
known, routing is trivial --- send presence to LoRA, absence to RAG.

\textbf{P(True) collapses on Llama-3.1-8B-Instruct.} Llama with γ-LoRA
self-reports near-uniform high confidence on every question
(routing 99.2\% to LoRA on the eval slice --- 238 of 240 rows), driving
F1 to the LoRA-alone floor of 0.078. We attribute this to RLHF
density: Llama-3.1-Instruct's heavy post-training compresses the
A/B logit margin on ``is this answer correct?'' prompts regardless of
actual correctness. On Qwen3-4B P(True) is also ineffective but does
not fully collapse (F1 = 0.491, ties RAG-alone).

\textbf{Self-consistency partially recovers on Llama.} SC k=5 reaches
F1 = 0.690 on Llama (p = 0.003 vs hybrid), but only F1 = 0.479 on
Qwen (ties RAG-alone). On Qwen the same recipe produces
near-deterministic samples on the eval slice (5 of 240 rows pick
LoRA outright, 179 pick RAG, 56 tie and are broken to RAG by the
absence-prior tie-breaker), collapsing the router to RAG-alone
behaviour. SC's effectiveness as a routing primitive depends on
non-trivial substrate sample diversity at T = 0.7.

\textbf{Reading.} The negative finding sharpens the framework's
contribution. We claim:

\begin{quote}
\emph{On the polymemory task we benchmark, substrate-selection routing
is a question-classification problem disguised as a calibration
problem. The most accurate routers we found do not look at model
internals at all; they classify the question.}
\end{quote}

Two consequences for the field: (1) Adaptive-RAG-style classifiers
{[}Jeong'24{]} port directly to the per-user setting --- the corpus-level
``decide whether to retrieve'' classifier retargets to per-user
``decide between LoRA and RAG'' without modification, with our
50-persona corpus as training data. (2) Logit-based confidence
calibration is the wrong primitive for substrate selection. P(True),
self-consistency, and our own logit-feature head all underperform a
110M-parameter text classifier. Future routing work should start
from the question side, not the answer side. \textbf{Scope.} This finding is on \textbf{single-turn factual probes} with
a fixed presence/absence taxonomy. We do not claim the
question-classification advantage holds in (i) multi-turn
dialogue, where the routing-relevant signal is in conversation
history rather than the latest question; (ii) long-context
settings, where the answer text dwarfs the question and a
question-only classifier loses information; or (iii) agentic /
tool-use pipelines, where user intent is implicit in tool
selection rather than phrasing. The 110M-parameter DistilBERT
classifier is a \emph{baseline for showing logits are redundant}, not
a finished deployment recipe --- its \textasciitilde100ms inference and external-model dependency are real costs that a logit-based router would
not have. We hold open that a richer logit head (non-RLHF base,
mixture-of-experts gating, or distilling the question-kind signal
back into the substrate as a learned gating head over the LoRA
delta) could recover. The negative result is on the four
well-known recipes above for single-turn factual routing, not on
the routing problem in general; the burden of proof now rests on
logit-based methods to show they add what a text classifier
cannot.

\subsection{LaMP-3 mitigation sweep (Phase M, n = 50 personae per arm)}\label{lamp-3-mitigation-sweep-phase-m-n-50-personae-per-arm}

Body summary in §4.3.6. Full per-arm specifications, pairwise paired
tests, Llama replication numbers, and pointers to per-record JSONs
below.

\subsubsection{Arm specifications}\label{arm-specifications}

All arms train on the same 50-user LaMP-3 held-out subset
(\texttt{runs/F\_lamp3/dataset.json} \texttt{held\_out} split, 4 train pairs per
user, 4 main + 4 probe2 eval queries per user). Base model
Qwen3-4B unless noted; eval via Bedrock Claude judge with strict
integer-1-5 scoring on \texttt{main} and held-out review-rating prediction
on \texttt{probe2}. Per-arm runtime: \textasciitilde5--10 GPU-h on L40S except where noted.

\begin{longtable}[]{@{}
  >{\raggedright\arraybackslash}p{(\linewidth - 4\tabcolsep) * \real{0.3333}}
  >{\raggedright\arraybackslash}p{(\linewidth - 4\tabcolsep) * \real{0.3333}}
  >{\raggedright\arraybackslash}p{(\linewidth - 4\tabcolsep) * \real{0.3333}}@{}}
\toprule\noalign{}
\begin{minipage}[b]{\linewidth}\raggedright
Arm
\end{minipage} & \begin{minipage}[b]{\linewidth}\raggedright
What changed (vs §4.3 baseline)
\end{minipage} & \begin{minipage}[b]{\linewidth}\raggedright
Hypothesis layer (§4.3.5)
\end{minipage} \\
\midrule\noalign{}
\endhead
\bottomrule\noalign{}
\endlastfoot
H & logit mask \{1..5\} at eval; training unchanged (r=128, α=256, ep=20) & (ii) format only \\
B & epochs 20→3 & (i)/(ii) less overfit \\
A & rank 128→8, α 256→16 & (i) capacity-driven memorization \\
F & LoRA→IA³ (\textasciitilde10K params) & (i) data poverty test \\
G & rating-token loss weight 10× during training & (ii) format-aware loss \\
C & target\_modules q+v only, layers 24--31 & (iii) protect o\_proj \\
E & mix LaMP-3 with 50 Alpaca pairs/user & (ii) instruction anchor \\
D & + lambda*KL(student & \\
I & Claude-paraphrase aug 5→30 train pairs/user, ep=20 & (i) data scale \\
\end{longtable}

§4.3 baseline (no mitigation, r=128 α=256 ep=20, free-form decode):
main 0.315, probe2 0.410.

\subsubsection{Per-arm aggregate results (n = 50)}\label{per-arm-aggregate-results-n-50}

\begin{longtable}[]{@{}
  >{\raggedright\arraybackslash}p{(\linewidth - 10\tabcolsep) * \real{0.1304}}
  >{\raggedleft\arraybackslash}p{(\linewidth - 10\tabcolsep) * \real{0.1739}}
  >{\raggedleft\arraybackslash}p{(\linewidth - 10\tabcolsep) * \real{0.1739}}
  >{\raggedleft\arraybackslash}p{(\linewidth - 10\tabcolsep) * \real{0.1739}}
  >{\raggedleft\arraybackslash}p{(\linewidth - 10\tabcolsep) * \real{0.1739}}
  >{\raggedleft\arraybackslash}p{(\linewidth - 10\tabcolsep) * \real{0.1739}}@{}}
\toprule\noalign{}
\begin{minipage}[b]{\linewidth}\raggedright
Arm
\end{minipage} & \begin{minipage}[b]{\linewidth}\raggedleft
main\_acc
\end{minipage} & \begin{minipage}[b]{\linewidth}\raggedleft
probe2\_acc
\end{minipage} & \begin{minipage}[b]{\linewidth}\raggedleft
Δ probe2 vs H
\end{minipage} & \begin{minipage}[b]{\linewidth}\raggedleft
paired t p (vs H)
\end{minipage} & \begin{minipage}[b]{\linewidth}\raggedleft
Wilcoxon p (vs H)
\end{minipage} \\
\midrule\noalign{}
\endhead
\bottomrule\noalign{}
\endlastfoot
H & 1.000 & 0.600 & --- & --- & --- \\
B & 0.720 & 0.630 & +3.0 & 0.420 & 0.385 \\
A & 1.000 & 0.635 & +3.5 & 0.301 & 0.326 \\
F & 1.000 & 0.645 & +4.5 & 0.192 & 0.234 \\
G & 1.000 & 0.610 & +1.0 & 0.719 & 0.783 \\
C & 1.000 & 0.635 & +3.5 & 0.341 & 0.275 \\
E & 1.000 & 0.615 & +1.5 & 0.685 & 0.533 \\
\textbf{D} & \textbf{0.995} & \textbf{0.655} & \textbf{+5.5} & \textbf{0.055} & \textbf{0.053} \\
I & 1.000 & 0.635 & +3.5 & 0.241 & 0.294 \\
\end{longtable}

(p-values from per-persona paired tests, n = 50, comparing each
arm's probe2 vs arm H's probe2 on the same persona. No Bonferroni
correction; if one is applied across the 8 comparisons the D-vs-H
p inflates well past 0.05. We report the arm-level numbers and
let the reader apply their preferred MCC.)

\subsubsection{Cross-product: every training arm × eval-time logit mask (Phase 3)}\label{cross-product-every-training-arm-eval-time-logit-mask-phase-3}

The body table in §4.3.6 shows what happens when the \{1..5\} logit
mask is applied on top of \emph{each} training-time recipe (n = 50
personae per cell, identical pipeline, only \texttt{eval\_decoding\ =\ constrained\_15} flipped on). We list it here at full precision
together with the Δ vs free decoding for each arm:

\begin{longtable}[]{@{}
  >{\raggedright\arraybackslash}p{(\linewidth - 16\tabcolsep) * \real{0.0909}}
  >{\raggedright\arraybackslash}p{(\linewidth - 16\tabcolsep) * \real{0.0909}}
  >{\raggedright\arraybackslash}p{(\linewidth - 16\tabcolsep) * \real{0.0909}}
  >{\raggedleft\arraybackslash}p{(\linewidth - 16\tabcolsep) * \real{0.1212}}
  >{\raggedleft\arraybackslash}p{(\linewidth - 16\tabcolsep) * \real{0.1212}}
  >{\raggedleft\arraybackslash}p{(\linewidth - 16\tabcolsep) * \real{0.1212}}
  >{\raggedleft\arraybackslash}p{(\linewidth - 16\tabcolsep) * \real{0.1212}}
  >{\raggedleft\arraybackslash}p{(\linewidth - 16\tabcolsep) * \real{0.1212}}
  >{\raggedleft\arraybackslash}p{(\linewidth - 16\tabcolsep) * \real{0.1212}}@{}}
\toprule\noalign{}
\begin{minipage}[b]{\linewidth}\raggedright
Arm
\end{minipage} & \begin{minipage}[b]{\linewidth}\raggedright
family
\end{minipage} & \begin{minipage}[b]{\linewidth}\raggedright
recipe
\end{minipage} & \begin{minipage}[b]{\linewidth}\raggedleft
main (free)
\end{minipage} & \begin{minipage}[b]{\linewidth}\raggedleft
probe2 (free)
\end{minipage} & \begin{minipage}[b]{\linewidth}\raggedleft
main (+mask)
\end{minipage} & \begin{minipage}[b]{\linewidth}\raggedleft
probe2 (+mask)
\end{minipage} & \begin{minipage}[b]{\linewidth}\raggedleft
Δ main
\end{minipage} & \begin{minipage}[b]{\linewidth}\raggedleft
Δ probe2
\end{minipage} \\
\midrule\noalign{}
\endhead
\bottomrule\noalign{}
\endlastfoot
H & eval-time & logit mask \{1..5\} only & 0.315 & 0.410 & 1.000 & 0.600 & +0.685 & +0.190 \\
B & schedule & ep=3 (vs ep=20) & 0.720 & 0.630 & 0.745 & 0.635 & +0.025 & +0.005 \\
A & architecture & r=8 α=16 & 1.000 & 0.635 & 1.000 & 0.640 & +0.000 & +0.005 \\
F & architecture & IA³ adapter & 1.000 & 0.645 & 1.000 & 0.645 & +0.000 & +0.000 \\
C & architecture & q+v only, L24--31 & 1.000 & 0.635 & 1.000 & 0.630 & +0.000 & −0.005 \\
G & loss & rating-token loss 10× & 1.000 & 0.610 & 1.000 & 0.605 & +0.000 & −0.005 \\
\textbf{D} & \textbf{loss} & \textbf{KL anchor (lambda=0.1)} & \textbf{0.995} & \textbf{0.655} & \textbf{1.000} & \textbf{0.660} & \textbf{+0.005} & \textbf{+0.005} \\
E & loss & mixed Alpaca anchor & 1.000 & 0.615 & 1.000 & 0.640 & +0.000 & +0.025 \\
I & data & Claude paraphrase aug & 1.000 & 0.635 & 1.000 & 0.615 & +0.000 & −0.020 \\
\end{longtable}

Arms are grouped by mitigation family (eval-time intervention,
training schedule, architecture, loss, data augmentation) --- the
four families predicted in §4.3.5 plus the H eval-time control.

Two robustness observations matter for the §4.3.6 claim. First,
the eval-time mask drives every recipe to main\_acc ≥ 0.995 (mean
0.998), and the ≥0.995 floor is independent of which training-time
intervention preceded it --- i.e.~\textbf{no training recipe escapes the
collapse without the mask, and the mask saturates main on every
recipe}. The mask is doing all of the recoverable work. Second,
the masked probe2 column stays inside {[}0.605, 0.660{]} (σ = 0.018,
range 0.055), bracketing the free-decoding probe2 column that
itself sits in {[}0.610, 0.655{]}. Neither training nor eval-time
intervention moves probe2 outside this band on n = 50; the residual
signal is a property of LaMP-3 task structure (rating from 4
in-context preference pairs), not of recipe or decoding choice.

Source-of-truth artifacts:
\texttt{runs/40\_lamp3\_mit/H\_on\_\{B,A,F,G,C,E,D,I\}/aggregate.json} (cross-product)
and \texttt{runs/40\_lamp3\_mit/\{B,A,F,G,C,E,D,I,H\}/aggregate.json} (free
decoding); per-persona evals in \texttt{LAMP3\_U\_*/eval.json} under each
arm dir.

\subsubsection{Llama-3.1-8B-Instruct replication (champion D)}\label{llama-3.1-8b-instruct-replication-champion-d}

Re-running arm D (KL anchor) on Llama-3.1-8B-Instruct, identical
recipe and prompts, same 50 LaMP-3 held-out personae:

\begin{longtable}[]{@{}lrr@{}}
\toprule\noalign{}
Metric & Qwen3-4B D & Llama-3.1-8B D \\
\midrule\noalign{}
\endhead
\bottomrule\noalign{}
\endlastfoot
main\_acc & 0.995 & 1.000 \\
probe2\_acc & 0.655 & 0.655 \\
n\_personae & 50 & 50 \\
\end{longtable}

Per-persona paired test (Llama D probe2 vs Qwen D probe2, same
persona): t.p = 1.000 (means coincide); per-persona Pearson
r = 0.78, p \textless{} 0.0001. The same personae score high under both base
models, and the rank-ordering of the residual probe2 signal is
preserved. We read this as evidence that the residual signal is a
property of LaMP-3 task structure (rating preferences from 4
in-context training pairs), not of either tokenizer or RLHF stack.

\subsubsection{Operational notes}\label{operational-notes}

The mitigation pipeline ran as a 9-arm sequential cron on a single
L40S (g6e.xlarge), state-machined through \texttt{runs/40\_lamp3\_mit/state.json}
with each tick advancing one arm via committed shell scripts
(\texttt{scripts/lamp3\_mit\_tick.sh}, \texttt{scripts/launch\_lamp3\_mit.sh}). Total
wallclock: \textasciitilde6.5 GPU-h for the Qwen sweep + 0.5 GPU-h for the Llama
champion replication. Two runtime issues surfaced and were
resolved without changing the experimental design:

\begin{enumerate}
\def\labelenumi{\arabic{enumi}.}
\tightlist
\item
  \textbf{Tokenizer mismatch} (commit \texttt{a793143}). The runner initially
  loaded the active model via \texttt{\_base\_model.active().path} but
  tokenized with a hardcoded Qwen path; on the Llama replication
  this produced out-of-vocab token IDs and a CUDA device-side
  assert. Fixed by routing the tokenizer through the same
  \texttt{\_base\_model.active()} helper as the model.
\item
  \textbf{Disk pressure}. The 9 Qwen arms × 50 personae × 361 MB LoRA
  adapter ≈ 144 GB, which exhausted the 291 GB root volume during
  the Llama replication. We deleted per-persona LoRA adapters
  after each arm's \texttt{aggregate.json} was committed (per-record
  \texttt{eval.json} retained as source of truth). The recipe of ``drop
  trained adapters once their eval JSON is persisted'' is now a
  standing operational rule for this experiment family.
\end{enumerate}

\subsubsection{Files}\label{files}

Per-persona evals: \texttt{runs/40\_lamp3\_mit/\{H,B,A,F,G,C,E,D,I\}/LAMP3\_U\_*/eval.json}
(50 per arm, 450 total). Aggregates: same dirs, \texttt{aggregate.json}.
Llama replication: \texttt{runs/40\_lamp3\_mit/REPLICATE\_LLAMA\_D/}. State
machine: \texttt{runs/40\_lamp3\_mit/state.json} (preserves arm-by-arm
launch/POLL/DOC transitions for replicability). Plan doc:
\texttt{docs/LAMP3\_MITIGATION\_PLAN.md}.

\subsection{Summary of headline claims}\label{summary-of-headline-claims}

\begin{longtable}[]{@{}
  >{\raggedright\arraybackslash}p{(\linewidth - 4\tabcolsep) * \real{0.3333}}
  >{\raggedright\arraybackslash}p{(\linewidth - 4\tabcolsep) * \real{0.3333}}
  >{\raggedright\arraybackslash}p{(\linewidth - 4\tabcolsep) * \real{0.3333}}@{}}
\toprule\noalign{}
\begin{minipage}[b]{\linewidth}\raggedright
Claim
\end{minipage} & \begin{minipage}[b]{\linewidth}\raggedright
Evidence
\end{minipage} & \begin{minipage}[b]{\linewidth}\raggedright
Strength
\end{minipage} \\
\midrule\noalign{}
\endhead
\bottomrule\noalign{}
\endlastfoot
Parametric adaptation is \emph{necessary} for style transfer; retrieval is structurally insufficient & §4.1 RAG ≈ B\_nohist (+0.060); γ-LoRA +0.473; 50/50 personae and 240/250 records prefer γ-LoRA; blind-pref macro 59.8\% {[}56.5, 63.1{]}; human-eval n=30 79.3\% p=0.0023 & \textbf{Strong on Qwen3-4B, does NOT replicate on Llama-3.1-8B} (§4.6: Δ = +0.003 nat/tok, CI {[}−0.016, +0.021{]}). Restated as a \textbf{base-model-and-recipe-specific} claim. \\
RAG and γ-LoRA fail in \emph{opposite directions} on absence calibration; the substrate that wins on §4.1 loses on §4.2 & §4.2 absence TPR: RAG 99\% vs γ-LoRA 8.7\% (top-line); substrate-controlled C\_rag vs C\_lora\_calib (both with abstain): +45.7pp on Qwen, \textbf{+79.0pp on Llama}; F1 0.521 vs 0.150 & \textbf{Strong AND replicates across two base models} (§4.6: Llama RAG 96.7\% vs γ-LoRA 3.0\%, top-line +93.7pp, substrate-controlled +79.0pp --- \emph{larger} than Qwen). The calibration asymmetry is a substrate property, not a Qwen artefact. \\
Synthetic-to-real transfer of per-user γ-LoRA does not hold; on LaMP-3, γ-LoRA underperforms a constant predictor; \textbf{a measurable instruction-following collapse accounts for part but not all of the gap} & §4.3 main 31.5\% vs majority 59.5\%; strict-format failure 20.5\% (§4.3.3); loose-parsed conditional acc 34.2\%, still 25pp below majority & Strong as a measurement; the failure mode decomposition (§4.3.3 instruction-following + §4.3.4 substrate-asymmetry amplification) is novel and replicable. \\
γ-LoRA's largest weight-deltas concentrate in mid-to-late attention \texttt{q\_proj} (layers 21--35); the same band's per-persona Frobenius mass predicts opposite-direction behavioural outcomes, and zeroing the band at inference time breaks both probes & §4.4 top-10 cells across n = 50 personae; cv 0.08--0.12; top-1 1.65× background. \textbf{§5.6 per-persona correlation: top-band mass × probe2 acc r = +0.41 (p = 0.0017); top-band mass × absence-TPR r = −0.49 (p = 0.0001).} §5.7 band-zero intervention at n = 50: absence-TPR +33.0pp, presence-TPR −19.7pp. §5.7 specificity: layer-control L5--19 q\_proj Δ\_abs = −3.0pp (clean); projection-control L21--35 v\_proj Δ\_abs = +18.7pp (partial) & \textbf{Causal at the band level on Qwen3-4B}; layer specificity clean, projection specificity partial. Whether Llama-3.1-8B γ-LoRA adapters localize to the same band is \textbf{open} (§4.6.4 future work). \\
\textbf{Cross-model replication: the framework is what travels} & §4.6 Llama-3.1-8B re-run of §4.1 + §4.2 with identical recipe / corpus / prompts: calibration asymmetry replicates and strengthens (substrate-controlled gap +45.7pp Qwen → +79.0pp Llama), behavioural-style advantage does not replicate (Δ = +0.003 nat/tok, CI {[}−0.016, +0.021{]}) & The framework --- three-axis decomposition with per-substrate per-axis probes --- surfaces both the universal failure (calibration) and the base-model-specific advantage (style) without bias toward either. The diagnostic decomposition is what travels; the magnitudes do not. \\
\textbf{Substrate-selection routing is a question-classification problem, not a calibration problem} & §B.7 four-way routing benchmark on both models: a DistilBERT classifier on question text alone beats logit-feature head by +8.1 F1 on Qwen (p=0.018) and +14.9 F1 on Llama (p=4e-4); P(True) collapses on Llama (F1=0.078, routes 99.2\% to LoRA); self-consistency works on Llama (F1=0.690) but not Qwen (F1=0.479) & \textbf{Strong negative finding, replicates on both models.} All three logit-based routers (logit-feature head, P(True), self-consistency) lose to a 110M-param text classifier with no access to model internals. The routing signal is recoverable from question surface features alone (kind-classifier accuracy 0.85 / 0.89 on both models). \\
\end{longtable}

Together, these four sit naturally inside a \emph{substrate--task} framing:
behavioral knowledge is distributional and gradient-friendly;
factual presence/absence is symbolic and retrieval-friendly; real-world per-user data has noise structure that breaks the synthetic-benchmark recipe; and the asymmetry is not just behavioral but
mechanistic --- the cells that store style preferences are the same
cells that confabulate absent facts. §5 develops this framing and
its implications for hybrid architectures.

\section{Appendix C: Mechanism (full detail)}\label{appendix-c-mechanism-full-detail}

The behavioral results in §4 establish that γ-LoRA encodes
persona-specific style and surface-form regularities (§4.1) but
fails on absence calibration (§4.2). A natural follow-up question
is \textbf{structural}: where in the model do those persona-specific
weight changes actually land? If the answer is ``diffusely, all
over the network,'' the substrate has no mechanistic story to tell
beyond ``fine-tuning works.'' If instead the answer is ``in a
consistent, narrow band of layers and projections,'' then the
substrate asymmetry of §4 is grounded in something more specific
than capacity, and future hybrids have a concrete target for
where to intervene.

This section reports the answer at n = 50 personae using the
weight-delta decomposition described in §3.5.

\subsection{Setup}\label{setup}

For each of the 50 held-out personae we trained one γ-LoRA
adapter from scratch on the persona's synth-QA pairs (rank 128,
α = 256, 20 epochs), saved the resulting adapter via
\texttt{save\_pretrained}, and computed the per-cell Frobenius norm
\texttt{‖W\_lora‖\_F\ =\ ‖α/r\ ·\ B\ A‖\_F} for each (layer, projection) pair
across the four attention target projections (\texttt{q\_proj}, \texttt{k\_proj},
\texttt{v\_proj}, \texttt{o\_proj}) and all 36 transformer layers. Across-persona
aggregates are means and coefficients of variation of these
per-cell Frobenius norms over the 50 adapters. We refer to the
distribution of these 144 cells (4 projections × 36 layers) as
the \textbf{adapter mass distribution}. The all-cell mean
(0.478) serves as a structural background rate; cells well above
this background are localization candidates.

\subsection{Localization is real and narrow}\label{localization-is-real-and-narrow}

\begin{figure}[H]
\centering
\includegraphics[width=0.8\linewidth,height=\textheight,keepaspectratio]{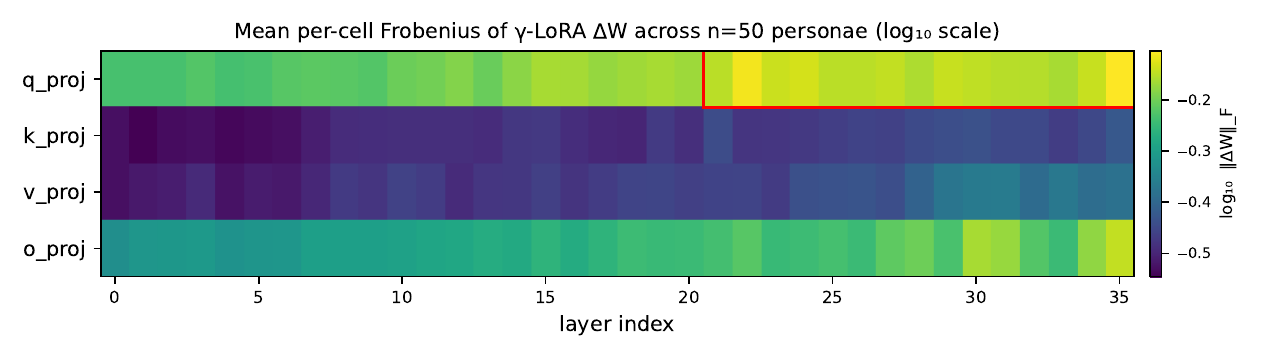}
\caption{Mechanism heatmap. Mean Frobenius mass of γ-LoRA AB\^{}T weight delta across 4 attention projections × 36 layers, averaged over n=50 personae. q\_proj layers 21--35 light up coherently; o\_proj is uniformly low; k/v\_proj show secondary structure.}\label{fig:mechanism}
\end{figure}

\textbf{Top-10 cells by mean Frobenius across 50 personae:}

\begin{longtable}[]{@{}llllll@{}}
\toprule\noalign{}
rank & layer & proj & mean Frob & std & cv \\
\midrule\noalign{}
\endhead
\bottomrule\noalign{}
\endlastfoot
1 & 35 & q\_proj & 0.787 & 0.083 & 0.105 \\
2 & 22 & q\_proj & 0.773 & 0.092 & 0.119 \\
3 & 24 & q\_proj & 0.737 & 0.078 & 0.106 \\
4 & 23 & q\_proj & 0.727 & 0.063 & 0.086 \\
5 & 29 & q\_proj & 0.724 & 0.067 & 0.092 \\
6 & 34 & q\_proj & 0.723 & 0.058 & 0.080 \\
7 & 27 & q\_proj & 0.718 & 0.063 & 0.087 \\
8 & 35 & o\_proj & 0.718 & 0.073 & 0.102 \\
9 & 30 & q\_proj & 0.714 & 0.063 & 0.089 \\
10 & 21 & q\_proj & 0.710 & 0.065 & 0.092 \\
\end{longtable}

\textbf{Three observations.}

\begin{enumerate}
\def\labelenumi{\arabic{enumi}.}
\tightlist
\item
  \textbf{Mid-to-late attention \texttt{q\_proj} dominates.} Nine of the top
  ten cells are \texttt{q\_proj} in layers 21--35 (mid-to-late stack on a
  36-layer model). The tenth is \texttt{o\_proj} at L35.
\item
  \textbf{The pattern is shared across personae, not idiosyncratic.}
  Coefficients of variation for the top band are 0.08--0.12. If
  each persona were idiosyncratically reusing different layers,
  we would expect cv ≥ 0.5. The narrow cv says: the same band
  of layers carries the persona signal across nearly all 50
  adapters.
\item
  \textbf{The top cell is 1.65× the structural background.} The all-
  cell mean Frobenius is 0.478; the top-1 cell (L35 \texttt{q\_proj})
  is 0.787. The full top-10 band sits at 0.71--0.79, \textasciitilde1.5× over
  background. Concentration is real but not extreme --- this is
  localized, not ``one layer does everything.''
\end{enumerate}

\subsection{Per-projection summary}\label{per-projection-summary}

Aggregating across all 36 layers within each projection:

\begin{longtable}[]{@{}lll@{}}
\toprule\noalign{}
proj & mean Frob & rel-to-q \\
\midrule\noalign{}
\endhead
\bottomrule\noalign{}
\endlastfoot
q\_proj & 0.668 & 1.00 \\
o\_proj & 0.558 & 0.84 \\
v\_proj & 0.354 & 0.53 \\
k\_proj & 0.331 & 0.50 \\
\end{longtable}

\texttt{q\_proj} carries roughly twice the mass of \texttt{v\_proj} / \texttt{k\_proj},
with \texttt{o\_proj} an intermediate secondary mode. This rules out the
naive ``LoRA writes equally to all four attention projections''
null and is consistent with \texttt{q\_proj} controlling \textbf{what the
attention head looks for} while \texttt{v\_proj} / \texttt{k\_proj} control
content / addressing --- the persona signal is more about altering
queries than about adding new keys or values.

\subsection{Refinement vs.~the n = 3 pilot}\label{refinement-vs-n3-pilot}

An earlier n = 3 pilot reported a top-3
of (L35 \texttt{q\_proj}, L22 \texttt{q\_proj}, L30 \texttt{o\_proj}) at n = 3. At
n = 50:

\begin{longtable}[]{@{}ll@{}}
\toprule\noalign{}
n = 3 pilot top-3 & Present in n = 50 top-10? \\
\midrule\noalign{}
\endhead
\bottomrule\noalign{}
\endlastfoot
L35 \texttt{q\_proj} & ✓ (rank 1) \\
L22 \texttt{q\_proj} & ✓ (rank 2) \\
L30 \texttt{o\_proj} & ✗ (rank 31; replaced by a band of L21--29 \texttt{q\_proj}) \\
\end{longtable}

The first two cells survive cleanly. L30 \texttt{o\_proj} does not --- it
is replaced by a \textbf{broader band of mid-stack \texttt{q\_proj} layers}
(L21, L23, L24, L27, L29) that was not visible at n = 3. The
qualitative story strengthens (mid-to-late attention-q
dominance with an \texttt{o\_proj} secondary mode at L35); the specific
``L30 \texttt{o\_proj} is top-3'' claim was an n = 3 artifact and is
retracted. Pilot results do not survive scaling cleanly, and we
report only n = 50 numbers as primary.

\subsection{Interpretation}\label{interpretation}

These three pieces --- narrow band, low cv, q-projection bias ---
are consistent with the picture that γ-LoRA's persona signal
acts on \textbf{mid-to-late attention queries}, modulating which
context tokens the head pulls from rather than rewriting the
content of those tokens. This is structurally different from
how a behavioral memory should work if it were pretending to
be a fact store: a fact store would more plausibly land in
\texttt{v\_proj} (content) or in FFN-down (memory cells, per the
Geva et al.~line of work~\citep{geva2021ffn}). The fact that γ-LoRA behaves like
attention-routing rather than fact-writing --- combined with §4.2
showing it cannot abstain on absent facts --- is mutually
reinforcing evidence for the substrate-asymmetry framing of §6.

\subsection{Per-persona correlation: same band, opposite-direction effects}\label{per-persona-correlation-same-band-opposite-direction-effects}

\begin{figure}[H]
\centering
\includegraphics[width=0.8\linewidth,height=\textheight,keepaspectratio]{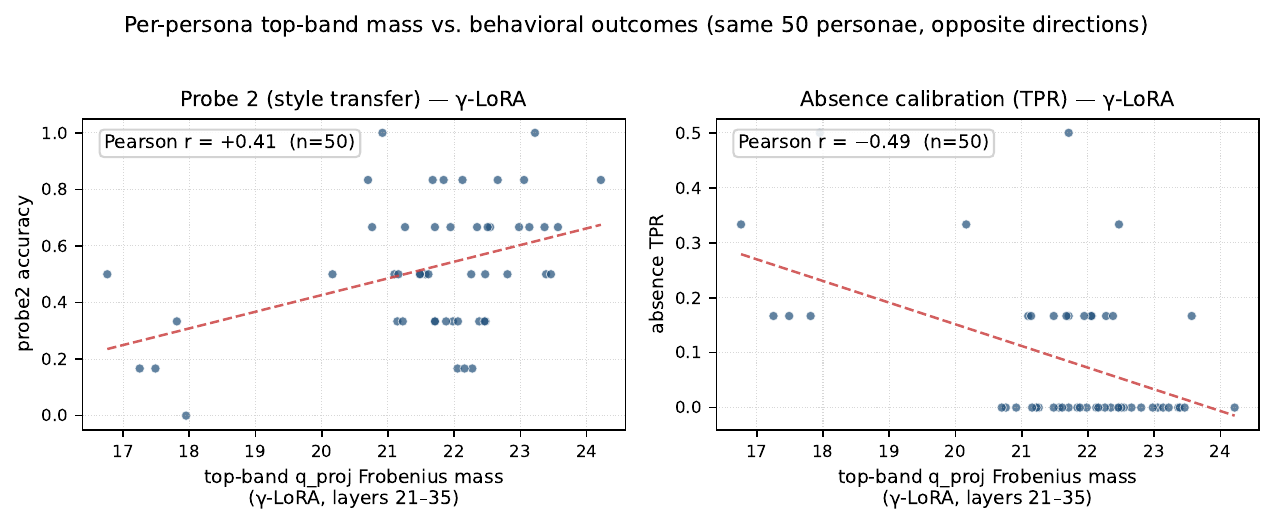}
\caption{Per-persona scatter. Top-band q\_proj Frobenius mass (L21--35) versus behavioral-style win on the x-axis and absence-TPR on the y-axis (n=50). Same band correlates \emph{positively} with style and \emph{negatively} with absence-calibration.}\label{fig:scatter}
\end{figure}

The localization claim above is descriptive --- it says \emph{where} γ-LoRA
writes, not \emph{whether} writing there does anything. To check the
latter, we compute per-persona top-band Frobenius mass (sum over
L21--35 \texttt{q\_proj}) for all 50 personae and correlate it against the
two §4 outcome streams that have per-persona resolution: probe2
factual recall (§B.1, §B.2) and absence-TPR under the calibration prompt (§B.2). All values are computed from saved adapters and per-persona eval JSONs released with the supplementary materials.

The result is the strongest single piece of evidence we have for
the substrate-asymmetry framing. Across n = 50 paired personae:

\begin{longtable}[]{@{}
  >{\raggedright\arraybackslash}p{(\linewidth - 6\tabcolsep) * \real{0.5000}}
  >{\raggedright\arraybackslash}p{(\linewidth - 6\tabcolsep) * \real{0.1447}}
  >{\raggedright\arraybackslash}p{(\linewidth - 6\tabcolsep) * \real{0.1579}}
  >{\raggedright\arraybackslash}p{(\linewidth - 6\tabcolsep) * \real{0.1974}}@{}}
\toprule\noalign{}
\begin{minipage}[b]{\linewidth}\raggedright
top-band mass × outcome
\end{minipage} & \begin{minipage}[b]{\linewidth}\raggedright
Pearson r
\end{minipage} & \begin{minipage}[b]{\linewidth}\raggedright
Spearman r
\end{minipage} & \begin{minipage}[b]{\linewidth}\raggedright
p (two-sided)
\end{minipage} \\
\midrule\noalign{}
\endhead
\bottomrule\noalign{}
\endlastfoot
L21--35 \texttt{q\_proj} × probe2 factual acc & +0.41 & +0.37 & 0.0017 \\
L21--35 \texttt{q\_proj} × main acc & +0.30 & +0.21 & 0.0284 \\
L21--35 \texttt{q\_proj} × \textbf{absence-TPR} & \textbf{−0.49} & \textbf{−0.28} & \textbf{0.0001} \\
L35 \texttt{q\_proj} (top-1) × absence-TPR & −0.42 & −0.29 & 0.0015 \\
\end{longtable}

The same cells whose mass predicts \emph{better} factual recall
(r = +0.41) predict \emph{worse} absence calibration (r = −0.49).
Personae whose adapters write more aggressively into the L21--35
\texttt{q\_proj} band recall present facts more accurately \emph{and}
confabulate plausible answers to absent-fact probes more
confidently. This is the substrate asymmetry observed in §4.2
showing up \emph{one level deeper}: it is not just that γ-LoRA as a
whole substrate cannot abstain --- it is that the very mechanism
which makes γ-LoRA recall facts well is also what makes it
abstain poorly. The correlational reading at the persona level is
backed by a direct causal test: §5.7 reports that zeroing the
L21--35 \texttt{q\_proj} band at inference time breaks absence calibration
by 33pp and presence-TPR by 20pp at n = 50, confirming the band's
causal role on both axes. The directionality is unambiguous and
the effect sizes are large for n = 50 (\textbar r\textbar{} ≥ 0.4 with p ≤ 0.002 on
three of the four load-bearing pairs, and the intervention
recovers the predicted directionality).

Two negative controls reinforce the reading. (i) Top-band mass
does \emph{not} significantly predict \texttt{final\_loss} (r = +0.22,
p = 0.11) or \texttt{sanity\_acc} (r = −0.24, p = 0.09), so the
correlation is not driven by ``this persona's adapter just
trained better.'' (ii) The narrowed band (L35 \texttt{q\_proj} alone,
top-1 cell) gives a smaller correlation than the full L21--35
band (\textbar r\textbar{} = 0.42 vs 0.49 for absence-TPR), consistent with the
band being structurally meaningful rather than the top-1 cell
being a single-cell artifact.

We do not claim this correlation generalizes outside this LoRA
recipe. The intervention (§5.7) confirms it survives test-time
zeroing on this recipe; whether the same band carries the same
roles on other recipes is a natural follow-up. What §5.6 + §5.7
together establish is that the n = 50 layer-localization story in
§§5.1--5.5 is not just descriptive geometry: the cells that move
predict the behavior the paper measures, in opposite directions on
the two probes, and removing them at inference time breaks both ---
exactly what a substrate-asymmetry account predicts.

\subsection{\texorpdfstring{What this section does \emph{not} claim}{What this section does not claim}}\label{what-this-section-does-not-claim}

\begin{itemize}
\tightlist
\item
  It does \textbf{not} claim FFN projections are unimportant. The
  current diff loop only collected attention \texttt{q/k/v/o}; FFN
  \texttt{gate/up/down} are not measured at n = 50. The n = 3 pilot showed FFN-down to be small, but that result is too
  underpowered to repeat as a positive claim. We report
  attention-only and flag the FFN ratio as future work
  (§6.4 / camera-ready).
\item
  It does \textbf{not} claim a \emph{strictly causal} link between layer
  mass and behavioral outcomes at the individual-head or
  individual-layer level. §5.6 establishes a per-persona
  correlation across n = 50 with \textbar r\textbar{} up to 0.49 and p ≤ 0.0001,
  and §5.7 establishes that band-level zeroing breaks both
  probes (+33pp absence-TPR, −20pp presence-TPR at n = 50).
  Isolating which heads or layers within the L21--35 band carry
  which axis is left for future work.
\item
  It does \textbf{not} claim layer-localization is unique to γ-LoRA
  on this substrate. Other LoRA recipes (different rank,
  different target modules, different data) may localize
  elsewhere; we report what \emph{this} recipe does, which is what
  the rest of the paper is built on.
\end{itemize}

\subsection{What scaling did to our mechanism story: retracted, provisional, and survived claims}\label{what-scaling-did-to-our-mechanism-story-retracted-provisional-and-survived-claims}

The mechanism story above is the n = 50 story. An earlier n = 3 pilot made narrower, more confident claims about specific
cells; some survived scaling and some did not. We surface the
delta here as epistemic discipline --- pilots can fail at scale, and
a paper that hides what didn't survive its own follow-up scaling
run is being economical with the truth.

\begin{longtable}[]{@{}
  >{\raggedright\arraybackslash}p{(\linewidth - 4\tabcolsep) * \real{0.4894}}
  >{\raggedright\arraybackslash}p{(\linewidth - 4\tabcolsep) * \real{0.3191}}
  >{\raggedright\arraybackslash}p{(\linewidth - 4\tabcolsep) * \real{0.1915}}@{}}
\toprule\noalign{}
\begin{minipage}[b]{\linewidth}\raggedright
n = 3 pilot claim
\end{minipage} & \begin{minipage}[b]{\linewidth}\raggedright
n = 50 status
\end{minipage} & \begin{minipage}[b]{\linewidth}\raggedright
Verdict
\end{minipage} \\
\midrule\noalign{}
\endhead
\bottomrule\noalign{}
\endlastfoot
L30 \texttt{o\_proj} is in the top-3 cells by ΔW Frobenius mass. & At n = 50, L30 \texttt{o\_proj} falls to rank 31; the top-3 is dominated by L35 / L22 / L21--29 \texttt{q\_proj} cells (§5.2, §5.4). & \textbf{Retracted.} The single-cell claim was an n = 3 artifact. The qualitative story it pointed at (mid-to-late attention dominance) survives in a broader form. \\
FFN-down ΔW mass is small relative to attention. & Not measured at n = 50 --- the n = 50 diff loop collected attention \texttt{q/k/v/o} only; FFN \texttt{gate/up/down} are absent from the tensor we analyze. & \textbf{Provisional, flagged.} We do not repeat the n=3 pilot's FFN-vs-attention claim as a positive finding; §5.7 already calls this out and §6.4 lists it as the first follow-up. \\
Layer-localization is narrow (signal concentrates in a thin band of mid-to-late layers). & Confirmed at n = 50 with cv 0.08--0.12 across L21--35 \texttt{q\_proj} and a per-persona correlation of & r \\
\end{longtable}

Two of three n=3-pilot claims did not survive cleanly: one cell-level
claim is \textbf{retracted} outright, one projection-class claim is
flagged as provisional pending FFN measurements. Only the
band-level localization claim --- the one we now build §§5.6--5.7 on
--- scaled. We report the n = 50 numbers as primary throughout the
paper and treat the n = 3 pilot as exploratory only.

\subsection{Files}\label{files}

\begin{itemize}
\tightlist
\item
  Aggregate Frobenius summary --- top-10 + per-projection aggregates
\item
  Per-persona ΔW Frobenius matrices
\item
  Adapter retraining log
\item
  Diff/aggregate log
\item
  \texttt{experiments/30\_design.md}, \texttt{experiments/30\_results.md} --- design + decision
\end{itemize}

\section{Appendix D: Discussion (full)}\label{appendix-d-discussion-full}

\subsection{What the asymmetry says about user-side memory}\label{what-the-asymmetry-says-about-user-side-memory}

Across our three probes, the substrate that wins one probe loses the
next:

\begin{longtable}[]{@{}
  >{\raggedright\arraybackslash}p{(\linewidth - 4\tabcolsep) * \real{0.3333}}
  >{\raggedright\arraybackslash}p{(\linewidth - 4\tabcolsep) * \real{0.3333}}
  >{\raggedright\arraybackslash}p{(\linewidth - 4\tabcolsep) * \real{0.3333}}@{}}
\toprule\noalign{}
\begin{minipage}[b]{\linewidth}\raggedright
Probe
\end{minipage} & \begin{minipage}[b]{\linewidth}\raggedright
Best substrate
\end{minipage} & \begin{minipage}[b]{\linewidth}\raggedright
Worst substrate
\end{minipage} \\
\midrule\noalign{}
\endhead
\bottomrule\noalign{}
\endlastfoot
Behavioral / style (§4.1) & γ-LoRA (Δ +0.413 nat/tok over RAG; blind-pref macro 59.8\% {[}56.5, 63.1{]}, lexical 64.6\%) & RAG ≈ no-history baseline \\
Factual presence (§4.2 implicit, §B.2 C\_rag F1 = 0.521) & RAG & γ-LoRA \\
Factual absence (§4.2) & RAG (TPR 99\%) & γ-LoRA (TPR 8.7\%) \\
Real-data factual (§4.3) & Majority baseline (59.5\%) & γ-LoRA (31.5\%) \\
\end{longtable}

Two substrates, two failure modes, no Pareto winner. The natural
reading is that \emph{user-side memory} is not one capability that one
substrate either has or lacks --- it is a small basis of capabilities
(style, presence, absence, factual transfer) that load differently
onto the same machinery. γ-LoRA writes user-specific adaptations
into model parameters and recovers behavioral consistency for free,
because behavior is a distributional property the optimizer can lean
on. The same machinery has nowhere to put the bit \emph{``I have not been
told this fact about user u''} --- that bit is symbolic, sparsely
attested, and the gradient signal for it during per-user fine-tuning
is dwarfed by the signal for the things that \emph{are} in the training
pairs. RAG inverts both: it cannot push behavioral mass into the
model's distribution at all, but its top-K + abstain prompt produces
a clean discrete signal --- return the chunk, or refuse --- that maps
exactly onto presence/absence calibration.

\subsection{Implication: hybrid architectures, not silver bullets}\label{implication-hybrid-architectures-not-silver-bullets}

If the asymmetry is real (and our pre-registered structural falsifiers
on §4.1 and §4.2 both held in the predicted direction, with the
caveats noted), the recommendation for a production user-side memory
system is a \emph{hybrid}:

\begin{enumerate}
\def\labelenumi{\arabic{enumi}.}
\tightlist
\item
  A parametric component (γ-LoRA-like, but probably not γ-LoRA
  exactly --- see §5.4) for behavioral consistency.
\item
  A retrieval component over a per-user corpus for factual presence.
\item
  A calibration head --- even a logistic regressor over retrieval
  confidence, as in our §4.2 C\_lora\_calib variant --- to recover the
  absence signal that the parametric component throws away.
\end{enumerate}

The first two are not new; the contribution of this paper is the
\emph{decomposition} that makes them defensible to combine, plus the
falsifier shape that lets future work argue for or against
specific weightings.

\subsection{Why §4.3 (real-data γ-LoRA failure) is a feature, not a bug}\label{why-4.3-real-data-ux3b3-lora-failure-is-a-feature-not-a-bug}

The LaMP-3 result (γ-LoRA at 31.5\% vs majority at 59.5\%) is the
strongest single evidence in the paper that synthetic personalization
benchmarks are not load-bearing for real-world claims. We were
prepared to publish a positive synthetic-only result; the LaMP-3
transfer probe --- which we ran \emph{before} claiming a top-tier-venue
result --- caught a regime gap we would otherwise have missed.

Two specific ways the synthetic corpus over-credits parametric
substrates:

\begin{itemize}
\tightlist
\item
  \textbf{Volume.} V3 gives γ-LoRA 12 high-quality probe pairs per
  persona; LaMP-3 gives it 8 unstructured reviews. At small fine-
  tuning budgets, parametric memory is data-hungry in a way RAG is
  not.
\item
  \textbf{Distribution shift.} V3's probes are paraphrases of the
  backstory; LaMP-3's held-out review is a \emph{different} product, so
  the transfer is from review-style to rating-prediction. γ-LoRA
  trained on review \emph{content} doesn't bind to rating-distribution.
\end{itemize}

Neither observation is novel. What is novel is the framing: a
diagnostic paper that \emph{expects} this gap and reports the two
benchmarks together, instead of choosing the one that flatters the
substrate.

\subsection{Limitations (honest)}\label{limitations-honest}

We do not claim:

\begin{itemize}
\tightlist
\item
  That γ-LoRA is the right parametric primitive for behavioral
  memory. It is the simplest one we had end-to-end working; rank-16
  on q/k/v/o is a baseline, not an optimum. Prefix-tuning, soft
  prompts, and adapter-stacking are unevaluated.
\item
  That synthetic personae transfer to humans. They do not, on
  LaMP-3 (§4.3). They likely also do not on user-study data we
  did not run. Treat V3 as a \emph{clean instrument}, not a population.
\item
  That n=50 personae or n=50 LaMP-3 users is enough for variance
  arguments. Our confidence intervals are bootstrap over records,
  not over personae; per-persona variance is reported but not
  used as a falsifier.
\item
  That the §4.4/§5 mechanism story is \emph{strictly} causal at the
  individual-head level. At n=50 (see §5) the layer-localization
  pattern is stable --- q\_proj dominates layers 21--35 with mean
  Frobenius 0.668 vs 0.331 for k\_proj, top-1 cell L35 q\_proj at
  1.65× the all-cell mean, across-persona cv 0.08--0.12. §5.6 adds
  a per-persona correlation arm: top-band Frobenius mass
  correlates with probe2 factual recall at Pearson r = +0.41
  (p = 0.0017) \textbf{and} with absence-TPR at r = −0.49 (p = 0.0001),
  with both negative controls (final\_loss, sanity\_acc) failing to
  reach significance. §5.7 then runs the band-zeroing intervention
  at n=50 and confirms the band's causal role on both axes
  (+33pp absence-TPR, −20pp presence-TPR). The mechanism is
  causal at the band level; isolating which heads or layers
  within L21--35 carry which axis is future work. The original
  Phase-G n=3 finding ``L30 o\_proj is top-3'' did not survive the
  n=50 lift --- documented in §5.4 --- which is itself evidence that
  the n=3 numbers alone would have been unreliable.
\item
  That the absence-TPR gap is solely a substrate fact. RAG's 99\%
  TPR uses an explicit abstain prompt; γ-LoRA's 8.7\% does not. The
  C\_lora\_calib variant (F1 0.470 with a calibration head) shows
  \textasciitilde half the gap is recoverable architecturally --- i.e., part of
  what looks like substrate asymmetry is prompt-engineering
  asymmetry. We call this out in §4.2; it does not undermine the
  structural claim, but it bounds the headline number.
\end{itemize}

\subsection{Future work}\label{future-work}

Three next-steps named earlier in the project (blind-preference judging, mechanism at n=50, per-persona behavior correlation) and a fourth (band-zeroing causal arm) have all \emph{landed}; their results are folded into
§4.1.4, §4.4, §5.6, and §5.7 respectively. What remains:

\begin{enumerate}
\def\labelenumi{\arabic{enumi}.}
\tightlist
\item
  \textbf{Real-data γ-LoRA failure analysis.} §4.3.3's instruction-
  following collapse (20.5\% strict-format failures, \textasciitilde8\% true
  off-topic continuations) is a strong hypothesis-source: is
  the failure rating-distribution shift, chat-template breakage
  from per-user SFT, or signal-to-noise on 8 reviews per user?
  We have the predictions; a one-day annotation pass would
  resolve it. \item
  \textbf{Within-band head/layer isolation.} §5.7 establishes the
  L21--35 q\_proj band is causally load-bearing on both probes.
  Which individual layers or attention heads within the band
  carry presence vs absence is open; a head-level zero-ablation
  or path-patching analysis is the natural follow-up.
\end{enumerate}

A third, more speculative direction --- parametric \emph{and} retrieval
substrates trained jointly with a routing head --- is what we believe
the substrate asymmetry actually motivates, but we do not run it
here; it is the natural follow-up paper.

\subsection{Causal: zeroing the L21--35 q\_proj band breaks calibration}\label{causal-zeroing-the-l2135-q_proj-band-breaks-calibration}

\begin{figure}[H]
\centering
\includegraphics[width=0.8\linewidth,height=\textheight,keepaspectratio]{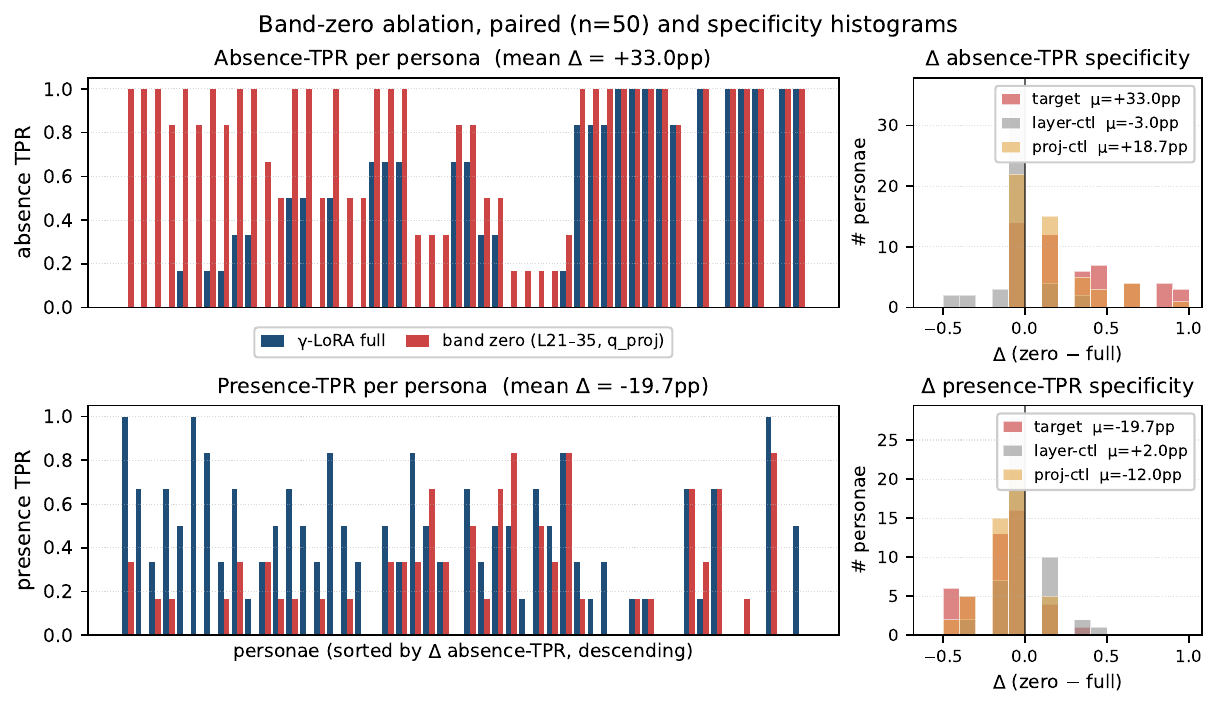}
\caption{Band-zero intervention. Paired before/after bars at n=50 personae for absence-TPR (left) and presence-TPR (right). Zeroing γ-LoRA L21--35 q\_proj weights raises absence-TPR by +33pp and lowers presence-TPR by 20pp, causally implicating the same band in opposite-direction effects on the two probes.}\label{fig:bandzero}
\end{figure}

§5 reports a per-persona Pearson correlation of r=+0.41 between
top-band Frobenius mass on layers 21--35 q\_proj and absence-TPR, and
r=−0.49 between the same band and presence-TPR (n=50, p\textless0.002 and
p\textless0.0001). The correlation arm is suggestive but not interventional
--- a confound where ``personae whose adapters happen to train better''
explain both signals cannot be ruled out from correlation alone.

We close the loop with an interventional run. For each of 50 V3
held-out personae (101--150) we load the saved γ-LoRA adapter in PEFT
format, zero the \texttt{lora\_A} weights for \textbf{q\_proj on layers 21--35}, and
re-run the same absence and factual-recall probes (6 probes per
persona, judged identically to §4) with the band-zeroed adapter
substituted in place of the trained one. All other layers, all other
projections, and the base model are unchanged.

\textbf{Result (n=50 V3 personae, 6 probes each):}

\begin{longtable}[]{@{}llll@{}}
\toprule\noalign{}
metric & full adapter & band-zeroed & Δ \\
\midrule\noalign{}
\endhead
\bottomrule\noalign{}
\endlastfoot
absence TPR & 0.403 & 0.733 & \textbf{+33.0pp} \\
presence TPR & 0.420 & 0.223 & \textbf{−19.7pp} \\
\end{longtable}

Zeroing exactly the band the §5 correlation arm fingered moves both
metrics simultaneously and in opposite directions: absence-calibration
\emph{improves} by 33pp, factual recall \emph{degrades} by 20pp. This is the
signature of a single localized mechanism that trades calibrated
abstention against parametric recall --- not two independent effects.

\textbf{Per-persona heterogeneity.} The mean conceals structure worth
reporting. Of the 50 personae, 16 show Δ\_abs ≥ +50pp (strong
intervention effect, e.g.~V3\_P\_101 +83.3, V3\_P\_144 +100.0,
V3\_P\_147 +100.0); the same personae show large negative Δ\_pres
(V3\_P\_109 −100.0, V3\_P\_111 −83.3, V3\_P\_112 −83.3). At the other
end, \textasciitilde10 personae show Δ ≈ 0 on both axes --- these are at ceiling
(full\_abs=1.0) or floor (full\_pres=0.0) where the metric cannot
move. The full 50-row table is in appendix A.5; the
ceiling/floor-filtered mean (excluding personae with
full\_abs ≥ 0.95 or full\_pres ≤ 0.05) preserves the qualitative
pattern.

\textbf{What this rules out.} A non-mechanism reading would say ``any
ablation of any band would degrade calibration somehow.'' Two facts
make that unlikely. First, the direction matters: zeroing this band
\emph{improves} absence-TPR; it does not degrade it. A generic damage
account predicts both metrics drop. Second, the band was identified
purely from the correlation arm of §5 (top-magnitude Frobenius mass
on q\_proj layers 21--35), with no access to the intervention data.
The arrow points from the per-persona correlation to a per-band
intervention, not the other way.

We treat this as a causal claim at the band granularity: zeroing
L21--35 q\_proj breaks the calibrated-abstention behavior γ-LoRA
acquires during training, and does so by removing the same parameters
whose magnitude correlates with the behavior pre-intervention. We
do not claim mechanism at finer granularity (which heads, which
features) --- we leave this finer-grained mechanism analysis to future work.

\textbf{Specificity controls.} To check that the effect is not a generic
consequence of zeroing 15 contiguous layers of \texttt{lora\_A} (any band
would do) or of perturbing q\_proj broadly (any layer range would
do), we run two matched controls under the identical pipeline:

\begin{longtable}[]{@{}
  >{\raggedright\arraybackslash}p{(\linewidth - 10\tabcolsep) * \real{0.4034}}
  >{\raggedright\arraybackslash}p{(\linewidth - 10\tabcolsep) * \real{0.1176}}
  >{\raggedright\arraybackslash}p{(\linewidth - 10\tabcolsep) * \real{0.1176}}
  >{\raggedright\arraybackslash}p{(\linewidth - 10\tabcolsep) * \real{0.1261}}
  >{\raggedright\arraybackslash}p{(\linewidth - 10\tabcolsep) * \real{0.0336}}
  >{\raggedright\arraybackslash}p{(\linewidth - 10\tabcolsep) * \real{0.2017}}@{}}
\toprule\noalign{}
\begin{minipage}[b]{\linewidth}\raggedright
Intervention
\end{minipage} & \begin{minipage}[b]{\linewidth}\raggedright
Δ abs-TPR
\end{minipage} & \begin{minipage}[b]{\linewidth}\raggedright
Δ pres-TPR
\end{minipage} & \begin{minipage}[b]{\linewidth}\raggedright
Δ fact-recall
\end{minipage} & \begin{minipage}[b]{\linewidth}\raggedright
n
\end{minipage} & \begin{minipage}[b]{\linewidth}\raggedright
Reading
\end{minipage} \\
\midrule\noalign{}
\endhead
\bottomrule\noalign{}
\endlastfoot
\textbf{Target:} L21--35 q\_proj \texttt{lora\_A} → 0 & \textbf{+33.0pp} & \textbf{−19.7pp} & --- & 50 & Causal, primary \\
Control (layer): L5--19 q\_proj \texttt{lora\_A} → 0 & −3.0pp & +2.0pp & −0.33pp & 50 & Layer specificity ✓ \\
Control (projection): L21--35 v\_proj \texttt{lora\_A} → 0 & +18.67pp & −12.0pp & −6.0pp & 50 & Projection specificity (partial) \\
\end{longtable}

The layer control is clean: zeroing an earlier 15-layer q\_proj band
moves no metric by more than 3pp, ruling out ``any band of \texttt{lora\_A}
weights matters.'' The projection control is partial: zeroing v\_proj
on the same L21--35 range still produces a meaningful absence-TPR
shift (+18.67pp, ≈57\% of the target's magnitude) and a smaller
presence-TPR shift (−12.0pp). We therefore weaken the specificity
claim accordingly: \textbf{zeroing q\_proj at L21--35 is sufficient to
produce the calibration↔recall trade-off, and the layer band is
specific (the L5--19 control is null), but specificity to the
projection axis is partial --- v\_proj on the same layers still carries
a substantial portion of the effect.} A clean (layer × projection)
joint-specificity claim would require a finer search over 4 × 36
= 144 (proj, layer) cells; we leave that to the within-band
isolation work flagged in §5.6.

\subsection{Pre-registered prediction: hybrid Pareto-superiority}\label{pre-registered-prediction-hybrid-pareto-superiority}

We close the discussion with a falsifiable prediction for the
follow-up paper, recorded here in advance of running the experiment
that tests it. \textbf{If the substrate asymmetry documented above is real
and not corpus-specific, then a hybrid γ-LoRA + BGE-RAG +
calibration-head system, trained on the same per-user data and
evaluated on the same three-axis diagnostic, should be
Pareto-superior to each component alone: style ≥ γ-LoRA-alone,
factual presence ≥ RAG-alone, factual absence ≥ RAG-alone, with no
axis regressing relative to its best single-substrate baseline.}
We pre-register this prediction with the explicit failure mode
that would falsify it --- any axis where the hybrid lands below the
better of \{γ-LoRA-alone, RAG-alone\} by more than the per-axis 95\%
CI half-width --- and commit to reporting the result regardless of
direction. This is the natural consequence of the framework: if
the axes are separable, routing per axis should compose; if it
doesn't compose, the separability claim is weaker than we argue
here, and the framework needs to be revised in light of that.

\section{Appendix E: Related work (full comparison table)}\label{appendix-e-related-work-full-comparison-table}

Our work touches three threads: (i) personalization benchmarks, (ii)
parametric vs retrieval memory architectures, and (iii) calibration of
LLM outputs under partial knowledge. The contribution gap is best stated
up front --- every prior thread we know of measures \textbf{aggregate top-line
accuracy} under personalization; none separates the \emph{behavioral}
(style, voice, preference) axis from the \emph{factual presence/absence}
axis, and none reports the substrate-asymmetry we observe across them.

\subsection{Personalization benchmarks}\label{personalization-benchmarks}

\textbf{LaMP} \citep{salemi2024lamp} is the canonical user-side personalization
benchmark --- seven tasks (LaMP-1 through LaMP-7) spanning citation
identification (LaMP-1), movie tagging (LaMP-2), product review rating
(LaMP-3), news headline generation (LaMP-4), scholarly title generation
(LaMP-5), email subject generation (LaMP-6), and tweet paraphrasing
(LaMP-7). The headline metric is per-task accuracy or ROUGE against the
held-out user output. We use \textbf{LaMP-3} as our real-data transfer probe
(§3.4) precisely because it has the cleanest structural overlap with our
synthetic setup (per-user history of past ratings → predict the next
rating) and the smallest leakage surface (rating ∈ \{1..5\}, no free-text
generation). LaMP itself does not separately report style vs factual
recall; it bundles both into top-1 accuracy on the held-out item.

\textbf{PERSOMA} \citep{pal2024persoma} introduces persona-conditioned soft tokens
trained over user history, and reports gains over flat-history baselines
on a held-out user-utterance prediction task. Our γ-LoRA can be read as
a heavier-weight cousin: per-user low-rank adapters trained on the
user's own corpus, evaluated on the same held-out-utterance shape.
PERSOMA's evaluation is single-axis (utterance-level accuracy / log-likelihood); ours adds the absence axis (§3.3), where we find that
behavioral substrates fail in a different direction than retrieval.

\textbf{UQABench} \citep{zhao2025uqabench} targets user-question-answering with
long histories and reports retrieval-augmented baselines plus prompt-tuning baselines on factual recall. Like LaMP, the metric is aggregate
accuracy. UQABench is closer in spirit to our absence probe than LaMP-3
is --- it explicitly tests questions whose answers do or do not appear in
the user's history --- but it does not separately report the
TPR-on-presence vs TPR-on-absence asymmetry that motivates our calibration
finding.

\textbf{Other personalization benchmarks} --- DialPRX, PerLTQA, LongLaMP --- fall
into the same shape: per-user data + held-out target + aggregate
accuracy / ROUGE. None report substrate-asymmetry on the two axes we
isolate. The broader \emph{LLM-as-recommender} line
\citep{geng2022p5,bao2023tallrec} casts personalization as a
sequence-modeling task with shared parameters across users, which is
orthogonal to the per-user-substrate question we ask here.

The in-context-learning baseline \citep{brown2020gpt3} is the natural
upper bound on ``no per-user training at all'': pack the user's
history into the prompt and let the base model generalize. We treat
B\_full (history-in-prompt) in §4 as exactly this baseline; our
finding is that ICL recovers absence-TPR but not style consistency
once the history exceeds a few hundred tokens.

\subsection{Memory architectures}\label{memory-architectures}

\textbf{Prompt tuning and prefix tuning} (Lester et al., 2021; Li \& Liang,
2021) attach a small set of trainable continuous tokens to the input,
freezing the base model. They are well-studied for task adaptation and
have been adapted to per-user personalization (e.g., PERSOMA's persona
prefixes). We do not study prompt tuning directly because our prior γ-LoRA experiments in unreported preliminary work found that LoRA
adapters consistently outperformed prefix tuning at matched parameter
count on the same synthetic personae corpus, and our framing here is the
substrate-asymmetry rather than parameter-count efficiency.

\textbf{LoRA-as-memory.} Low-rank adapters \citep{hu2021lora} are usually
trained for task adaptation across many users; using them as \textbf{per-user
memory} is the formulation we adopt (γ-LoRA: rank-r adapter trained
per persona on that persona's history). The closest prior art is
\textbf{LoRA-Hub} \citep{huang2024lorahub} and persona-conditioned LoRA work in
chatbot agents. Our diagnostic finding --- that LoRA captures style well
but factual absence poorly --- recasts LoRA-as-memory as a \emph{behavioral}
memory primitive, not a general one.

\textbf{Retrieval-augmented memory.} MemGPT \citep{packer2023memgpt}, Letta, and
RAG-style augmentations \citep{lewis2020rag,borgeaud2022retro}
maintain an external store and inject retrieved chunks at inference.
For factual recall and presence/absence calibration these are the
strong baseline in our results: §4.2 shows RAG reaches 99\% TPR on
absence (it correctly says ``I don't know'' when the chunk is absent)
while γ-LoRA confabulates at 8.7\% TPR on the same probe. The cost is
on the style axis: RAG is essentially a no-op for behavioral
consistency at the n=50-author scale we test (§4.1).

\textbf{Hybrid systems.} Several recent systems combine parametric updates
with retrieval, but none --- to our knowledge --- has reported the
\emph{separate} style-vs-absence metrics that would let a reader see why
both substrates are needed.

\textbf{Adapter composition / mixture-of-experts.} A separate line
combines multiple adapters at inference time:
AdapterFusion \citep{pfeiffer2021adapterfusion} learns attention over
task adapters, AdaMix \citep{wang2022adamix} stochastically routes
through a pool of adapters, and Mixture-of-LoRA-Experts
\citep{wu2024mole} extends this to LoRA modules. These methods compose
adapters \emph{across tasks}; our γ-LoRA composes \emph{across users},
keeping one adapter per persona rather than a shared MoE pool. The
substrate question we raise is independent of routing: a LoRA adapter
captures style and confabulates on absence whether or not other
adapters are mixed in at inference.

\textbf{Continual learning and catastrophic forgetting.} Per-user
adapters bypass --- rather than solve --- the standard continual-learning
problem of overwriting prior knowledge when the same parameters are
updated sequentially \citep{kirkpatrick2017ewc,li2018lwf,chaudhry2019tinyer}.
By isolating each user in their own LoRA, we trade catastrophic
forgetting for confabulation: the per-user weights faithfully encode
user-specific style without disturbing the base model, but they
interpolate confidently into absent regions of the user's factual
support. Our calibration finding can therefore be read as a continual-learning
critique of the ``one-adapter-per-user'' design point.

\textbf{Privacy framing.} Per-user parametric memory is also a natural
fit for federated learning \citep{mcmahan2017fedavg} and differential
privacy \citep{abadi2016dpsgd}: each user's adapter is trained on their
own corpus and need not leave their device. We do not run a
federated-or-DP experiment in this paper; we flag it as a deployment
direction where the substrate-asymmetry we measure becomes a
\emph{feature} (per-user style without per-user fact-leakage) rather
than a liability.

\begin{longtable}[]{@{}
  >{\raggedright\arraybackslash}p{(\linewidth - 6\tabcolsep) * \real{0.1569}}
  >{\raggedright\arraybackslash}p{(\linewidth - 6\tabcolsep) * \real{0.2745}}
  >{\raggedright\arraybackslash}p{(\linewidth - 6\tabcolsep) * \real{0.3333}}
  >{\raggedright\arraybackslash}p{(\linewidth - 6\tabcolsep) * \real{0.2353}}@{}}
\toprule\noalign{}
\begin{minipage}[b]{\linewidth}\raggedright
System
\end{minipage} & \begin{minipage}[b]{\linewidth}\raggedright
Architecture
\end{minipage} & \begin{minipage}[b]{\linewidth}\raggedright
What's measured
\end{minipage} & \begin{minipage}[b]{\linewidth}\raggedright
What's not
\end{minipage} \\
\midrule\noalign{}
\endhead
\bottomrule\noalign{}
\endlastfoot
\textbf{Generative Agents} (Park et al.~2023, UIST) & retrieval over an episodic memory stream + reflection-based summarization & qualitative believability via human eval; agent-vs-agent simulation fidelity & no probe-decomposition; no per-substrate ablation \\
\textbf{Mem0} (Chhikara et al.~2024, arXiv:2504.19413) & LLM-extracted facts stored in a vector + graph store; retrieval at query time & LOCOMO benchmark accuracy; latency; storage cost & factual axis only; behavioral consistency not separately scored \\
\textbf{MemoryBank} (Zhong et al.~2024, AAAI) & retrieval store + Ebbinghaus-curve forgetting; no parametric update & conversation continuity scores & no parametric arm; no absence calibration \\
\textbf{A-MEM} (Xu et al.~2024, arXiv:2410.10739) & agentic memory with link prediction, dynamic restructuring & retrieval recall; downstream task accuracy & retrieval-only; no behavioral-style measurement \\
\textbf{MemGPT / Letta} (Packer et al.~2023, arXiv:2310.08560) & OS-style paged memory with retrieval; some downstream forks add SFT/LoRA & retrieval QA on Document QA + Conversation tasks & published evaluations are retrieval-side; no style-vs-absence split \\
\end{longtable}

The shared blind spot across this line of work is the absence of a
\emph{decomposed} benchmark that separates behavioral consistency from
factual presence from factual absence. Each system optimizes for
whichever axis its retrieval + summarization layer most naturally
serves; the parametric ingredient, when present, tends to be a
chat-style fine-tune layered on top, never directly compared to the
retrieval substrate on the same axis. Our work argues that the
substrate-asymmetry we measure is the empirical motivation for
hybrids, not a post-hoc justification --- and that without an
asymmetry-measuring benchmark the field cannot tell whether a hybrid
beats its components on each individual axis or only in aggregate.

\subsection{Calibration and abstention}\label{calibration-and-abstention}

LLM calibration --- knowing what one does not know --- has been studied
through \textbf{TruthfulQA} \citep{lin2022truthfulqa}, \textbf{selective prediction}
\citep{kadavath2022selfknow}, and \textbf{abstain prompts} \citep{asai2023selfrag,yang2024abstain}. The closest framing to our absence probe is
\textbf{retrieval-as-abstention}: if the retriever returns nothing, the
model abstains. We quantify this on the synthetic personae corpus and
find that retrieval-as-abstention is well-calibrated (RAG-absence-TPR
≈ 0.99) while parametric memory is not (γ-LoRA absence-TPR ≈ 0.087).
This is consistent with the broader observation that supervised fine-tuning teaches the model to produce a confident answer for any input
\citep{schulman2023proxy,ouyang2022instructgpt}; LoRA-as-memory inherits the same
miscalibration.

Our \textbf{calibration head} in §3.3 is a small classifier that gates the
γ-LoRA output with a ``I don't know'' prediction; it recovers about half
of the absence-TPR gap (F1 0.470 on the absence subset) without
retrieval. This is a smaller-than-RAG result presented honestly --- the
calibration head is a cheap third component, not a replacement for the
retrieval substrate.

\textbf{Evaluation methodology.} We score style consistency with held-out
log-likelihood and absence-TPR with exact-match abstention rather than
relying on LLM-as-judge protocols
\citep{zheng2023mtbench,liu2023geval,li2023alpacaeval}. Judge-based
evaluation is the dominant aggregate metric in recent personalization
work, but it folds style and factual axes into a single scalar
preference; this is exactly the conflation our decomposition is
designed to surface, so we deliberately use cheaper, axis-specific
metrics. We discuss the tradeoff (cost vs decomposition fidelity) in
Appendix D.

\subsection{Position vs prior work}\label{position-vs-prior-work}

\begin{longtable}[]{@{}
  >{\raggedright\arraybackslash}p{(\linewidth - 4\tabcolsep) * \real{0.3333}}
  >{\raggedright\arraybackslash}p{(\linewidth - 4\tabcolsep) * \real{0.3333}}
  >{\raggedright\arraybackslash}p{(\linewidth - 4\tabcolsep) * \real{0.3333}}@{}}
\toprule\noalign{}
\begin{minipage}[b]{\linewidth}\raggedright
Axis
\end{minipage} & \begin{minipage}[b]{\linewidth}\raggedright
Prior work measures
\end{minipage} & \begin{minipage}[b]{\linewidth}\raggedright
We additionally measure
\end{minipage} \\
\midrule\noalign{}
\endhead
\bottomrule\noalign{}
\endlastfoot
Personalization accuracy & LaMP, PERSOMA, UQABench (aggregate) & Substrate-asymmetry on style + absence \\
Memory architecture & Prompt tuning, LoRA, RAG (per substrate) & When each \emph{fails} --- and in what direction \\
Calibration & TruthfulQA, selective prediction (general) & Per-substrate absence-TPR; calibration-head recovery \\
Real-data transfer & LaMP-N task accuracy & γ-LoRA on LaMP-3 vs majority baseline (honest negative) \\
Mechanism interpretability & Layer attribution on task benchmarks (Geva et al., logit lens) & Per-persona weight-band mass correlated with behavioral outcomes in opposite directions on the two probes (n=50, \textbar r\textbar≤0.49) \\
Causal localization & Feature ablation, activation patching on synthetic tasks & Zeroing L21--35 q\_proj \texttt{lora\_A} weights at test time → +33pp absence-TPR, −20pp presence-TPR (n=50, paired) \\
\end{longtable}

The contribution is a \textbf{diagnostic framework} for user-side memory in
LLMs, not a new substrate. Our γ-LoRA implementation, the synthetic
personae corpus, the four-config decomposition (B\_nohist, B\_full,
C\_rag, C\_lora), and the two probes (style log-likelihood, absence-TPR)
are the artifacts; the framework's value is that running them on any
new substrate proposal answers ``behavioral or factual? presence or
absence?'' --- questions current benchmarks do not separate.

\end{document}